\documentclass[sigconf]{acmart}

\usepackage{booktabs} 

\usepackage{url,mathrsfs}
\usepackage{amsmath,wrapfig,color}
\usepackage{amsfonts}
\usepackage{graphicx}
\usepackage{subfigure,algorithm}

\setcopyright{rightsretained}

\acmDOI{10.475/123_4}

\acmConference[2018]{}{}{}
\acmYear{2018}
\copyrightyear{2018}

\begin{document}
\title{Several Tunable GMM Kernels}

\author{Ping Li}
\affiliation{%
  \institution{Baidu Research USA}
}
\email{pingli98@gmail.com}

\begin{abstract}
\noindent While tree methods have been popular in practice, researchers and practitioners are also  looking for simple algorithms which can reach  similar accuracy of trees. In 2010, ~\cite{Proc:ABC_UAI10} developed the method of ``abc-robust-logitboost''  and compared it with other supervised learning methods on datasets used by the deep learning literature. In this study, we propose a series of ``tunable GMM kernels'' which are simple  and perform largely comparably to tree methods on the same datasets. Note that ``abc-robust-logitboost''~\cite{Proc:ABC_UAI10} substantially improved the original ``GDBT'' in that (a) it developed a tree-split formula based on second-order information of the derivatives of the loss function; (b) it developed a new set of derivatives for multi-class classification formulation.

\vspace{0.08in}

\noindent In the prior study~\cite{Proc:Li_KDD17}, the ``generalized min-max'' (GMM) kernel was shown to have good performance compared to the ``radial-basis function'' (RBF) kernel. However, as demonstrated in this paper, the original GMM kernel is often not as competitive as tree methods on the datasets used in the deep learning literature. Since the original GMM kernel has no parameters, we propose tunable GMM kernels by adding tuning parameters in various ways. Three basic (i.e., with only one parameter) GMM kernels are the ``$e$GMM kernel'', ``$p$GMM kernel'', and ``$\gamma$GMM kernel'', respectively.  Extensive experiments show that they are able to produce good results for a large number of classification tasks. Furthermore, the basic kernels can be combined  to  boost the performance.

\vspace{0.08in}

\noindent For large-scale machine learning tasks, it is crucial that learning methods should be able to scale up with the size of the training samples. It has been known that the original GMM kernel can be  efficiently linearized (i.e., achieving the result of a nonlinear kernel at the cost of a linear kernel). As demonstrated in this paper, several tunable GMM kernels also inherit this nice property in that they can also be efficiently linearized.

\end{abstract}



\maketitle

\section{Introduction}

Kernel methods~\cite{Book:Scholkopf_02} are an important part of machine learning. Among many types of kernels, the linear kernel and the ``radial basis function'' (RBF) kernel are probably the most well-known. Recently, the ``generalized min-max'' (GMM) kernel~\cite{Report:Li_GMM16} was introduced for large-scale search and machine learning, owing to its efficient linearization  via either hashing or  the Nystrom method~\cite{Article:Nystrom1930}. For defining the GMM kernel, the first step is a simple transformation on the original data. Consider, for example,   the original data vector $u_i$, $i=1$ to $D$. We define the following transformation, depending on whether an entry $u_i$ is positive or negative:
\begin{align}\label{eqn_transform}
 \left\{\begin{array}{cc}
\tilde{u}_{2i-1} = u_i,\hspace{0.1in} \tilde{u}_{2i} = 0&\text{if } \ u_i >0\\
\tilde{u}_{2i-1} = 0,\hspace{0.1in} \tilde{u}_{2i} =  -u_i &\text{if } \ u_i \leq 0
\end{array}\right.
\end{align}
For example, when $D=2$ and $u = [-4\ \ 6]$, the transformed data vector becomes $\tilde{u} = [0\ \ 4\ \ 6\ \ 0]$. The GMM kernel is defined~\cite{Report:Li_GMM16} as  follows:
\begin{align}
GMM(u,v) = \frac{\sum_{i=1}^{2D}\min\{\tilde{u}_i,\tilde{v}_i\}}{\sum_{i=1}^{2D} \max\{\tilde{u}_i,\tilde{v}_i\}}
\end{align}

Even though the GMM kernel has no tuning parameter, it performs surprisingly well for classification tasks as empirically demonstrated in~\cite{Report:Li_GMM16} (also see Table~\ref{tab_UCI} and Table~\ref{tab_MNIST}), when compared to the linear kernel and best-tuned radial basis function (RBF) kernel:
\begin{align}
RBF(u,v;e) = e^{-\lambda_e\left(1-\frac{\sum_{i=1}^{D}u_iv_i}{\sqrt{\sum_{i=1}^{D} u_i^2\sum_{i=1}^{D}v_i^2}}\right)}
\end{align}
where $\lambda_e>0$ is the tuning parameter.

 Furthermore, the (nonlinear) GMM kernel can be efficiently linearized via  hashing~\cite{Report:Manasse_CWS10,Proc:Ioffe_ICDM10,Proc:Li_KDD15} (or the Nystrom method~\cite{Article:Nystrom1930}). This means we can use the linearized GMM kernel for large-scale machine learning tasks essentially at the cost of linear learning. \\

Given the deceiving simplicity of the GMM kernel and its surprising  performance compared to the RBF kernel, researchers and practitioners  might be seriously interested in asking two questions:
\begin{enumerate}
\item Does the GMM kernel perform comparably to more sophisticated learning methods such as trees, at least in the context of supervised learning when  features are already available?\\

\item Can one improve the accuracy of the original (tuning-free) GMM kernel, for example, by adding tuning parameters?
\end{enumerate}

This papers aims at addressing these two questions. It turns out that, in many datasets, the original (tuning-free) GMM kernel can be substantially improved, by adding tuning parameters. Furthermore, we report a comparison study using a set of public datasets in the deep learning literature~\cite{Proc:Larochelle_ICML07}. This set of datasets were used by an empirical study~\cite{Proc:ABC_UAI10} to compare several boosting \& tree methods with deep nets.  This paper will show that tunable GMM kernels can achieve comparably accuracy as trees.

\subsection{Tunable GMM Kernels}

In order to improve the performance of the original (tuning-free) GMM kernels, we propose three basic tunable GMM kernels:
\begin{align}
&e\text{GMM}(u,v;\lambda_e) = e^{-\lambda_e\left(1-\frac{\sum_{i=1}^{2D}\min\{\tilde{u}_i,\tilde{v}_i\}}{\sum_{i=1}^{2D} \max\{\tilde{u}_i,\tilde{v}_i\}}\right)}\\
&p\text{GMM}(u,v;p) =  \frac{\sum_{i=1}^{2D}\left(\min\{\tilde{u}_i,\tilde{v}_i\}\right)^p}{\sum_{i=1}^{2D}\left( \max\{\tilde{u}_i,\tilde{v}_i\}\right)^p}\\
&\gamma \text{GMM}(u,v;\gamma) =  \left(\frac{\sum_{i=1}^{2D}\left(\min\{\tilde{u}_i,\tilde{v}_i\}\right)}{\sum_{i=1}^{2D}\left( \max\{\tilde{u}_i,\tilde{v}_i\}\right)}\right)^\gamma
\end{align}
and the combinations of the basic tunable GMM kernels:
\begin{align}
&p\gamma\text{GMM}(u,v;p,\gamma) =  \left(\frac{\sum_{i=1}^{2D}\left(\min\{\tilde{u}_i,\tilde{v}_i\}\right)^p}{\sum_{i=1}^{2D}\left( \max\{\tilde{u}_i,\tilde{v}_i\}\right)^p}\right)^\gamma\\
&ep\text{GMM}(u,v;\lambda_e,p) =  e^{-\lambda_e\left(1-\frac{\sum_{i=1}^{2D}\left(\min\{\tilde{u}_i,\tilde{v}_i\}\right)^{p}}{\sum_{i=1}^{2D}\left( \max\{\tilde{u}_i,\tilde{v}_i\}\right)^{p}}\right)}\\
&e\gamma\text{GMM}(u,v;\lambda_e,\gamma) =  e^{-\lambda_e\left(1-\left(\frac{\sum_{i=1}^{2D}\left(\min\{\tilde{u}_i,\tilde{v}_i\}\right)}{\sum_{i=1}^{2D}\left( \max\{\tilde{u}_i,\tilde{v}_i\}\right)}\right)^\gamma\right)}\\
&ep\gamma\text{GMM}(u,v;\lambda_e,p,\gamma) =  e^{-\lambda_e\left(1-\left(\frac{\sum_{i=1}^{2D}\left(\min\{\tilde{u}_i,\tilde{v}_i\}\right)^{p}}{\sum_{i=1}^{2D}\left( \max\{\tilde{u}_i,\tilde{v}_i\}\right)^{p}}\right)^\gamma\right)}
\end{align}

In this study, we will provide an empirical study on kernel SVMs based on the  tunable GMM kernels. Perhaps not surprisingly, the improvements can be substantial on many datasets. In particular, we will also compare them with  tree methods on 11 datasets used by the deep learning literature~\cite{Proc:Larochelle_ICML07} and later by~\cite{Proc:ABC_UAI10}.

\subsection{The GMM Kernels versus Tree Methods}

\cite{Proc:ABC_ICML09,Proc:ABC_UAI10} developed several boosting \& tree methods including ``abc-mart'', ``robust logitboost'', and ``abc-robust-logitboost'' and demonstrated their  performance on  11 datasets used by the deep learning literature~\cite{Proc:Larochelle_ICML07}. The good accuracy was achieved by establishing the second-order tree-split formula and new derivatives for multi-class logistic loss. See Table~\ref{tab_MNIST} for more information on those datasets.

\begin{figure}[h!]
\begin{center}
\includegraphics[width=3in]{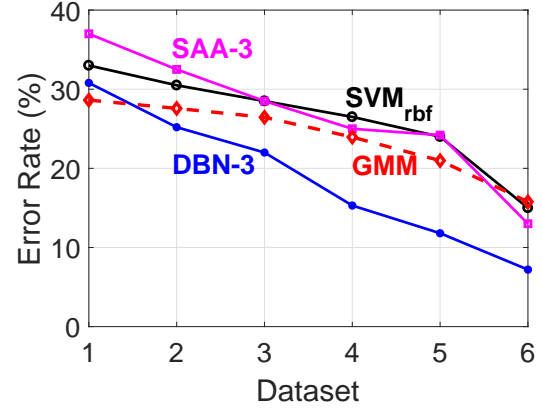}
\includegraphics[width=3in]{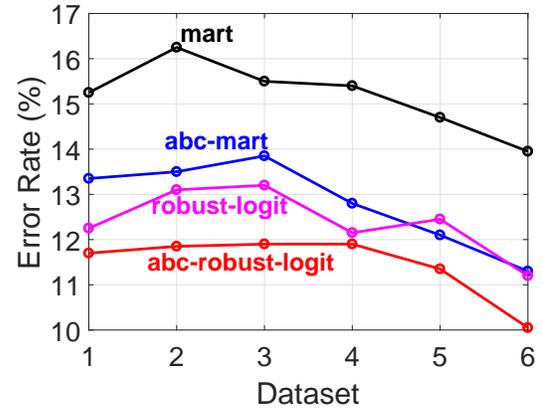}

\end{center}
\vspace{-0.15in}
\caption{Classification test error rates (lower the better) on 6 datasets (M-Noise1, M-Noise2, ..., M-Noise6) listed in Table~\ref{tab_MNIST} which were used by the deep learning literature~\cite{Proc:Larochelle_ICML07} and later by~\cite{Proc:ABC_UAI10} for comparing tree methods. On these datasets, the results of the GMM kernel (dashed curve in upper panel) are not as accurate as the results produced by the four tree \& boosting algorithms (bottom panel).}\label{fig_GMM_Trees}\vspace{-0.1in}
\end{figure}

Figure~\ref{fig_GMM_Trees} presents the classification accuracy results on 6 datasets, suggesting that the GMM kernel (upper panel) does not perform as well as tree methods (bottom panel). This observation has motivated us to develop tunable GMM kernels. Later in the paper, Figure~\ref{fig_Noise6} will show that the tunable GMM kernels are able to produce roughly comparable results as trees.

\subsection{Connection to the Resemblance Kernel}

The GMM kernel is  related to several similarity measures widely used in data mining and web search. When the data are nonnegative,  GMM becomes the  ``min-max'' kernel, which has been studied in the literature~\cite{Proc:Kleinberg_FOCS99,Proc:Charikar,Report:Manasse_CWS10,Proc:Ioffe_ICDM10,Proc:Li_KDD15}. When the data are binary (0/1), GMM becomes the well-known ``resemblance'' similarity. The minwise hashing algorithm for approximating resemblance has been a highly successful tool in web search for numerous  applications~\cite{Proc:Broder_WWW97,Proc:Fetterly_WWW03,Proc:Li_Konig_WWW10,Proc:Bendersky_WSDM09,Article:Forman09,Proc:Cherkasova_KDD09,
Article:Dourisboure09,Proc:Chierichetti_KDD09,Proc:Gollapudi_WWW09,Proc:Najork_WSDM09,Proc:HashLearning_NIPS11}. Note that for the $p$GMM kernel and nonnegative data, when $p\rightarrow0$, it also approaches the resemblance.

\subsection{Linearization of Nonlinear Kernels}

It is common in practice to use linear   learning algorithms such as  logistic regression or linear SVM. It is also known that one can often improve the performance of linear methods by using nonlinear algorithms such as kernel SVMs, if the computational/storage burden can be resolved. A straightforward  implementation of a nonlinear kernel, however,  can be difficult for large-scale datasets~\cite{Book:Bottou_07}.  For example, for  a small dataset with merely $60,000$ data points, the $60,000 \times 60,000$ kernel matrix  has $3.6\times 10^9$ entries.  In practice, being able to linearize nonlinear kernels becomes highly beneficial. Randomization (hashing) is a popular tool for kernel linearization. After data linearization, we can then apply our favorite linear learning packages such as LIBLINEAR~\cite{Article:Fan_JMLR08} or SGD (stochastic gradient descent)~\cite{URL:Bottou_SGD}. In this study, we focus on linearizing the $p$GMM kernels via hashing and we will also discuss how to linearize the $e$GMM and $\gamma$GMM kernels.

\vspace{0.08in}

Next, we  present an experimental study on the a large number of classification tasks using the variety of kernels we have discussed.


\begin{table*}[h!]
\caption{\textbf{Public classification datasets  and $l_2$-regularized kernel SVM results}. We report the test classification accuracies for the linear kernel, the best-tuned RBF kernel, the original (tuning-free) GMM kernel, the best-tuned $e$GMM,  $p$GMM, and $\gamma$GMM kernels, at their individually-best SVM regularization  $C$ values. All datasets are from the UCI repository except for the last 11 datasets, which were  used by~\cite{Proc:Larochelle_ICML07,Proc:ABC_UAI10} for testing deep learning algorithms and tree methods.\vspace{-0.1in}
}
\begin{center}{
{\begin{tabular}{l r r r c c c l l l}
\hline \hline
Dataset     &\# train  &\# test  &\# dim &linear  &RBF  &GMM &$e$GMM ($\lambda_e$) &$p$GMM ($p$) &$\gamma$GMM ($\gamma$)   \\
\hline
Car &864 &864 &6 &71.53   &94.91  &98.96   &99.31 (2)    &{\bf99.54} (2) &99.31 (6)\\
Covertype25k &25000 &25000 &54 &62.64 &{82.66} &82.65  &{88.32} (20)   &83.25 (0.6) &{\bf88.34} (20) \\
CTG &1063 &1063 &35 &60.59   &89.75  &88.81   &88.81 (0.01)  &{\bf100.00} (0.1) &90.78 (0.3)\\
DailySports &4560 &4560 &5625  & 77.70 &97.61 &{99.61} &{99.61} (0.2) &{99.61} (0.6) &{\bf 99.63} (0.8) \\
DailySports2k&2000&7120&5625& 72.16  &93.71   &98.99  &99.00 (0.1)  &{99.07} (.75) &{\bf 99.16} (0.3) \\
Dexter &300 &300 &19999 &92.67   &93.00  &94.00   &94.00 (17)  &{94.67} (0.5) & {\bf95.67}(0.2) \\
EEGEye &7490 &7490 &14 &61.46   &86.82   &78.54   &95.51 (1000)  &87.65 (15)  &91.20 (60)\\
Gesture &4937 &4936 &32 & 37.22   &61.06   &{65.50}   &{66.67} (1.9)   &66.33 (0.6)&{\bf67.16} (2.6)\\
ImageSeg &210 &2100 &19 &83.81   &91.38   &{95.05}   &95.38 (1.2) & {\bf95.57} (0.6) &95.38 (1.8)\\
Isolet2k &2000 &5797 &617 & 93.95   &{95.55} &95.53 &{95.55} (0.2) &95.53 (1.0) &{\bf95.60}(0.8)\\
MHealth20k&20000&20000&23&72.62   &82.65   &{85.28} &85.33  (0.5) &{\bf86.69} (0.5)  &85.31 (1.3) \\
MiniBooNE20k&20000&20000&50&88.42   &93.06   &{93.00}  &93.01 (0.2)  &{\bf93.69} (0.6) & 93.01 (1.1) \\
MSD20k &20000 &20000 &90 &66.72 &68.07 &{71.05}  &71.18 (0.2)   &{\bf71.84} (0.5) &71.44 (0.6)\\
Magic &9150 &9150 &10 &78.04   &84.43  &{87.02} &86.93 (0.3)  &{\bf87.57} (0.5) &87.09 (0.8)\\
Musk &3299 &3299 &166  & 95.09 &\textbf{99.33} &99.24 &99.24 (0.3) &99.24 (1.0) &99.24 (1.0) \\
Musk2k&2000&4598&166&94.80   &97.63   &{98.02}  &{98.02} (0.01)   &{\bf98.06} (1.25) &{\bf98.06} (0.5) \\
PageBlocks &2737  &2726 &10 &95.87   &{97.08}   &96.56   &{96.56} (1.4) &{\bf97.30} (0.1) & 96.64(0.8)\\
Parkinson &520&520&26&61.15   &66.73   &{69.81}   &{\bf70.19} (0.6) &69.81 (1.0) & {\bf70.19} (1.7)\\
PAMAP101 &20000 &20000 &51 &76.86   &96.68  &{98.91}  &98.91 (0.1) &{\bf99.00} (1.5) & 98.92 (1.1) \\
PAMAP102 &20000 &20000 &51 &81.22   &95.67  &{\bf98.78}  &98.77 (0.01) &{\bf98.78} (2) &{\bf98.78}(1.7)\\
PAMAP103 &20000 &20000 &51 & 85.54  &97.89  &{99.69}  &{\bf99.70} (0.01) &99.69 (1.0) &{\bf99.70} (0.8)\\
PAMAP104 &20000 &20000 &51 &84.03  &97.32   &{99.30}  &{\bf99.31} (0.6) &99.30 (1.0) &{\bf99.31} (1.3)\\
PAMAP105 &20000 &20000 &51 &79.43  &97.34   &{99.22} &{99.24} (1.1) &99.22 (0.75) &{\bf99.26}(1.8) \\
PIMA &384 &384 &8 &66.67   &71.35   &76.30  &77.08 (12)  &76.56 (0.75)   &76.82 (9.5)\\
RobotNavi &2728 &2728 &24 &69.83   &90.69   &{96.85}   &96.77 (0.1) &{\bf98.20} (0.1) &97.65 (0.3) \\
Satimage &4435 &2000 &36 &72.45   &85.20   &{90.40}   &{\bf91.85} (35)  &90.95 (5) & 91.35(9.5)\\
SEMG1 &900 &900 &3000  &26.00   &\textbf{43.56}   &41.00 &41.22 (0.1) &42.89 (0.25)  &42.11 (1.7)\\
SEMG2 &1800 &1800 &2500  &19.28  &29.00    &{54.00} &54.00 (0.3) &{\bf56.11} (2) & 55.22 (0.6) \\
Sensorless &29255 &29254 &48 &61.53 &93.01  &{99.39} &99.38 (0.01) &{\bf99.76} (0.5)  &99.62 (0.5) \\
Shuttle500 &500 &14500 &9 &91.81   &99.52  &{99.65}   &99.65 (0.1)  &{99.66} (0.5) &{\bf99.68} (0.4) \\
SkinSeg10k&10000&10000&3& 93.36   &99.74 &{99.81} &{\bf99.90} (20)   &99.85 (5) &99.87 (8.5)   \\
SpamBase &2301&2300&57&   85.91&   92.57 & {94.17}   &94.13  (0.6) &{\bf95.78} (0.25) &94.17 (1.0)\\
Splice &1000&2175&60&85.10   &90.02   &{95.22}   &{\bf96.46} (5)    &95.26 (1.25)  &{\bf 96.46} (5)\\
Theorem &3059&3059&51&   67.83   &70.48   &{71.53}   &{\bf71.69} (1.6)  &71.53 (1.0) &{\bf 71.76} (2.1) \\
Thyroid &3772 &3428 &21 &95.48   &97.67   &{98.31}   &98.34 (0.3)  &{\bf99.10} (0.1) &98.63 (0.6)\\
Thyroid2k &2000&5200&21&   94.90   &97.00  &{98.40}   &98.40  (0.01)  &{\bf98.96} (0.1) &98.62 (0.6) \\
Urban &168&507&147&   62.52   &51.48  &{66.08}   & 65.68 (0.5)  &{\bf83.04} (0.1) &67.26 (0.2) \\
Vertebral &155&155&6&   80.65   &83.23 &{89.04}   &{\bf89.68} (1.4)   &89.04 (1.0) & {\bf89.68}(1.1)\\
Vowel &264 &264 &10 &39.39 &94.70  &{96.97}  &{\bf98.11} (5)   &96.97 (1.0) &{\bf98.11} (4) \\
Wholesale &220&220&6&   89.55   &90.91   &{93.18}   &93.18 (0.6)  &{\bf93.64} (1.25)  &{\bf93.64} (0.2) \\
Wilt &4339&500&5&62.60   &83.20   &{87.20}  &{\bf87.60} (1.1) & 87.40 (0.75) &{\bf87.60} (1.7)\\
YoutubeAudio10k&10000&11930&2000&   41.35   &48.63   &50.59 &50.60 (0.01)  &{\bf51.84} (0.6)  & 50.84 (0.9)   \\
YoutubeHOG10k&10000&11930&647&   62.77   &66.20   &{68.63}  &68.65 (0.01)  &{\bf72.06} (0.5) &68.63 (1.1) \\
YoutubeMotion10k&10000&11930&64& 26.24  &28.81   &{31.95} &{\bf33.05} (4) &32.65 (0.6) & 32.98 (4.5) \\
YoutubeSaiBoxes10k&10000&11930&7168&46.97   &49.31   &51.28  &51.22 (0.001)   &{\bf52.15} (0.6) & 51.39 (0.8) \\
YoutubeSpectrum10k&10000&11930&1024&26.81   &33.54   &{39.23}  &39.27 (0.1)   &{\bf41.23} (0.5) & 39.28 (1.1) \\
\hline
M-Basic   &12000 &50000 &784 & 89.98   &{\bf97.21}   &96.34  &96.47 (1.2)  & 96.40 (0.5) &96.84 (2.3)  \\
M-Image &12000 &50000 &784 & 70.71 &77.84  &{80.85} &81.20 (1.5) &\textbf{89.53} (50) & 81.32 (2.1) \\
M-Noise1 &10000 &4000 &784 &60.28   &66.83    &{71.38} &71.70 (0.5)   &\textbf{85.20} (80) &71.90 (2.8) \\
M-Noise2 &10000 &4000 &784 & 62.05  & 69.15   &{72.43} &72.80 (3)   &\textbf{85.40} (70) & 72.95 (2.8) \\
M-Noise3 &10000 &4000 &784 &65.15   &71.68   &{73.55} &74.70  (3) &\textbf{86.55} (50) & 74.83 (3) \\
M-Noise4 &10000 &4000 &784 & 68.38  &75.33   &{76.05} &76.80 (2.5)  &\textbf{86.88} (60) &77.03 (2.8)\\
M-Noise5 &10000 &4000 &784 &72.25   &78.70  &{79.03} &79.48  (3)  &\textbf{87.33} (30) & 79.70 (3.5)\\
M-Noise6 &10000 &4000 &784 &78.73   &{85.33} &84.23  &84.58 (2)  &\textbf{88.15} (20) &84.68 (4)\\
M-Rand &12000 &50000 &784 & 78.90   &{85.39}   &84.22 &84.95 (4)  &\textbf{89.09} (40) &85.17 (3.5)\\
M-Rotate &12000 &50000   &784 &47.99  &\textbf{89.68}  & 84.76 &86.02 (1.6) &{86.52} (0.25) & 87.33 (2.1)\\
M-RotImg &12000 &50000 &784 &31.44  &{45.84}   & 40.98 &42.88 (4)  &\textbf{54.58}  (20) &43.22 (3.5) \\
\hline\hline
\end{tabular}}
}
\end{center}\label{tab_UCI}\vspace{-0.5in}

\end{table*}

\begin{figure*}[t]
\begin{center}
\mbox{
\includegraphics[width=1.7in]{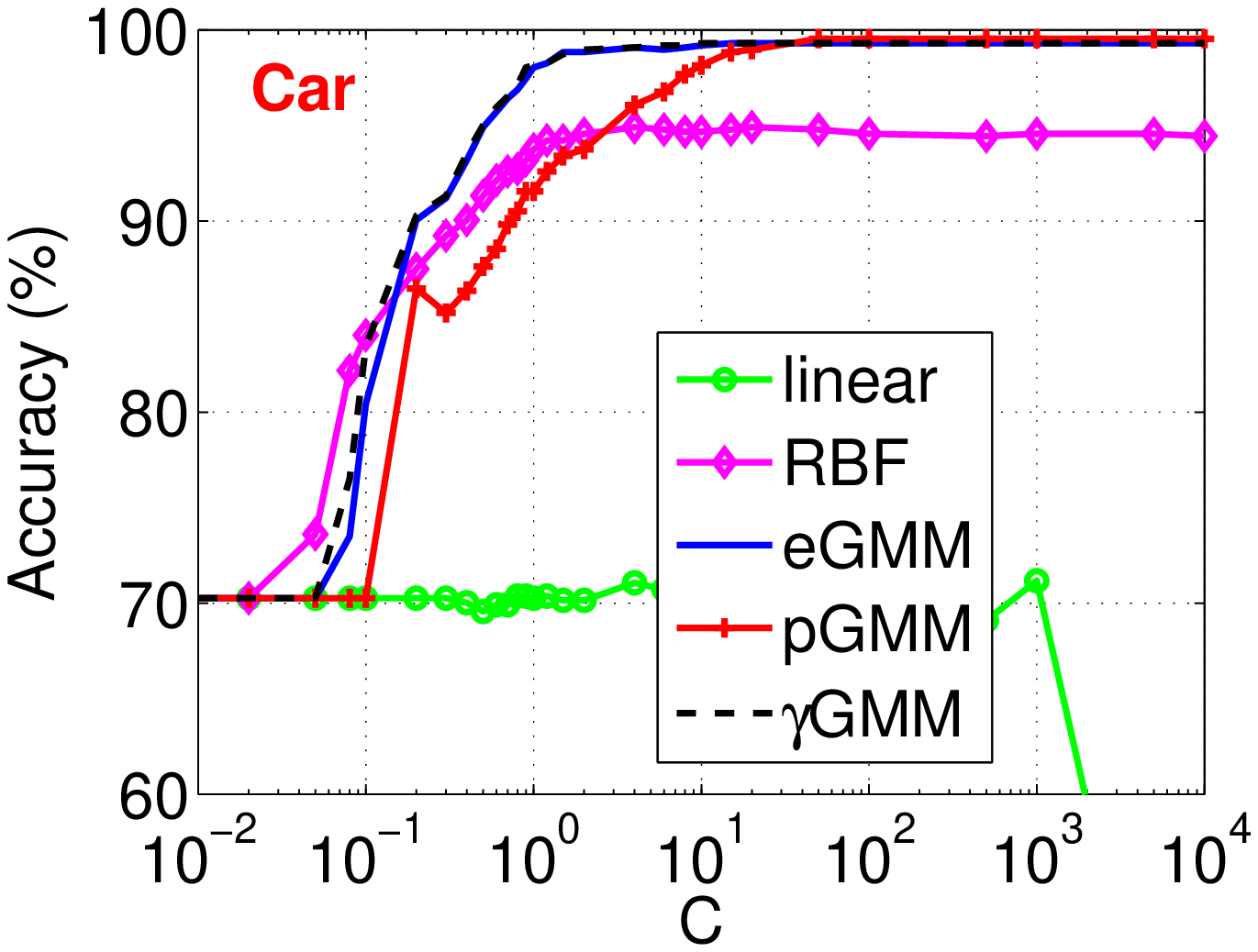}\hspace{-0in}
\includegraphics[width=1.7in]{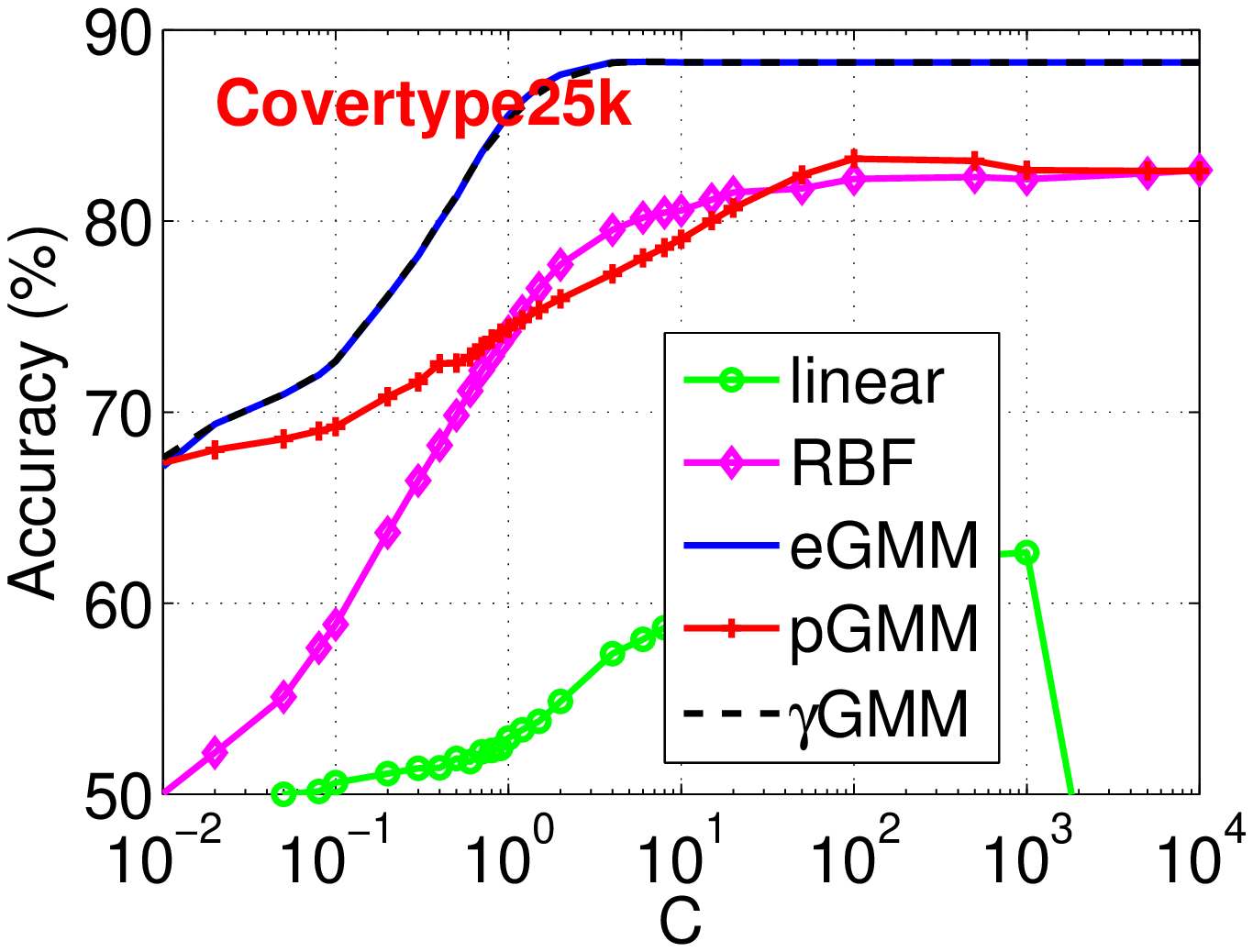}\hspace{-0in}
\includegraphics[width=1.7in]{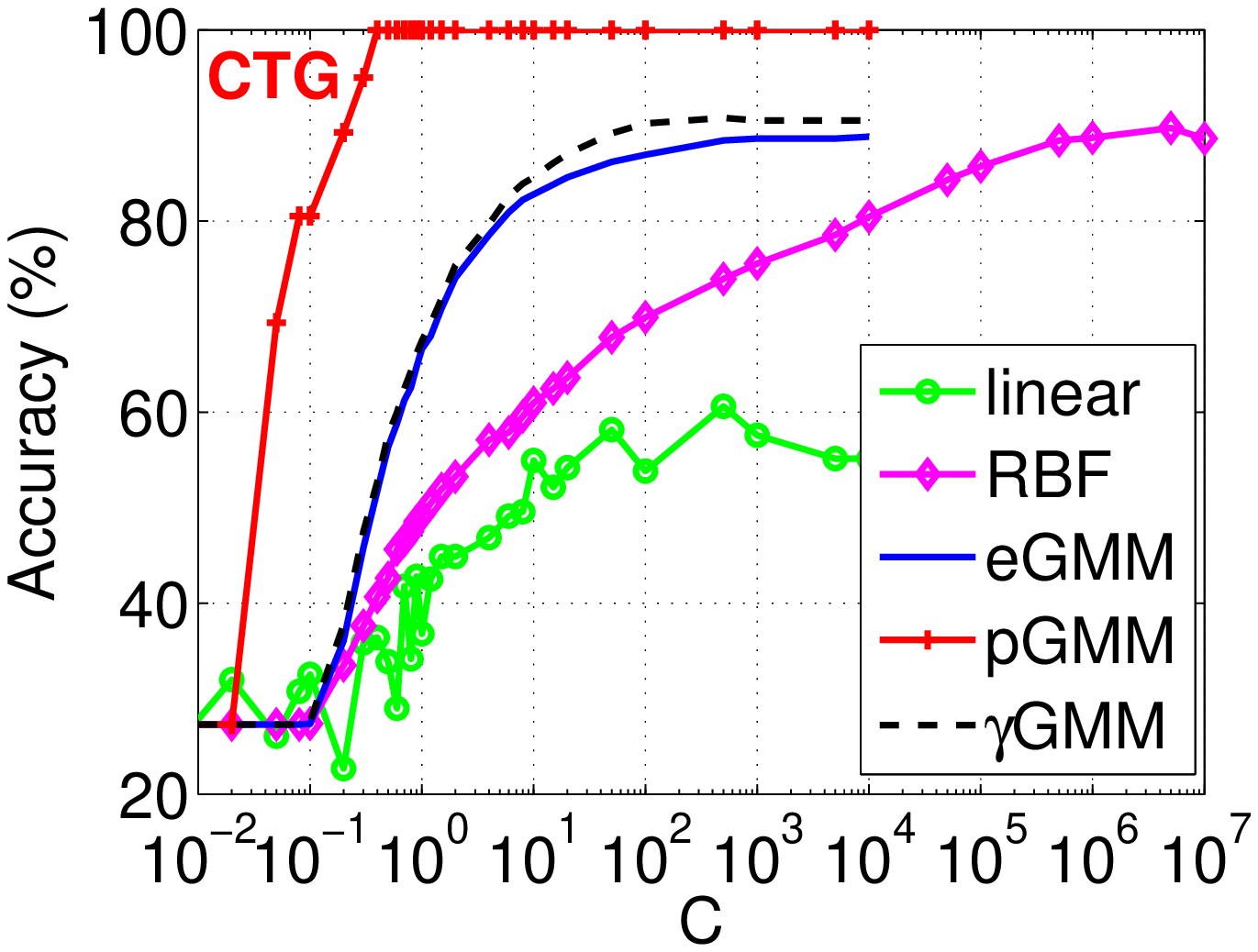}\hspace{-0in}
\includegraphics[width=1.7in]{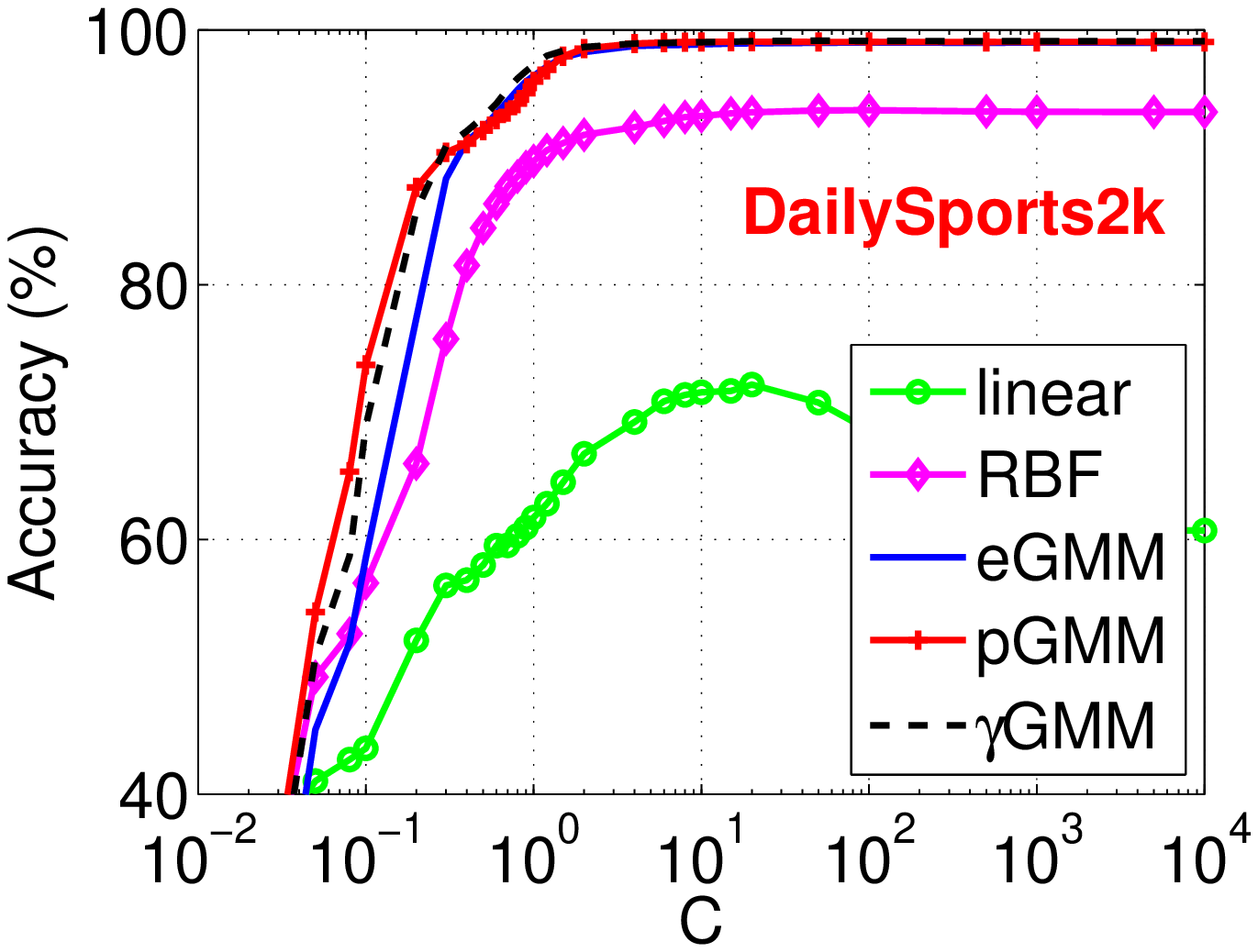}
}
\mbox{
\includegraphics[width=1.7in]{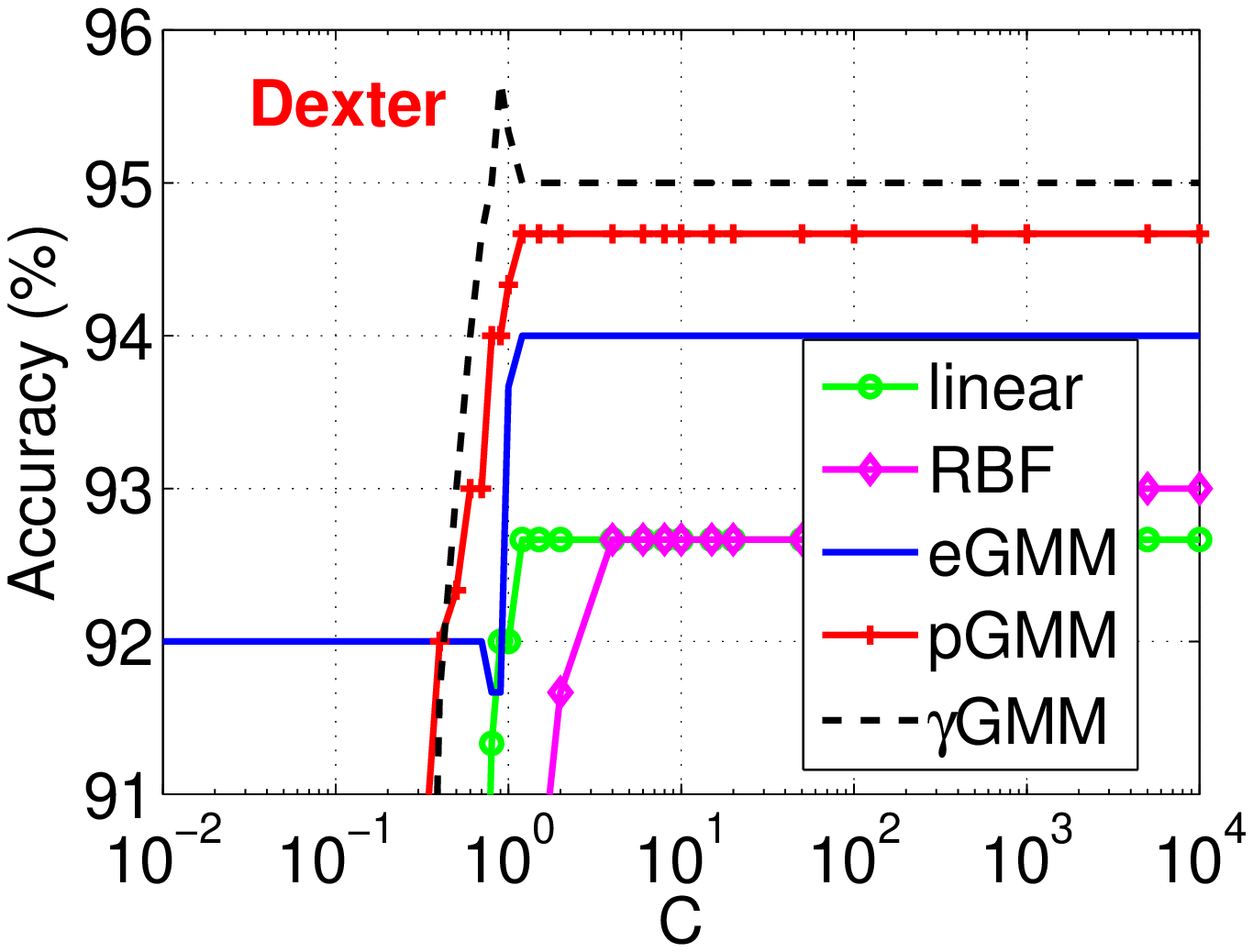}\hspace{-0in}
\includegraphics[width=1.7in]{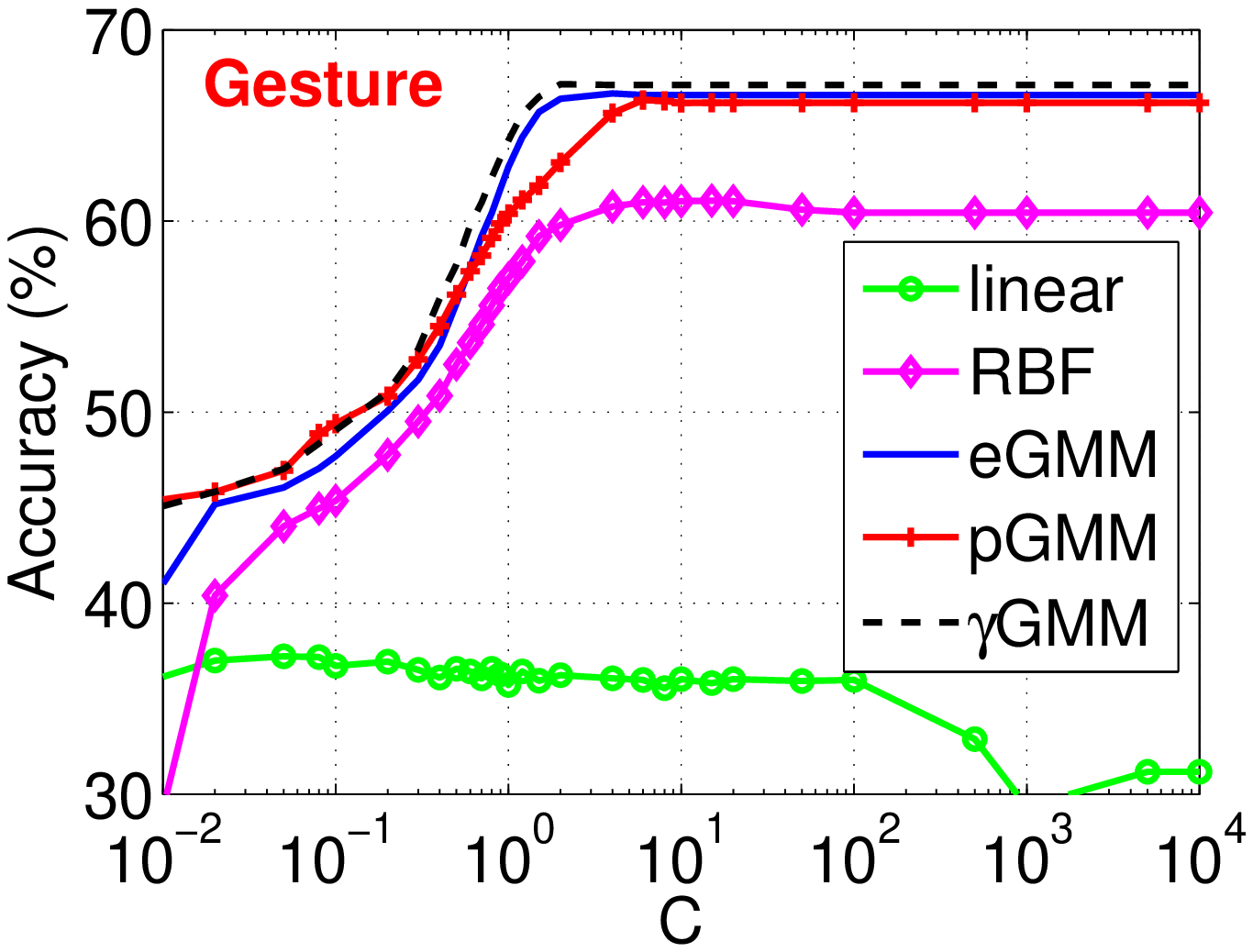}\hspace{-0in}
\includegraphics[width=1.7in]{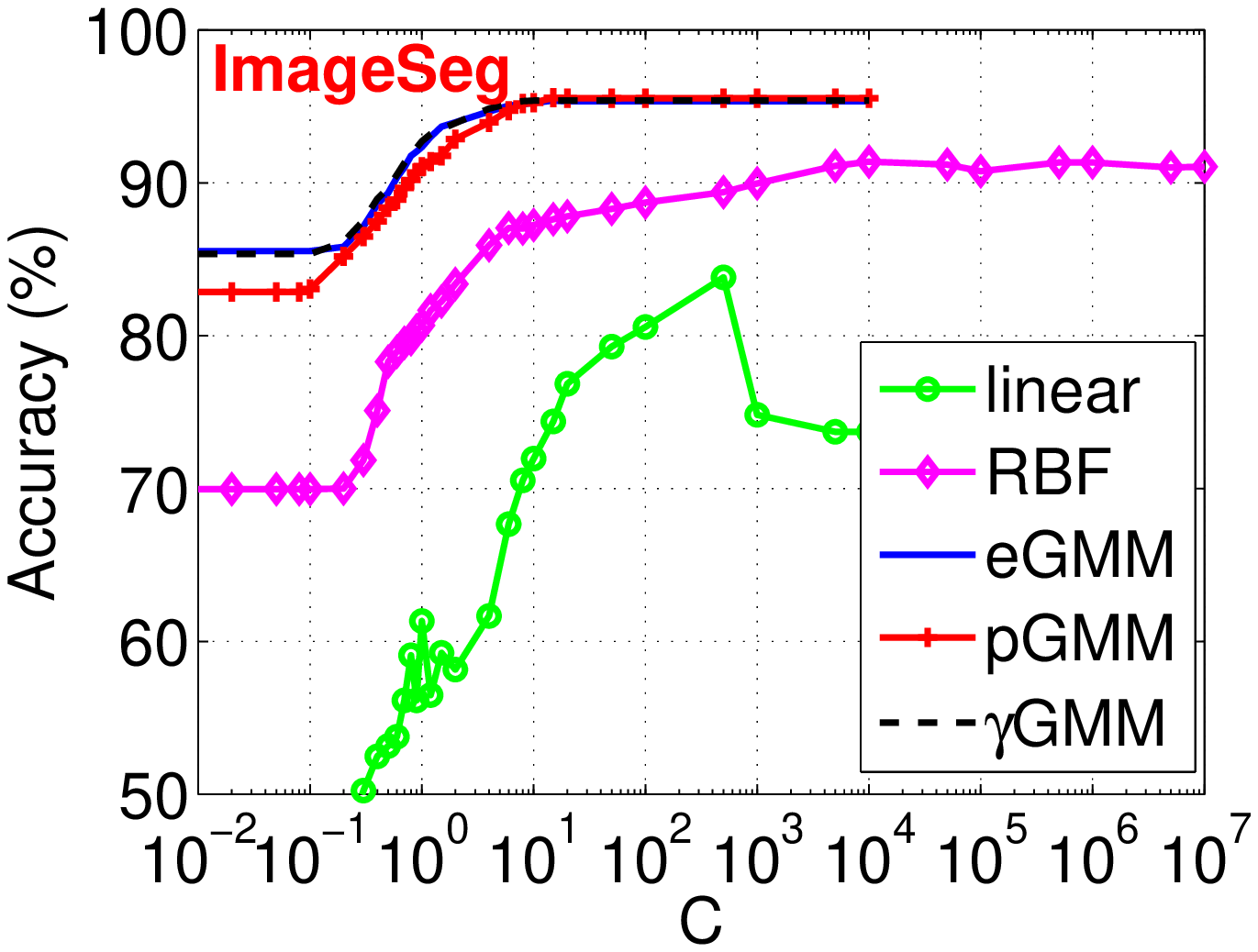}\hspace{-0in}
\includegraphics[width=1.7in]{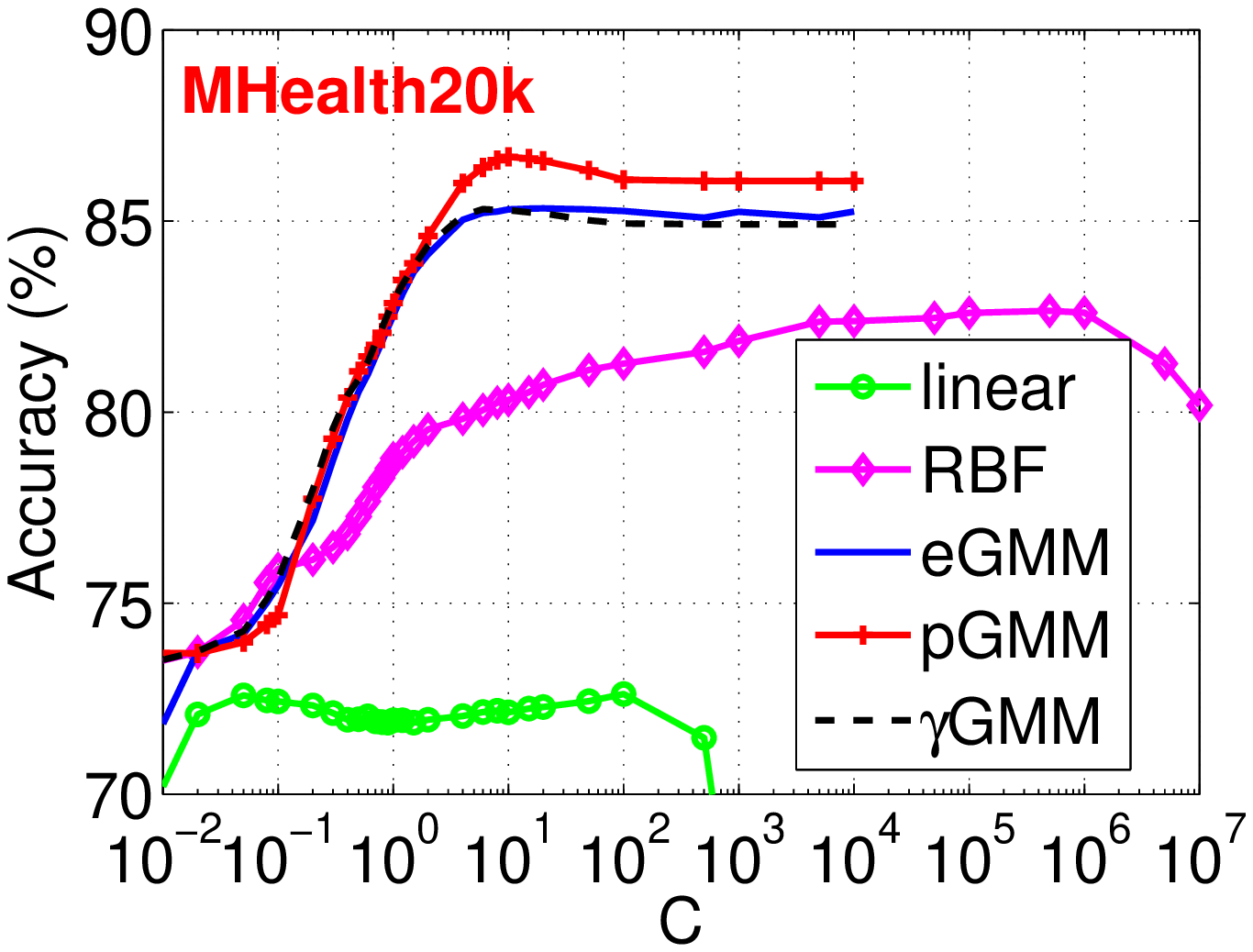}
}
\mbox{
\includegraphics[width=1.7in]{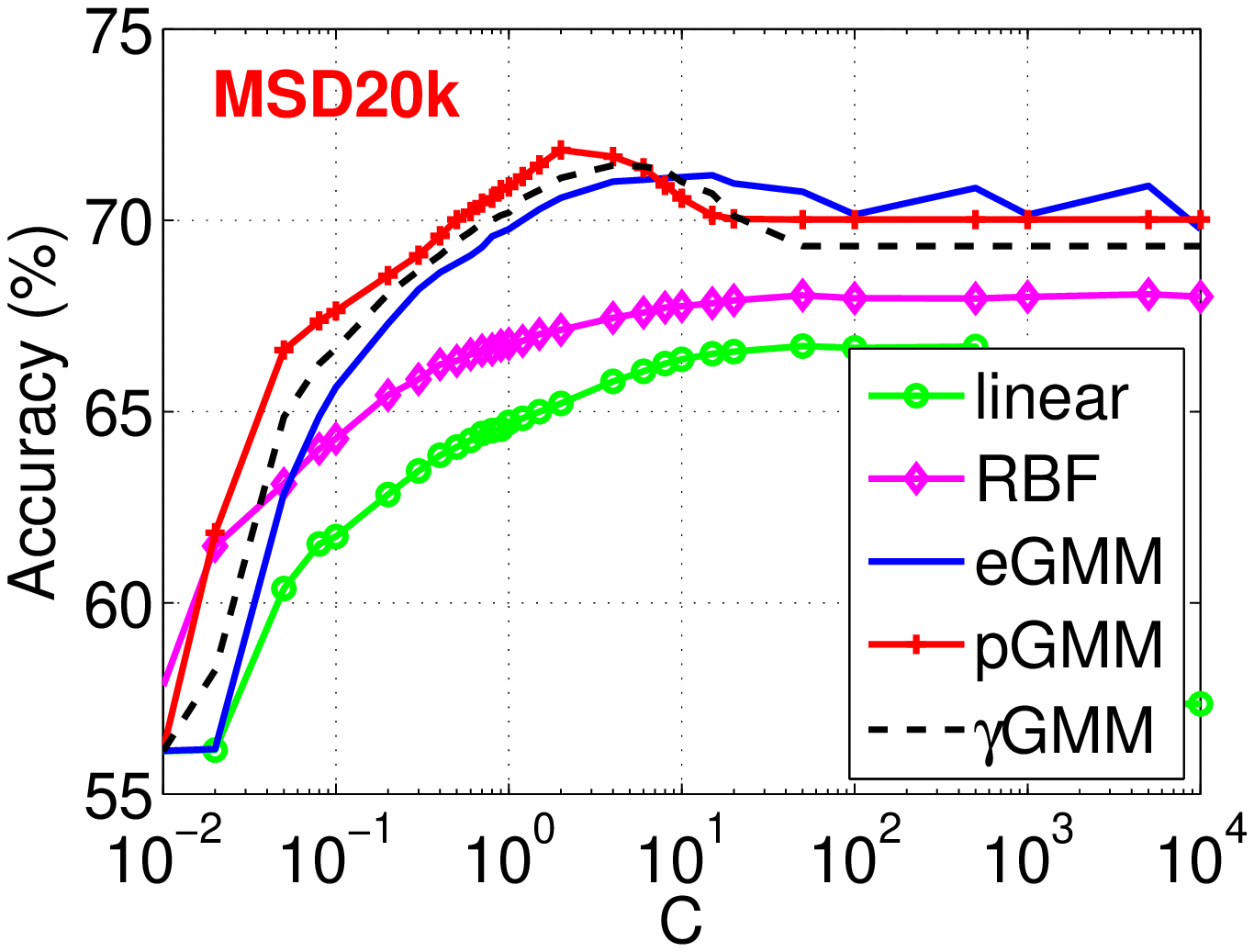}\hspace{-0in}
\includegraphics[width=1.7in]{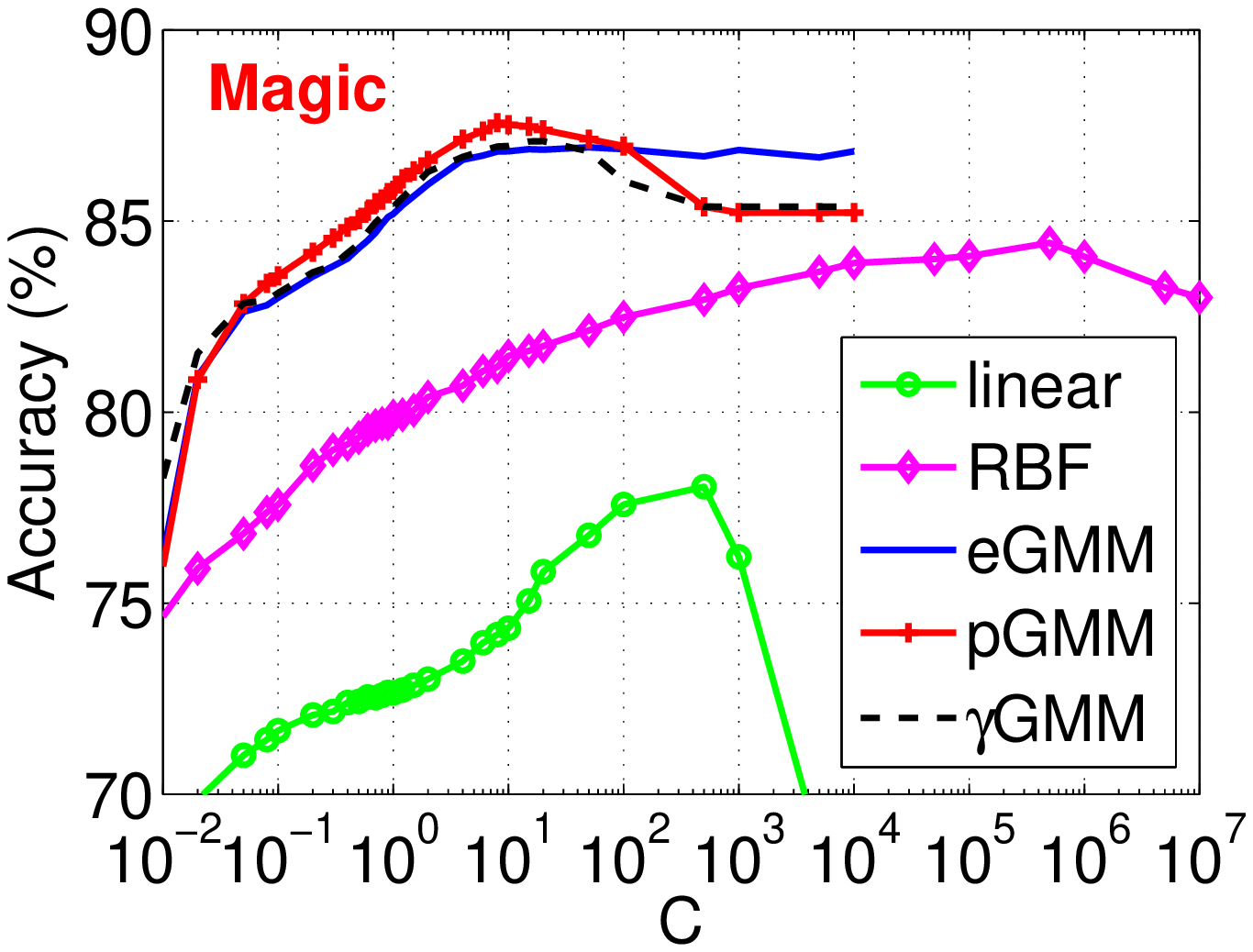}\hspace{-0in}
\includegraphics[width=1.7in]{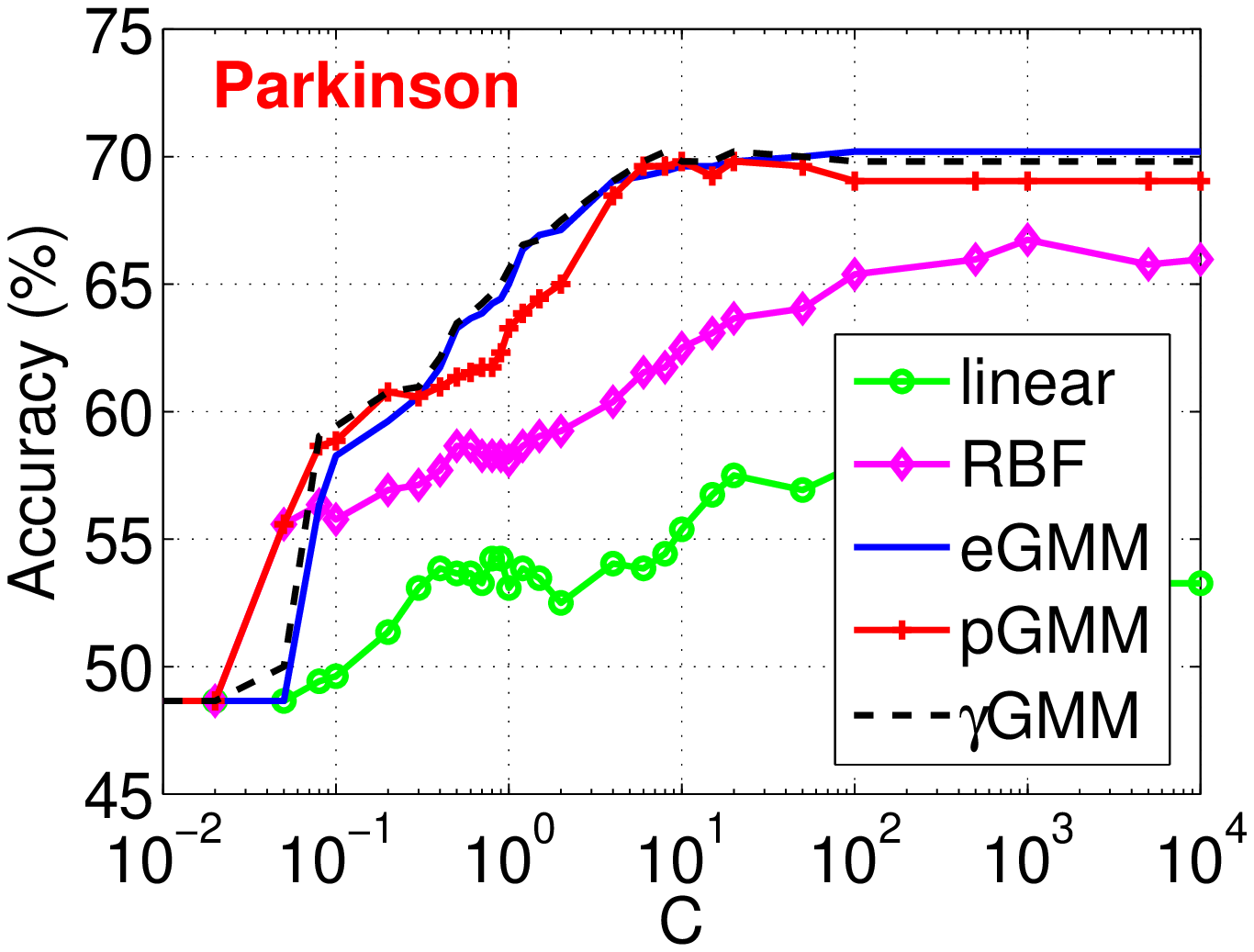}\hspace{-0in}
\includegraphics[width=1.7in]{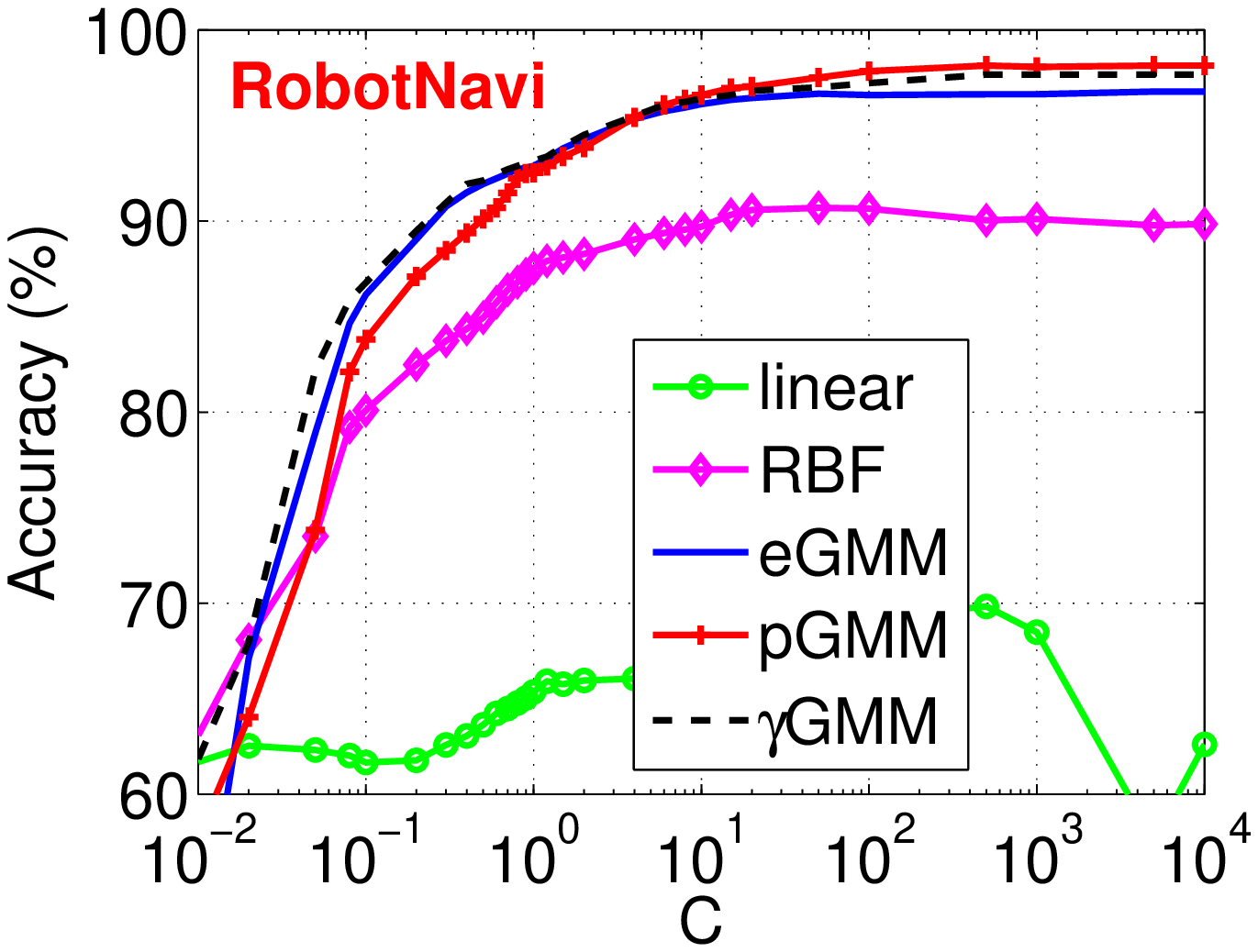}
}

\mbox{
\includegraphics[width=1.7in]{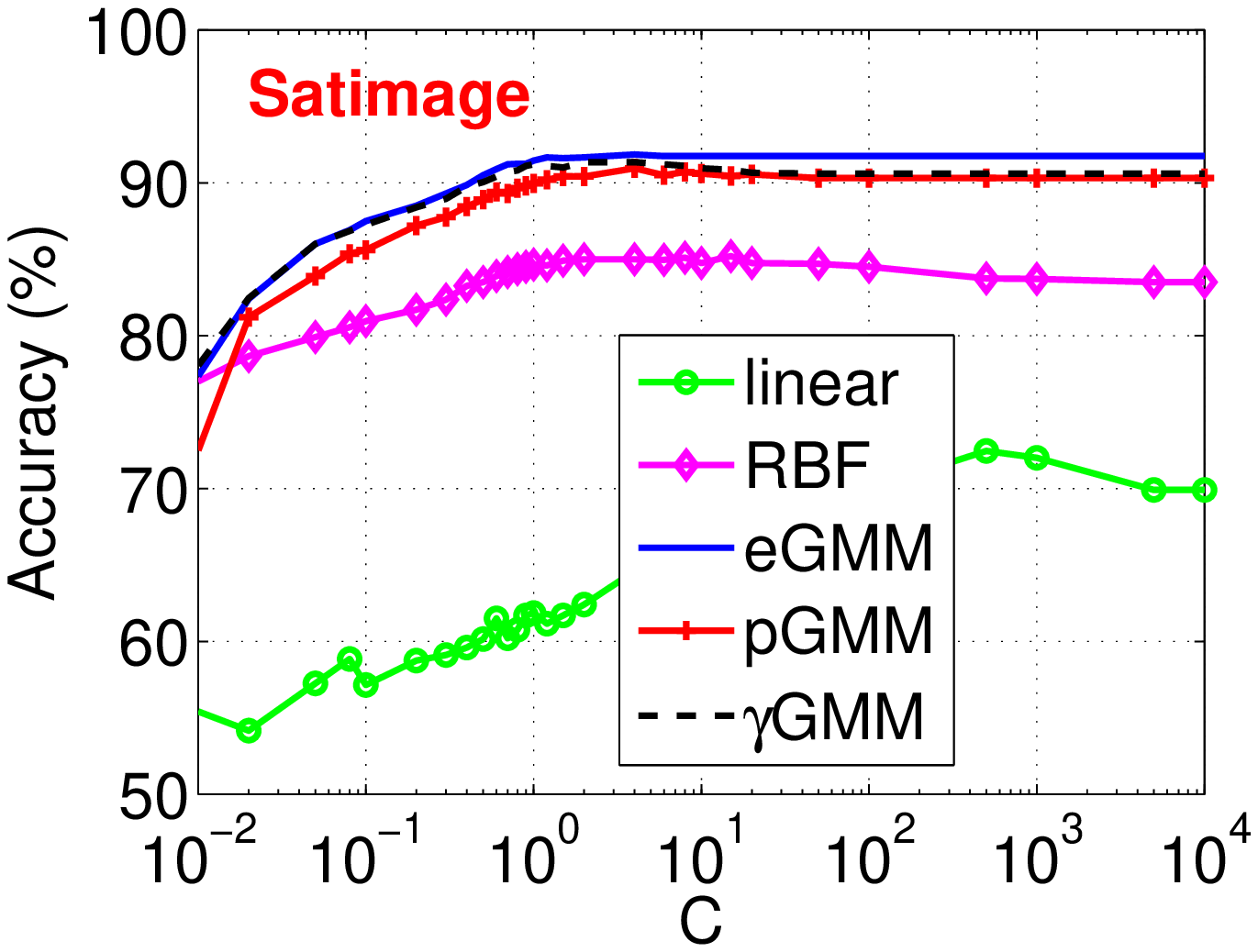}\hspace{-0in}
\includegraphics[width=1.7in]{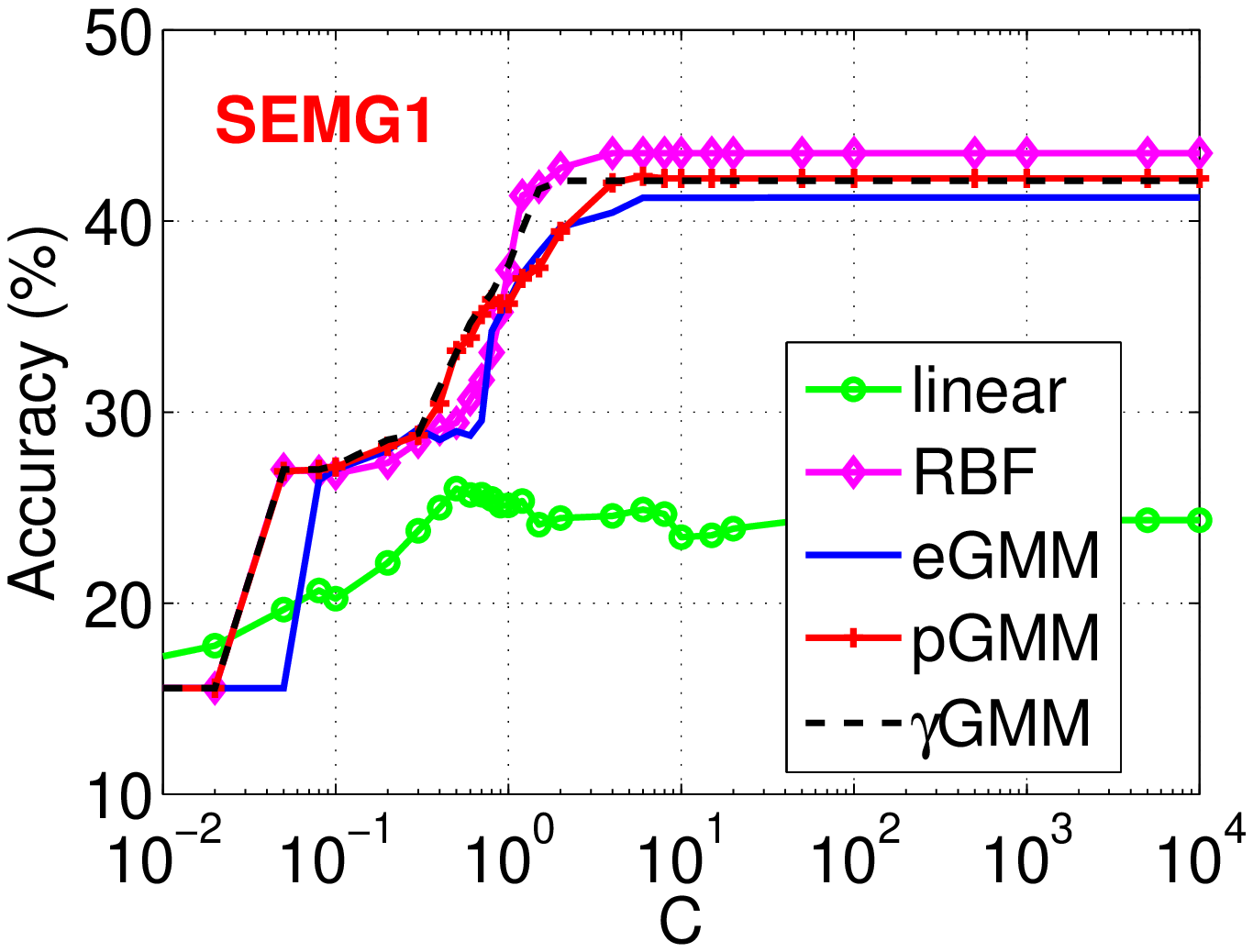}\hspace{-0in}
\includegraphics[width=1.7in]{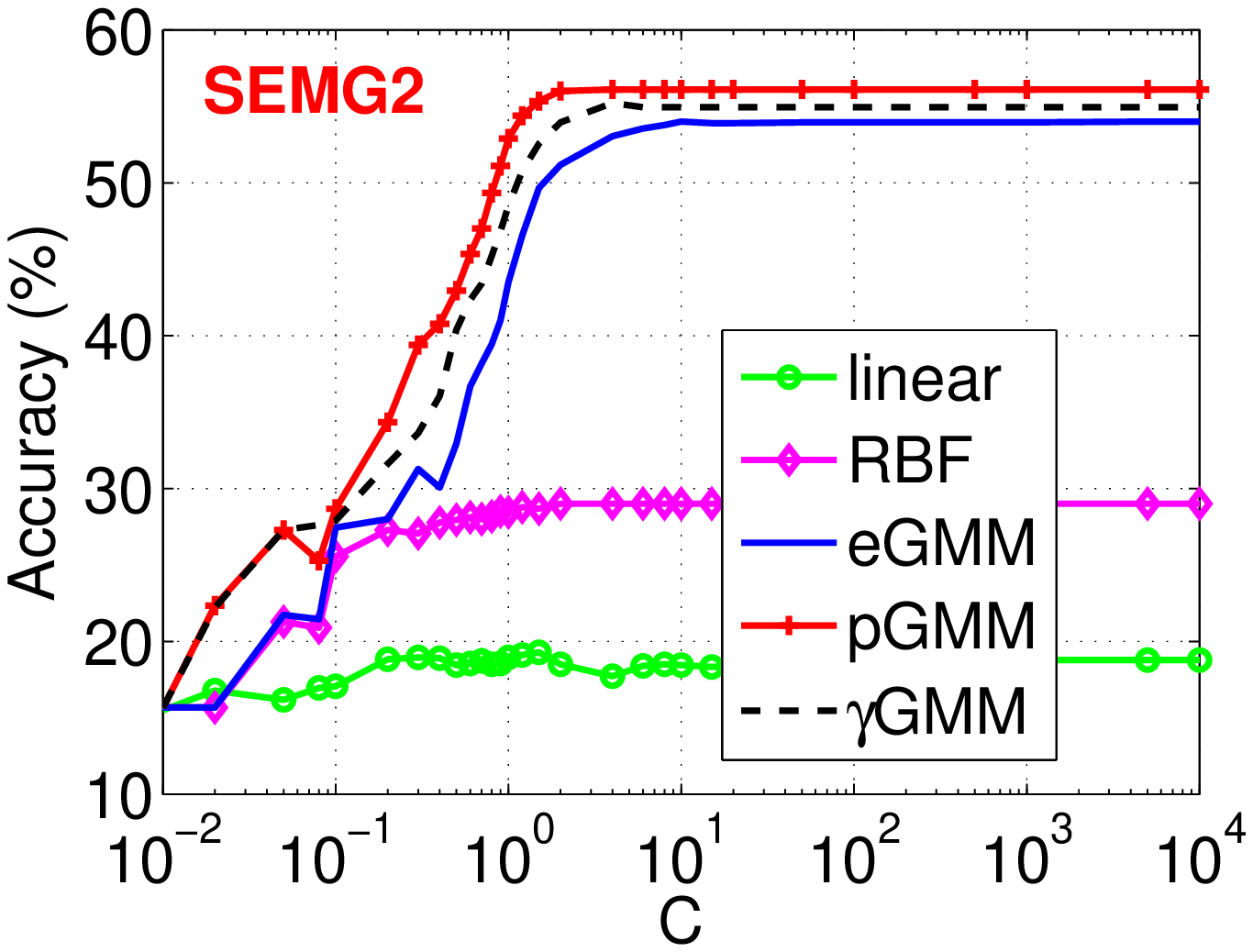}\hspace{-0in}
\includegraphics[width=1.7in]{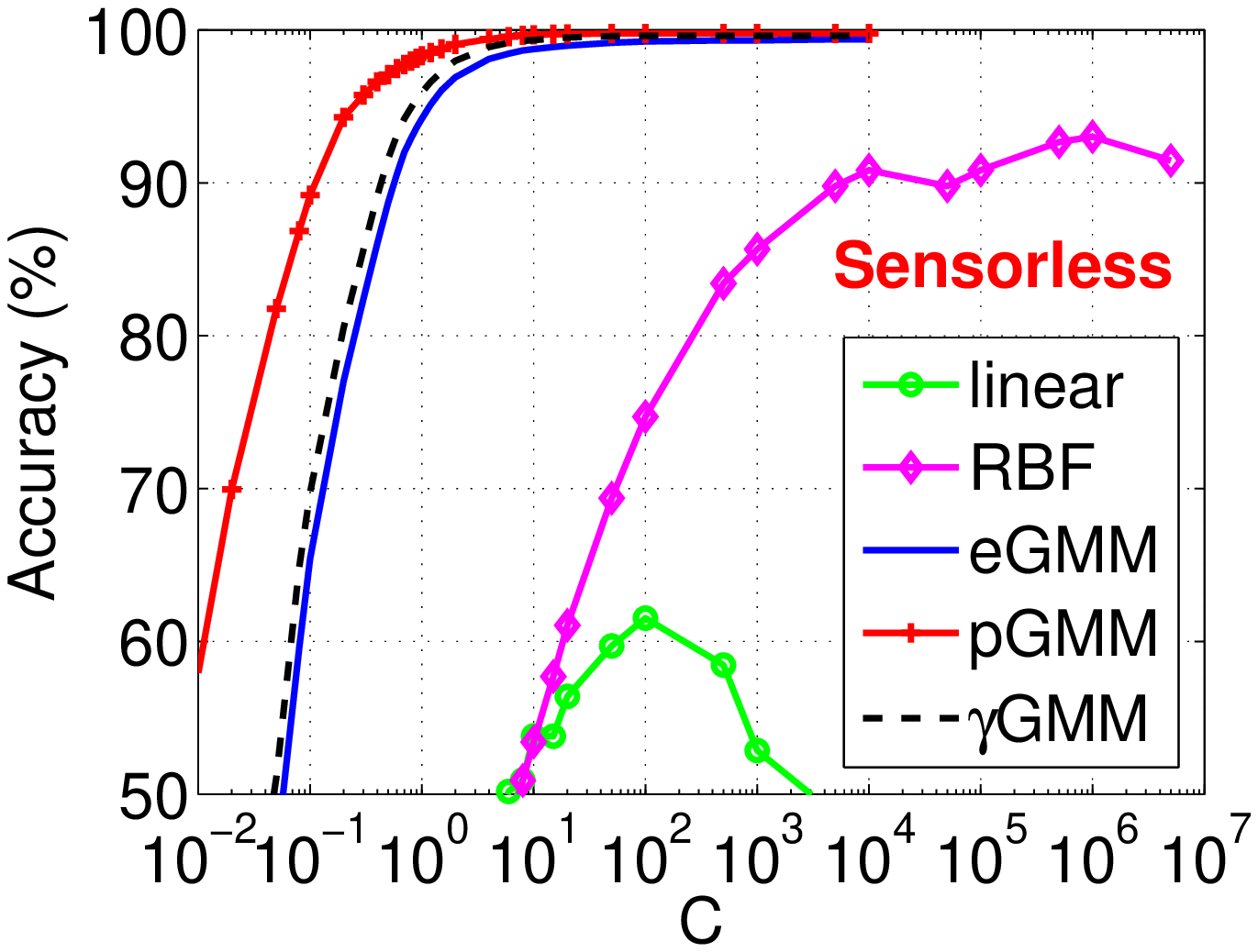}
}

\mbox{
\includegraphics[width=1.7in]{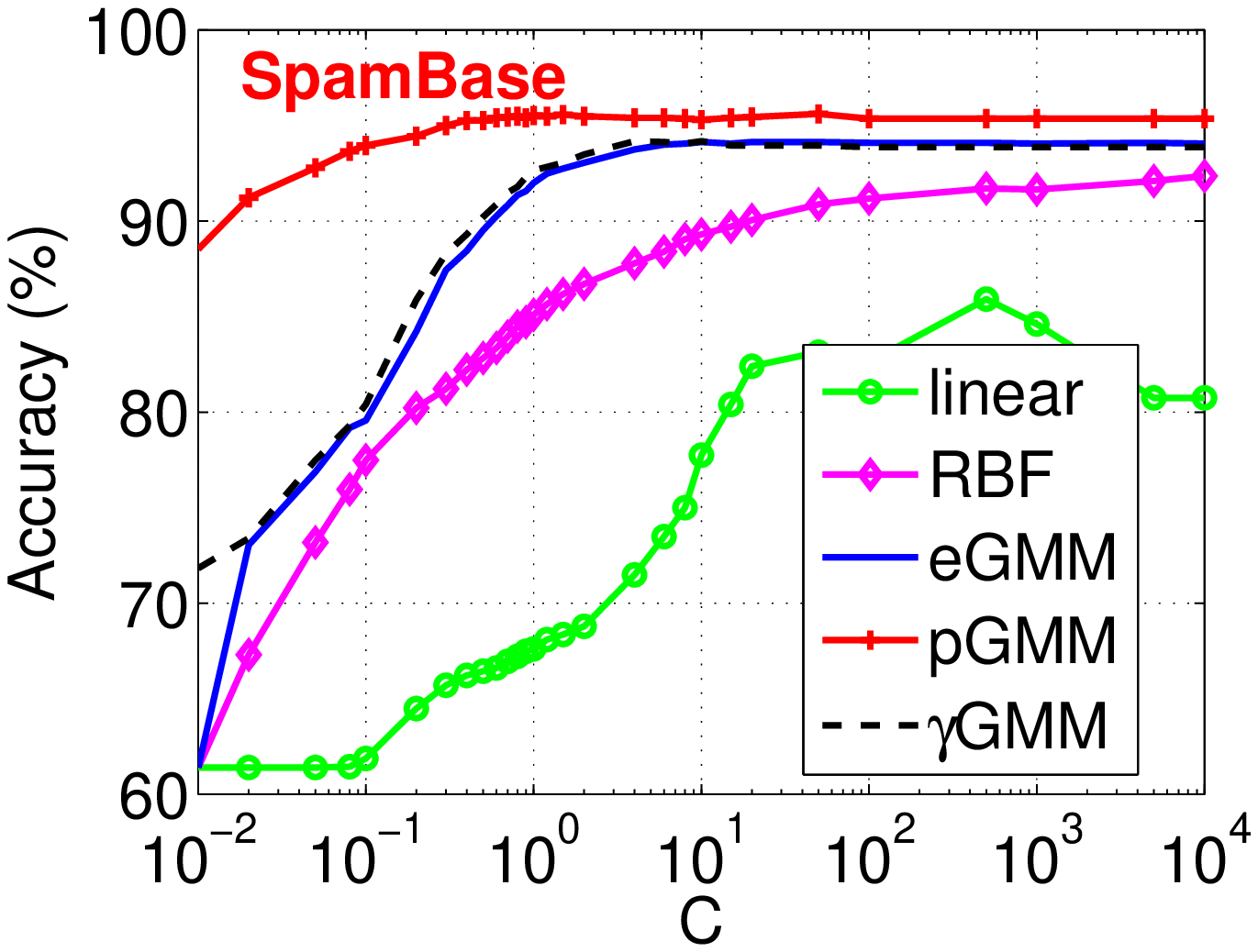}\hspace{-0in}
\includegraphics[width=1.7in]{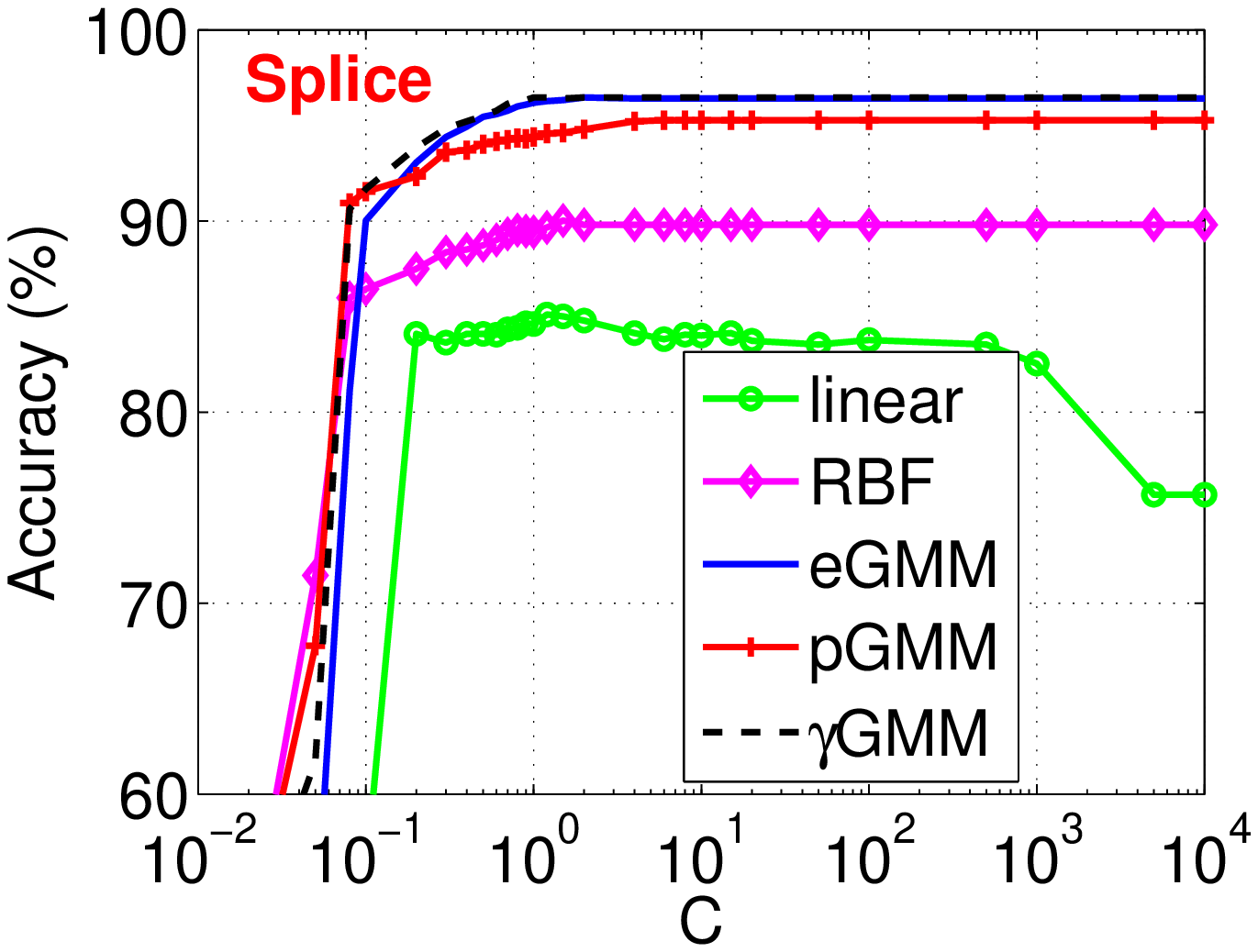}\hspace{-0in}
\includegraphics[width=1.7in]{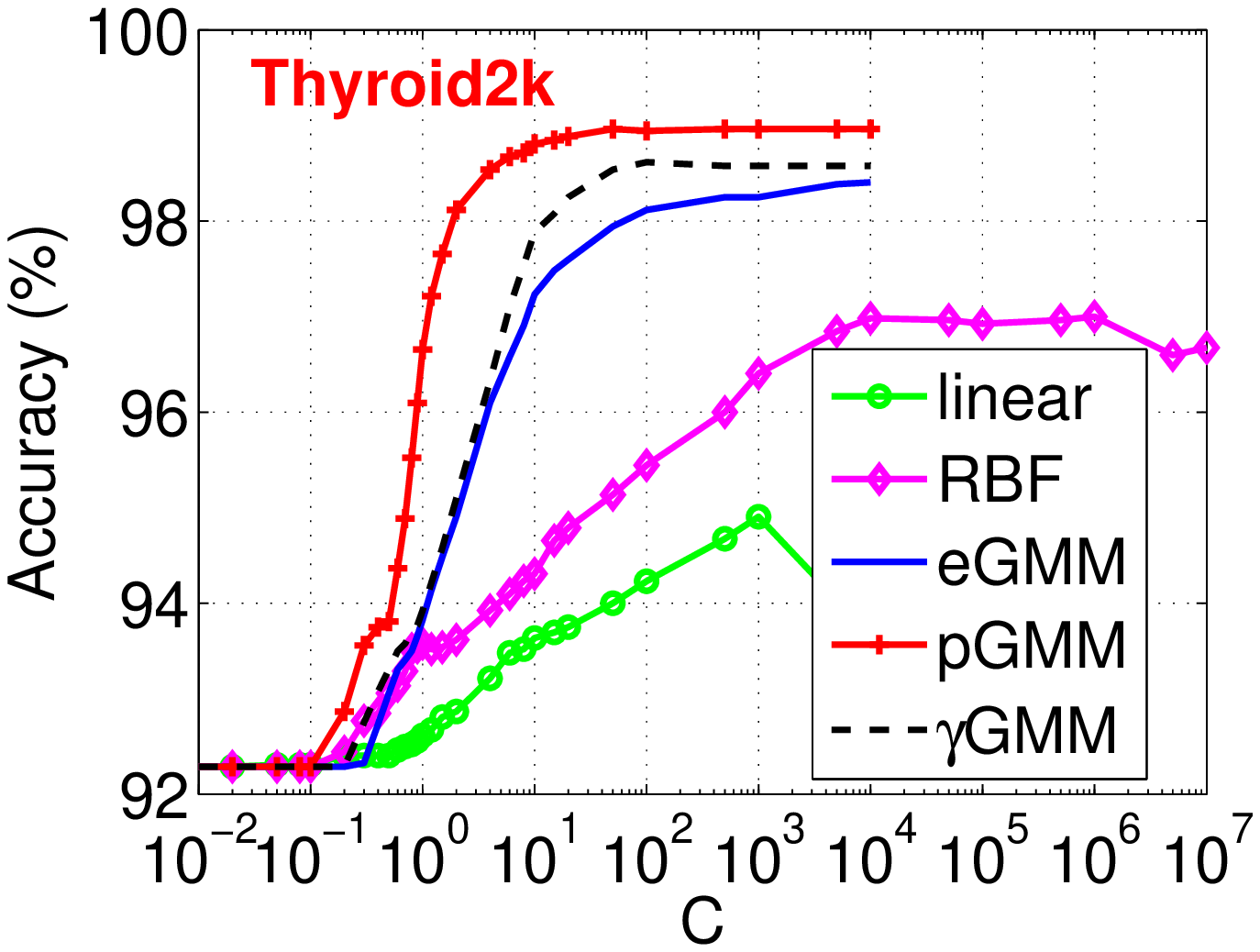}\hspace{-0in}
\includegraphics[width=1.7in]{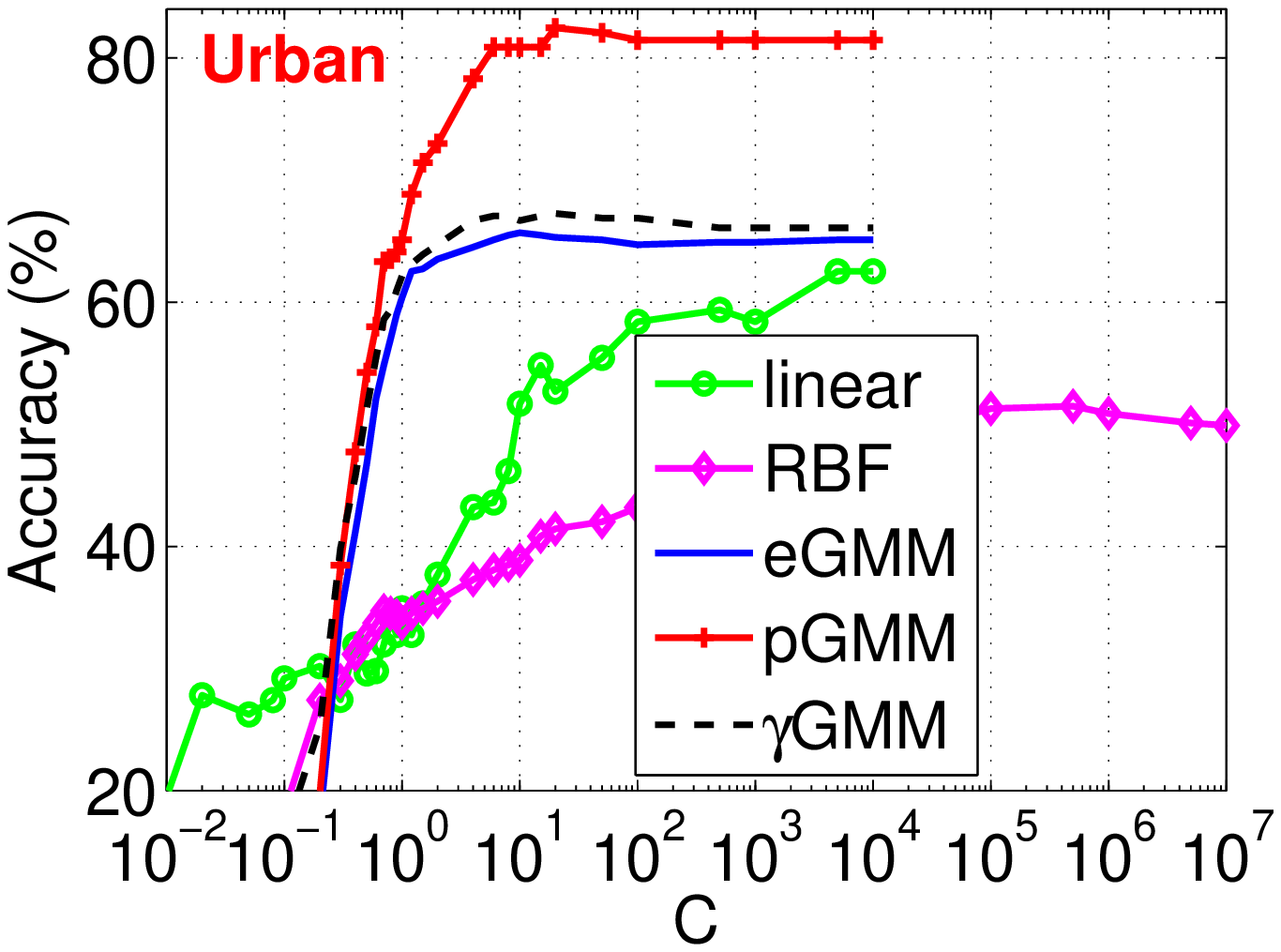}
}

\mbox{
\includegraphics[width=1.7in]{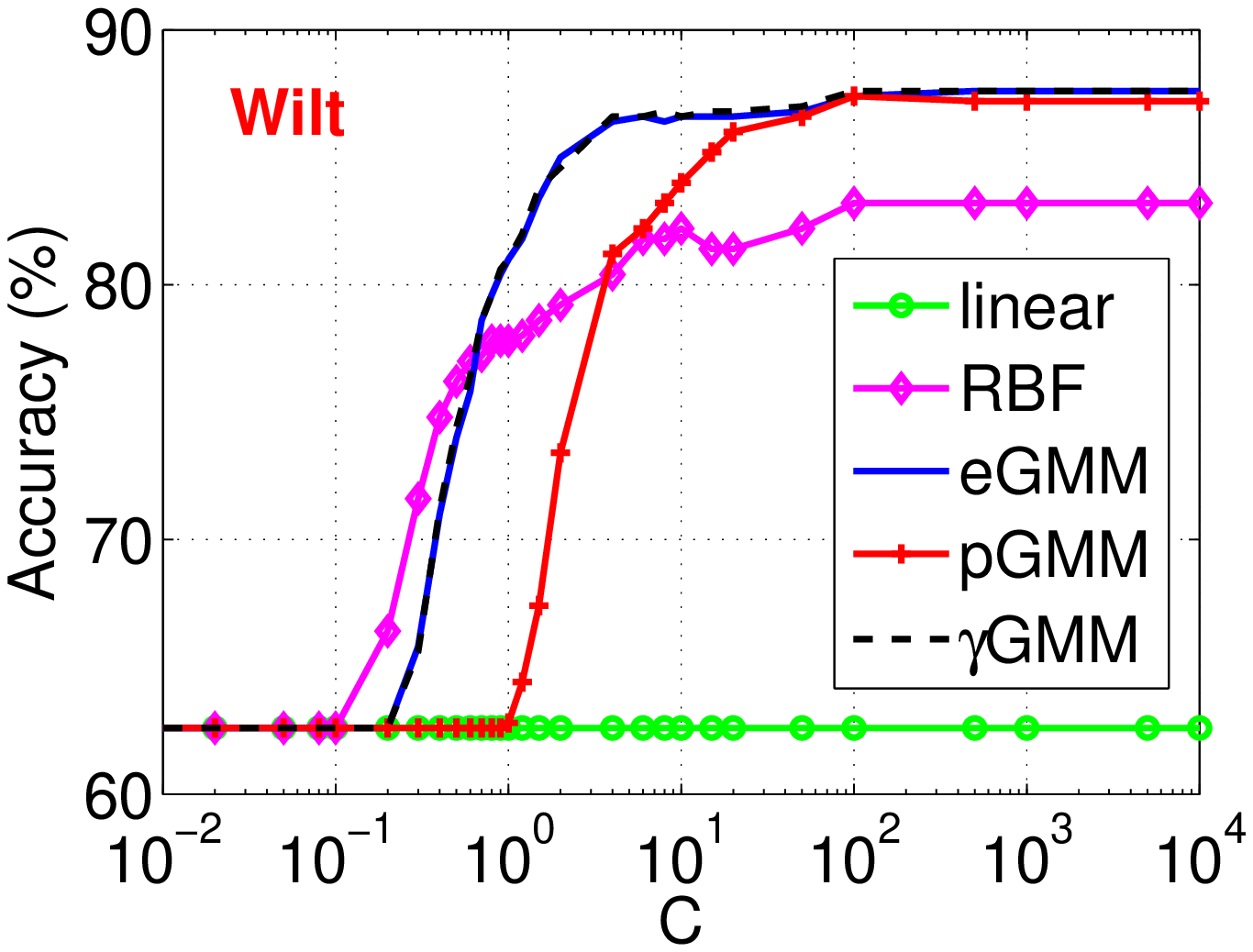}\hspace{-0in}
\includegraphics[width=1.7in]{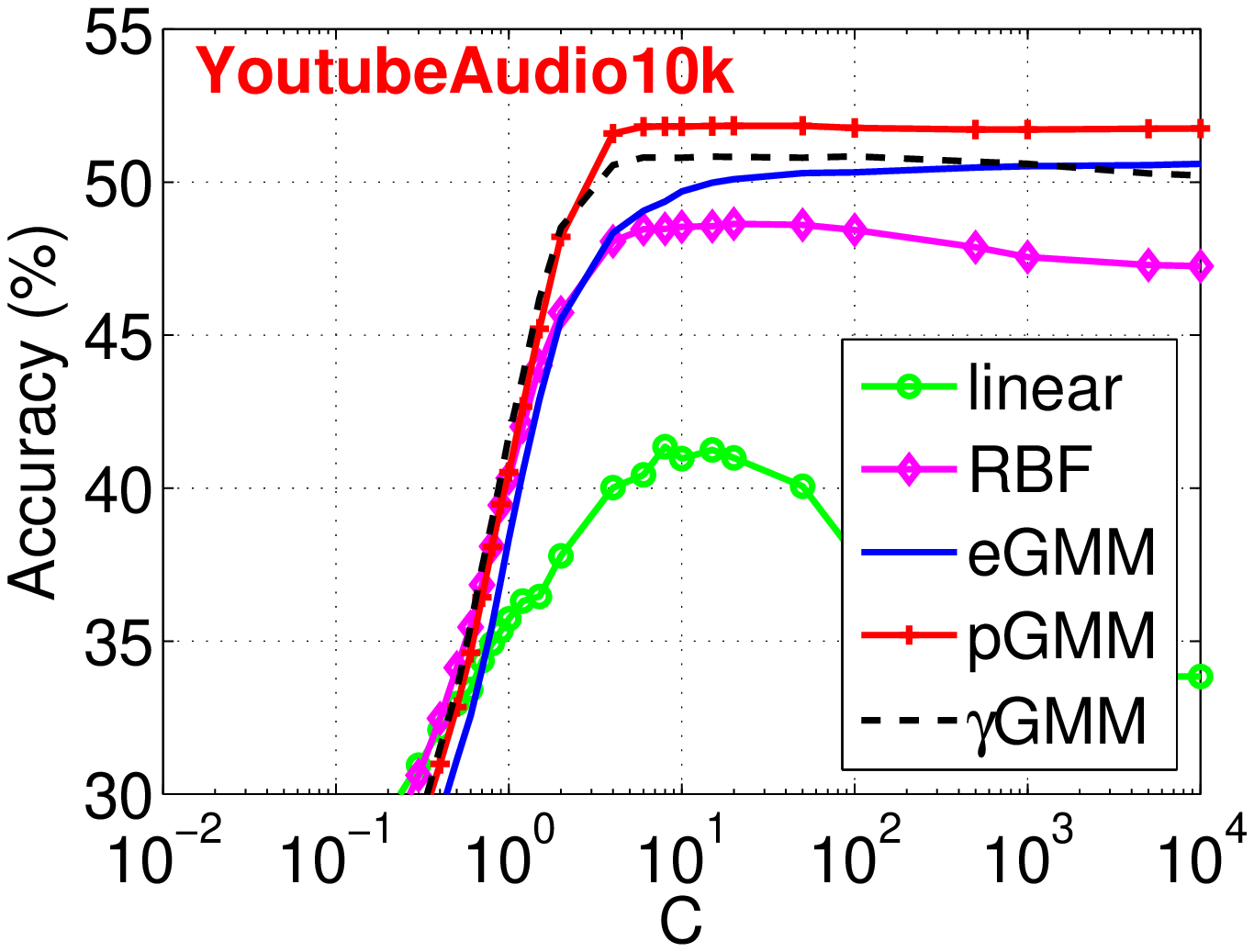}\hspace{-0in}
\includegraphics[width=1.7in]{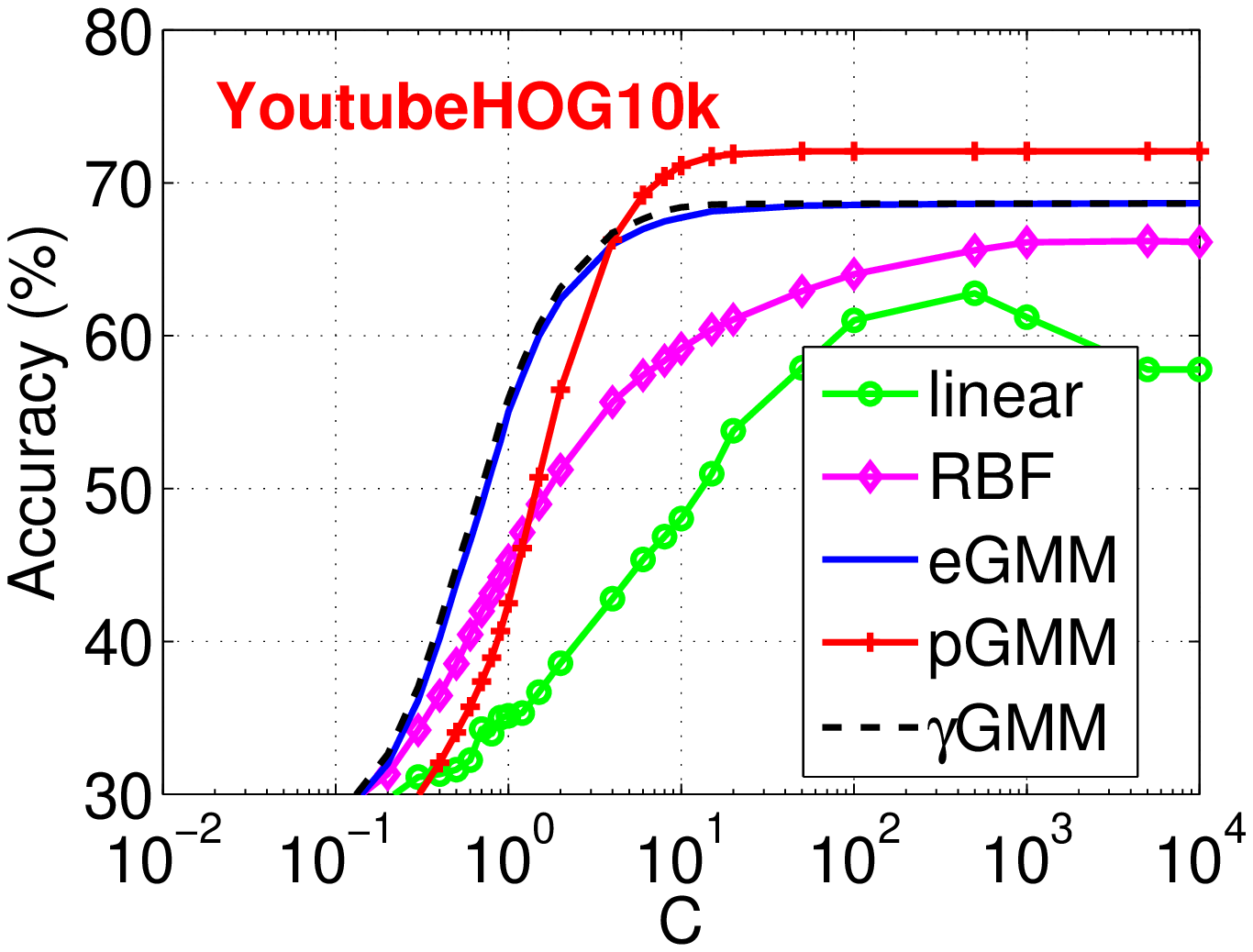}\hspace{-0in}
\includegraphics[width=1.7in]{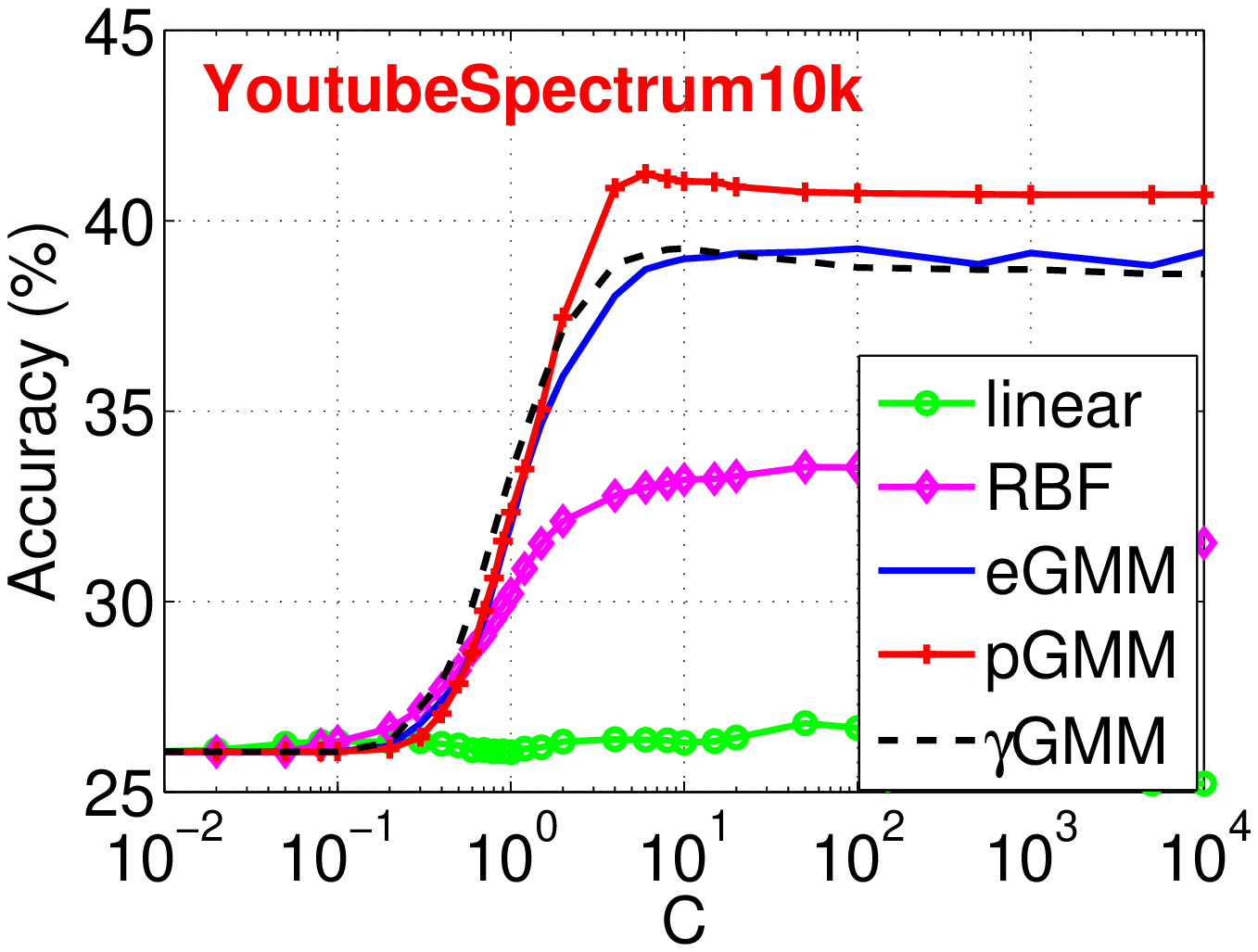}
}
\mbox{
\includegraphics[width=1.7in]{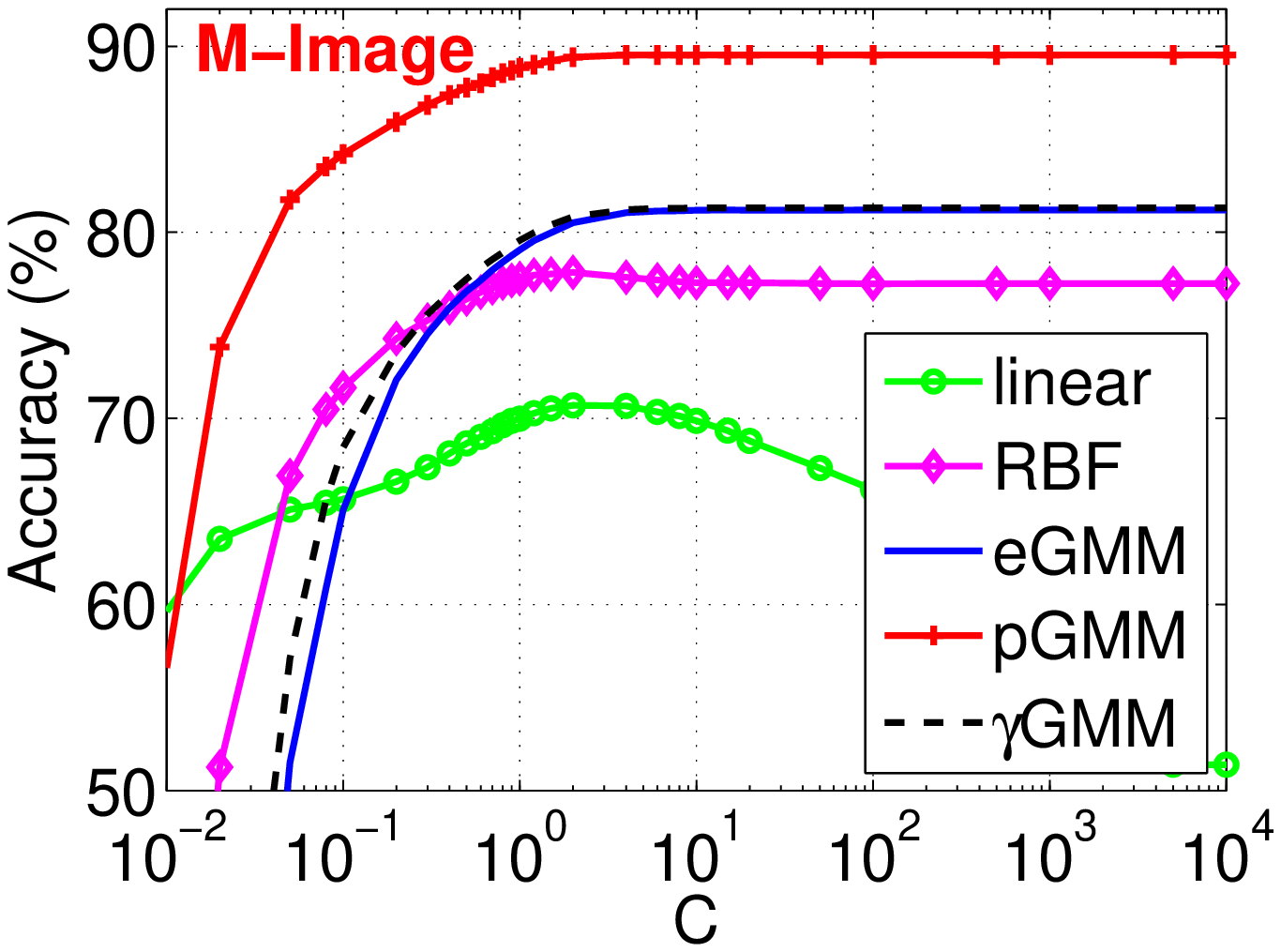}\hspace{-0in}
\includegraphics[width=1.7in]{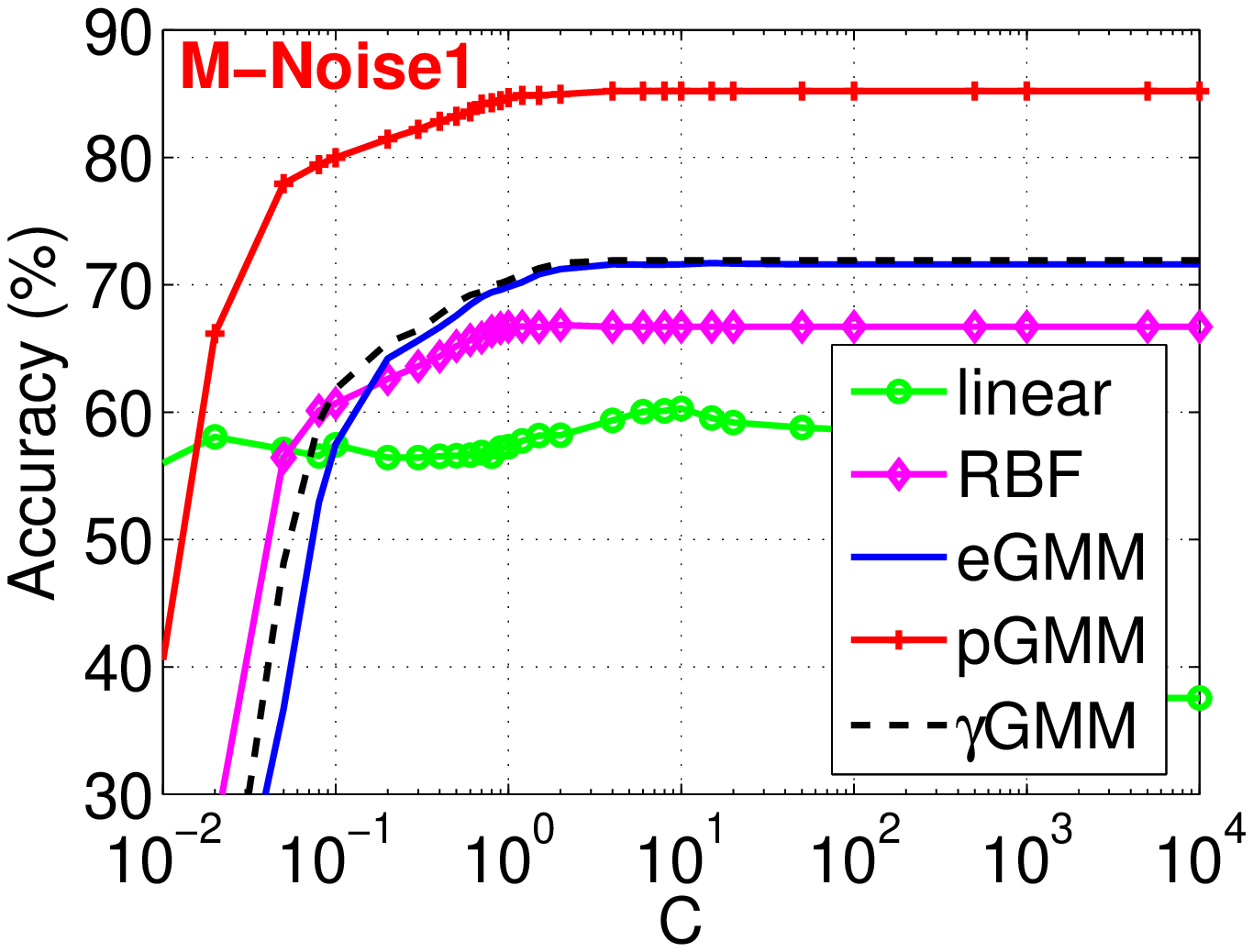}\hspace{-0in}
\includegraphics[width=1.7in]{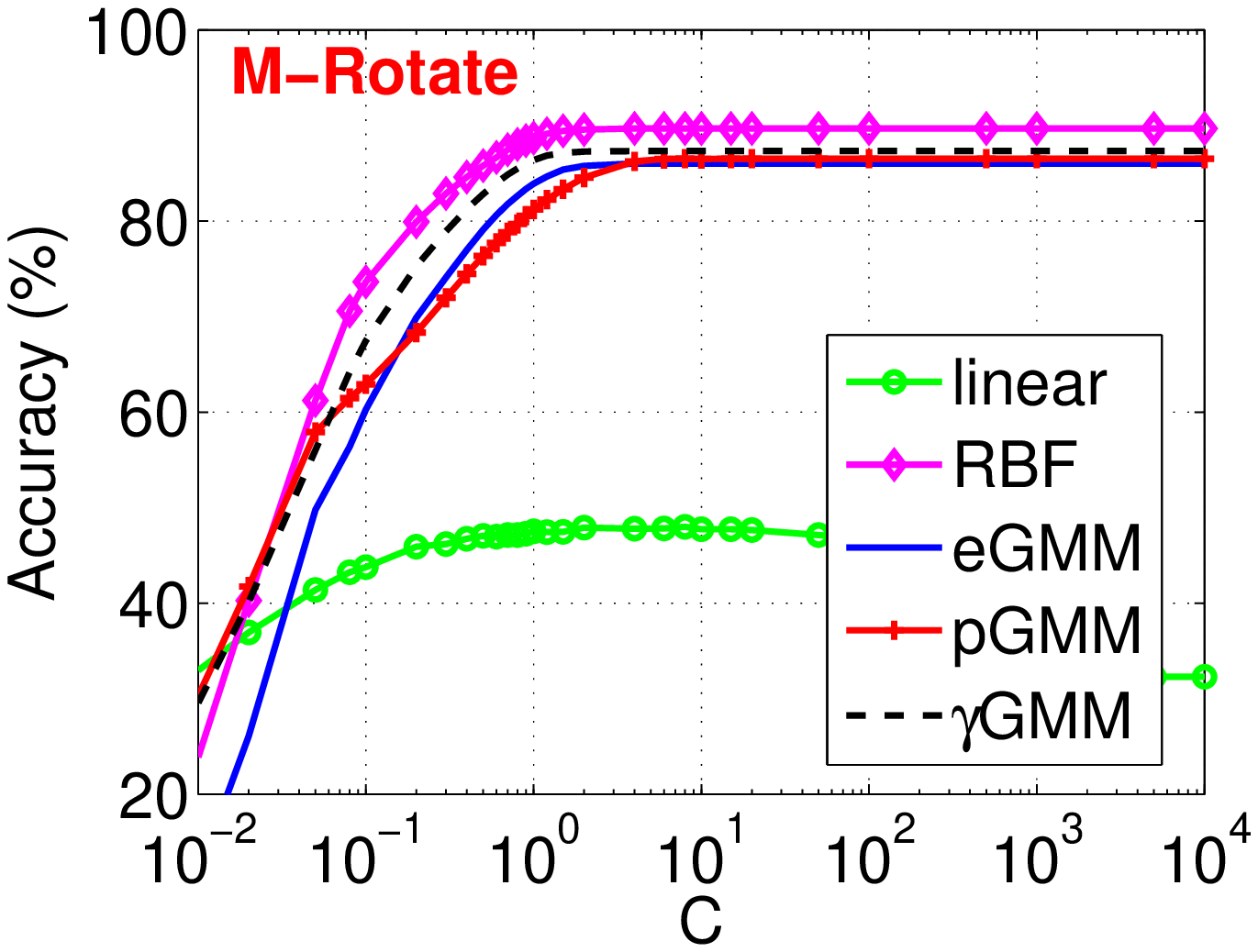}
\includegraphics[width=1.7in]{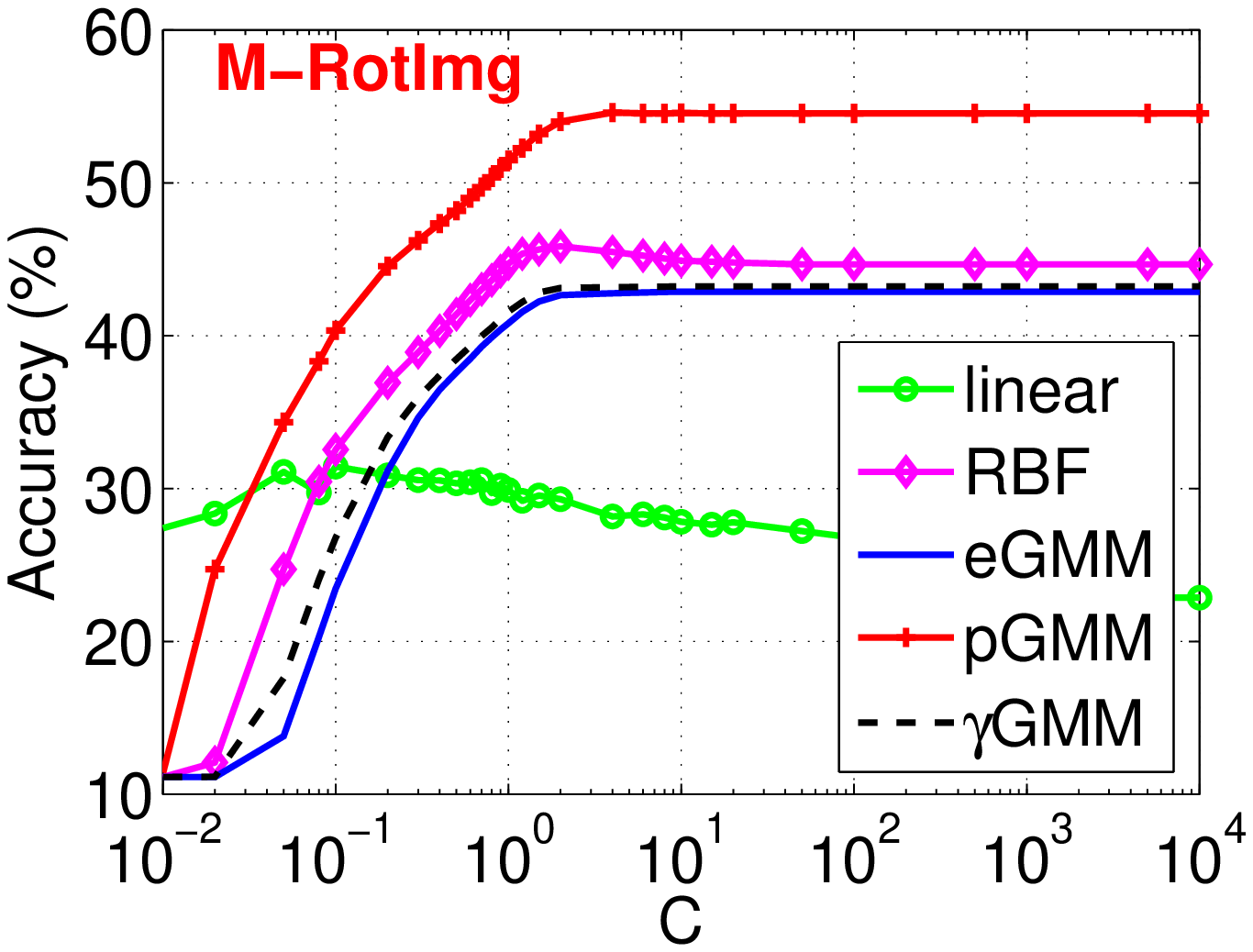}
}

\end{center}
\vspace{-0.2in}
\caption{Test classification accuracies of various kernels using LIBSVM pre-computed kernel functionality. The results are presented  with respect to $C$, which is the $l_2$-regularized kernel SVM parameter. For RBF, $e$GMM, $p$GMM, and $\gamma$GMM, at each $C$, we report the best test accuracies from a wide range of kernel parameter values.}\label{fig_SVM}
\end{figure*}

\newpage\clearpage

\section{Experimental Study on Kernel SVMs}\label{sec_kernel}

\subsection{Basic Kernels: $e$GMM, $p$GMM, $\gamma$GMM}

Table~\ref{tab_UCI} lists a large number of publicly available  datasets from the UCI repository plus the 11 datasets (the last 11 datasets whose names start with ``M-'') used by the deep learning literature~\cite{Proc:Larochelle_ICML07}. In this table, we report the kernel SVM test classification results for a variety of kernels: linear,  RBF, GMM, $e$GMM, $p$GMM, $\gamma$GMM.

In all the experiments, we adopt the $l_2$-regularization (with a regularization parameter $C$) and report the test classification accuracies at the best $C$ values in Table~\ref{tab_UCI}. More detailed results for a wide range of $C$ values are reported in Figures~\ref{fig_SVM}. To ensure repeatability, we use the LIBSVM pre-computed kernel functionality, at the significant cost of disk space. For the RBF kernel, we exhaustively experiment with 58 different values of $\lambda_e\in\{$0.001, 0.01, 0.1:0.1:2, 2.5, 3:1:20 25:5:50, 60:10:100, 120, 150, 200, 300, 500, 1000$\}$. Basically, Table~\ref{tab_UCI}   reports the best results among all $C$  and $\lambda_e$ values in our experiments. Here, 3:1:20 is the matlab notation, meaning that the iterations stat at 3 and terminate at 20, at a space of 1.

\vspace{0.08in}

For the $e$GMM kernel, we experiment with the same set of (58) $\lambda_e$ values as for the RBF kernel. For the $p$GMM kernel, however, because we have to materialize (store) a kernel matrix for each $\gamma$, disk space becomes a  serious concern. Therefore, for the $p$GMM kernel, we only search in the range of $p\in\{0.05, 0.1, 0.15, 0.2, 0.25, 0.3, 0.4,\\ 0.5, 0.6, 0.75, 1, 1.25, 1.5,  2, 5, 10, 15, 20, 25, 30:10:100\}$. For the $\gamma$GMM kernel, we experiment with $\gamma\in\{0.05, 0.1:0.1:3, 3.5:0.5:10, 11:1:20, 30:10:100\}$.

\begin{figure}[h!]
\begin{center}

\mbox{
\includegraphics[width=1.8in]{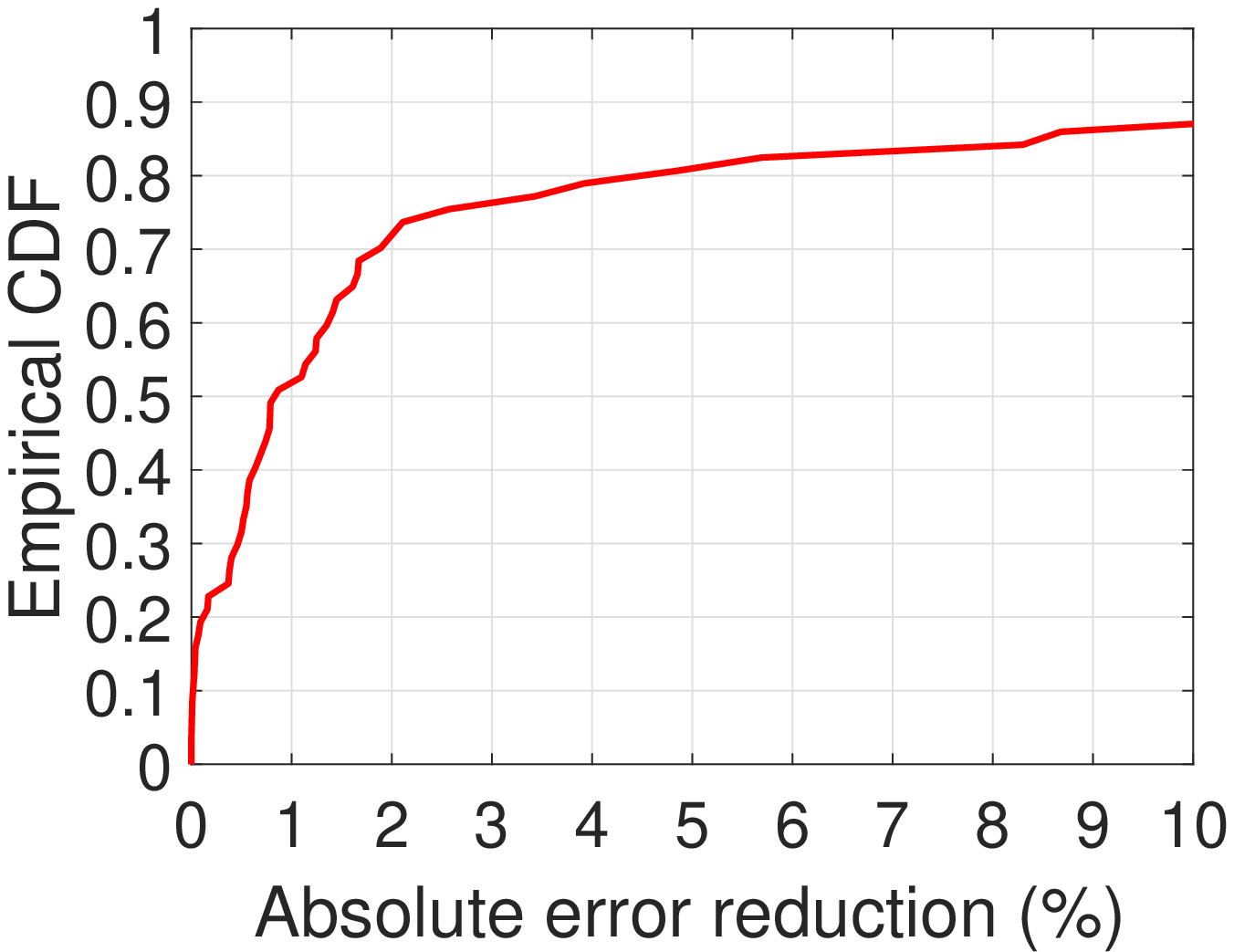}\hspace{-0.14in}
\includegraphics[width=1.8in]{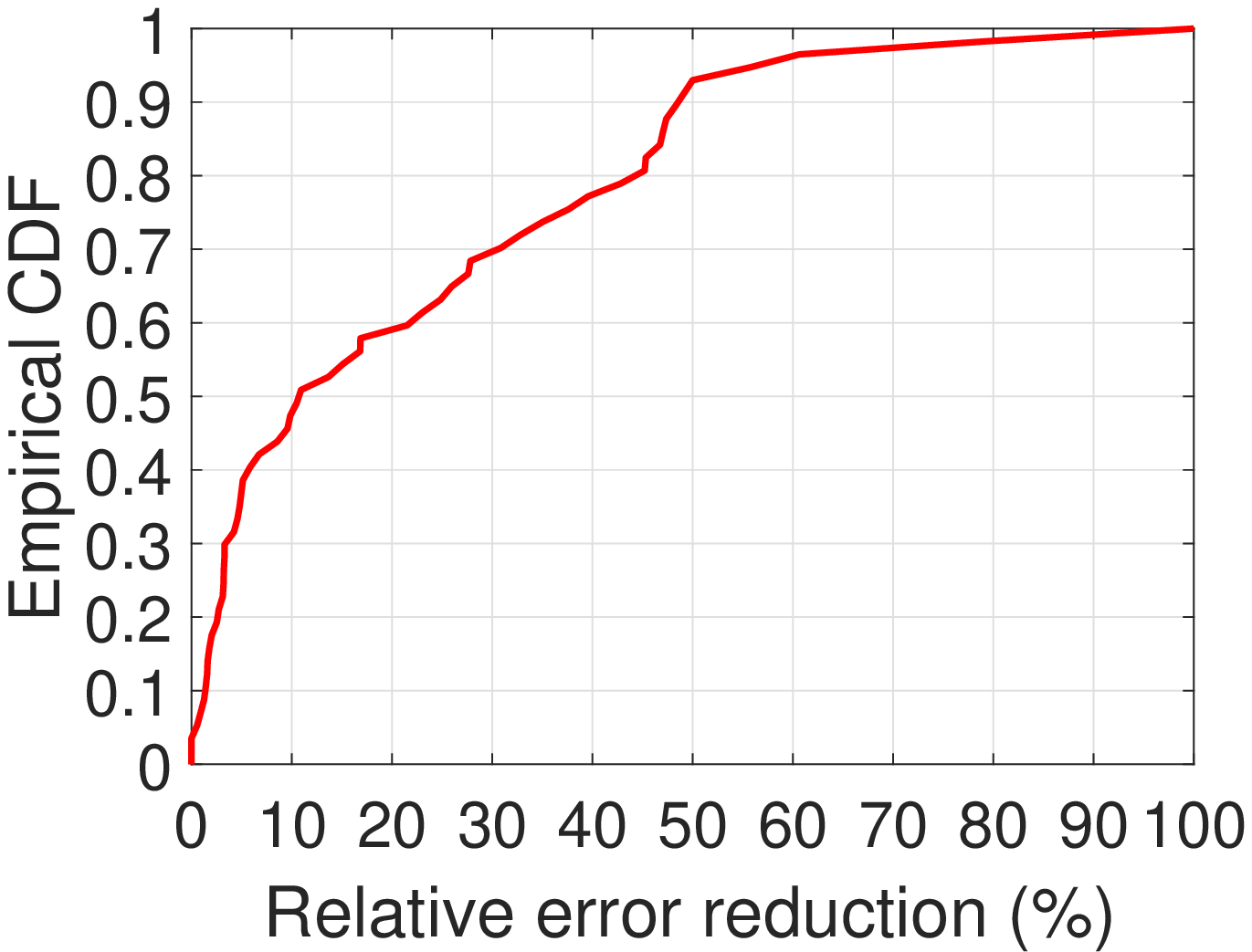}
}
\end{center}
\vspace{-0.15in}
\caption{Empirical CDFs of the absolute (left panel)  and relative (right panel) error reductions (\%) obtained by using one of the $e$GMM, $p$GMM, or $\gamma$GMM kernels, compared to using the original GMM kernel. There are in total 57 datasets.}\label{fig_Imp}\vspace{-0.05in}
\end{figure}

The classification results in Table~\ref{tab_UCI}  and Figures~\ref{fig_SVM} confirm that the $e$GMM,  $p$GMM, and $\gamma$GMM kernels typically improve the original GMM kernel. On a good fraction of datasets, the improvements can be very substantial. Figure~\ref{fig_Imp} quantifies the improvements by plotting the empirical CDF (cumulative distribution function) of the absolute (left panel) and relative (right panel) error reduction (in \%), obtained from using one of the $e$GMM, $p$GMM, or $\gamma$GMM kernels, compared to using the original GMM kernel. \vspace{-0.12in}

For example, the left panel of Figure~\ref{fig_Imp} says that, out of a total of 57 datasets (in Table~\ref{tab_UCI}), about $50\%$  of the datasets exhibit an improvement of $>1\%$ absolute error reduction, and about $20\%$ of the datasets exhibit an improvement of $>5\%$ absolute error reduction. The right panel  says that about $50\%$ of the datasets exhibit an improvement of $>10\%$ relative error reduction, and about $20\%$ of the data exhibit an improvement of $>45\%$ relative error reduction.

\subsection{$ep$GMM and $p\gamma$GMM}

Combining the three basic tunable GMM kernels produces four more tunable GMM kernels. Naturally we expect that, in many datasets, adding more tuning parameters would further improve the accuracies. In Table~\ref{tab_MNIST}, we report the kernel SVM experimental results on the 11 datasets used by the deep learning literature~\cite{Proc:Larochelle_ICML07}, for the $ep$GMM and $p\gamma$GMM kernels. Since there are two parameters, the results are obtained by a two-dim grid search.

The results in Table~\ref{tab_MNIST} illustrate that the additional improvements can be also quite substantial. For example, on the M-RotImg dataset, the accuracy of the original GMM kernel is $40.98\%$, and the accuracy of the $p$GMM kernel is $54.58\%$. However, the accuracy of $p\gamma$GMM kernel becomes $59.92\%$.

\begin{table*}[t]
\caption{We add the results (test classification accuracies) for the $ep$GMM and $p\gamma$GMM kernels  as the last two columns, for the 11 datasets used in the deep learning literature~\cite{Proc:Larochelle_ICML07} and later by~\cite{Proc:ABC_UAI10} for comparing tree methods.\vspace{-0.1in}}
\begin{center}{
{\begin{tabular}{l r r r r c c c c c c c}
\hline \hline
Dataset  &\# train  &\# test  &\# dim   &linear  &RBF  &GMM &$e$GMM &$p$GMM &$\gamma$GMM &$ep$GMM&$p\gamma$GMM \\
\hline
M-Basic  &12000 &50000&784  & 89.98   &{\bf97.21}   &96.34  &96.47  & 96.40  &96.84 &{96.71} &97.00 \\
M-Image  &12000 &50000&784 & 70.71 &77.84  &{80.85} &81.20 &89.53 &81.32 &{89.96} &{\bf90.96}\\
M-Noise1 &10000 &4000 &784 &60.28   &66.83    &{71.38} &71.70   &85.20 &71.90 &{85.58} &{\bf87.13}  \\
M-Noise2 &10000 &4000 &784  & 62.05  & 69.15   &{72.43} &72.80   &85.40 &72.95 &{86.05} &{\bf87.53}\\
M-Noise3 &10000 &4000 &784 &65.15   &71.68   &{73.55} &74.70   &86.55  &74.83 &{87.10} &{\bf88.28} \\
M-Noise4 &10000 &4000 &784 & 68.38  &75.33   &{76.05} &76.80  &86.88 &77.03 &{87.43} &{\bf88.80}\\
M-Noise5 &10000 &4000 &784 &72.25   &78.70  &{79.03} &79.48   &87.33 &79.70 &{88.30}&{\bf89.18}  \\
M-Noise6 &10000 &4000 &784 &78.73   &{85.33} &84.23  &84.58  &88.15 &84.68 &{88.85} &{\bf89.78} \\
M-Rand &12000 &50000&784  & 78.90   &{85.39}   &84.22 &84.95   &89.09 &85.17 &{89.43} &{\bf90.63}\\
M-Rotate &12000 &50000&784   &47.99  &\textbf{89.68}  & 84.76 &86.02 &86.56 &87.33 &{88.36}  &{89.06} \\
M-RotImg &12000 &50000&784 &31.44  &{45.84}   & 40.98 &42.88   &54.58 &43.22 &{55.73} &{\bf59.92}\\
\hline\hline
\end{tabular}}
}
\end{center}\label{tab_MNIST}

\end{table*}

\vspace{0.08in}

We choose to show experiment on these 11 datasets because the prior work~\cite{Proc:ABC_UAI10} in 2010 already conducted a thorough empirical study of a series of tree \& boosting methods on the same datasets.

\vspace{-0.1in}

\subsection{Comparisons with Trees}

\vspace{-0.2in}

\begin{figure}[h!]
\begin{center}
\includegraphics[width=3in]{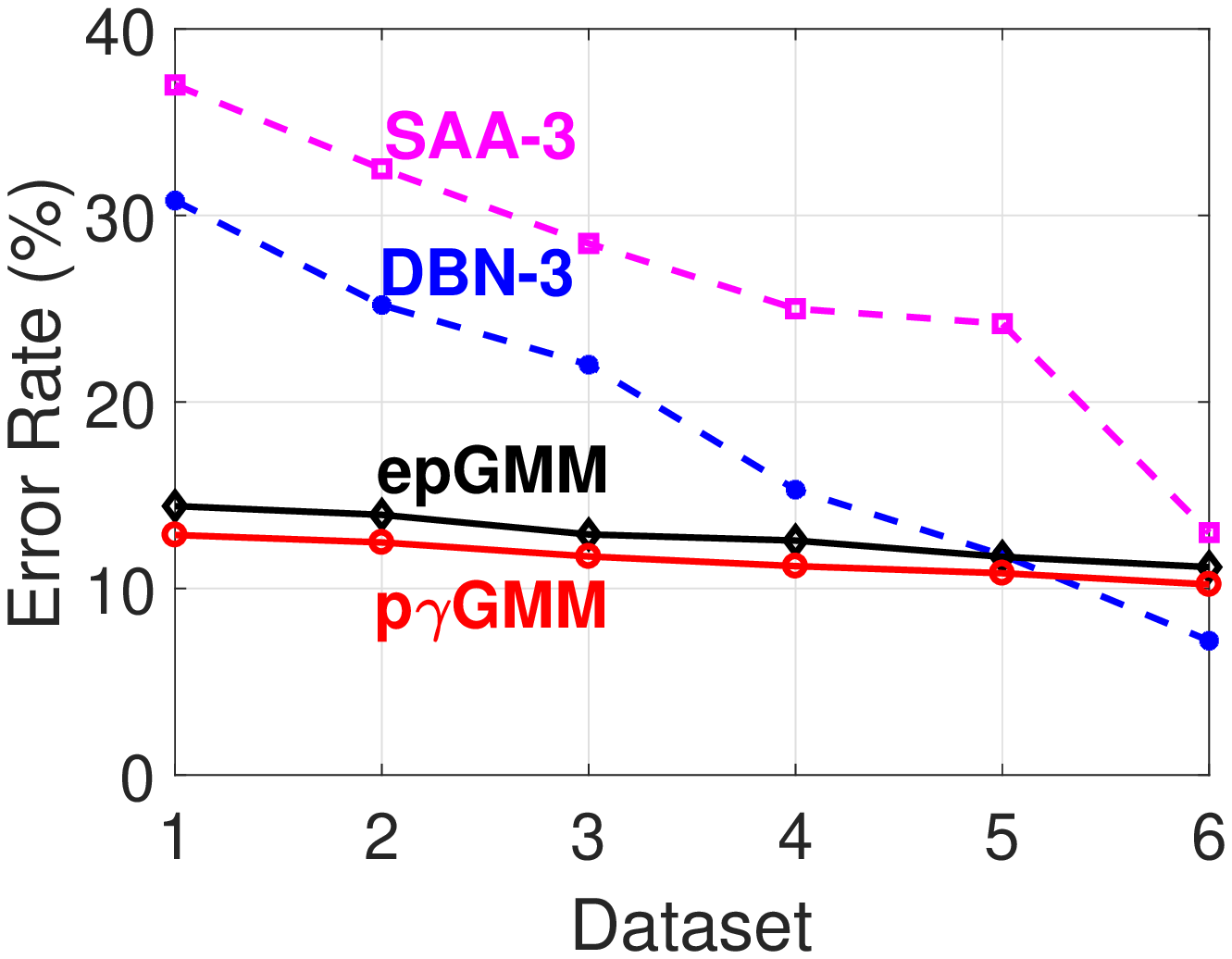}

\vspace{-0.25in}

\includegraphics[width=3in]{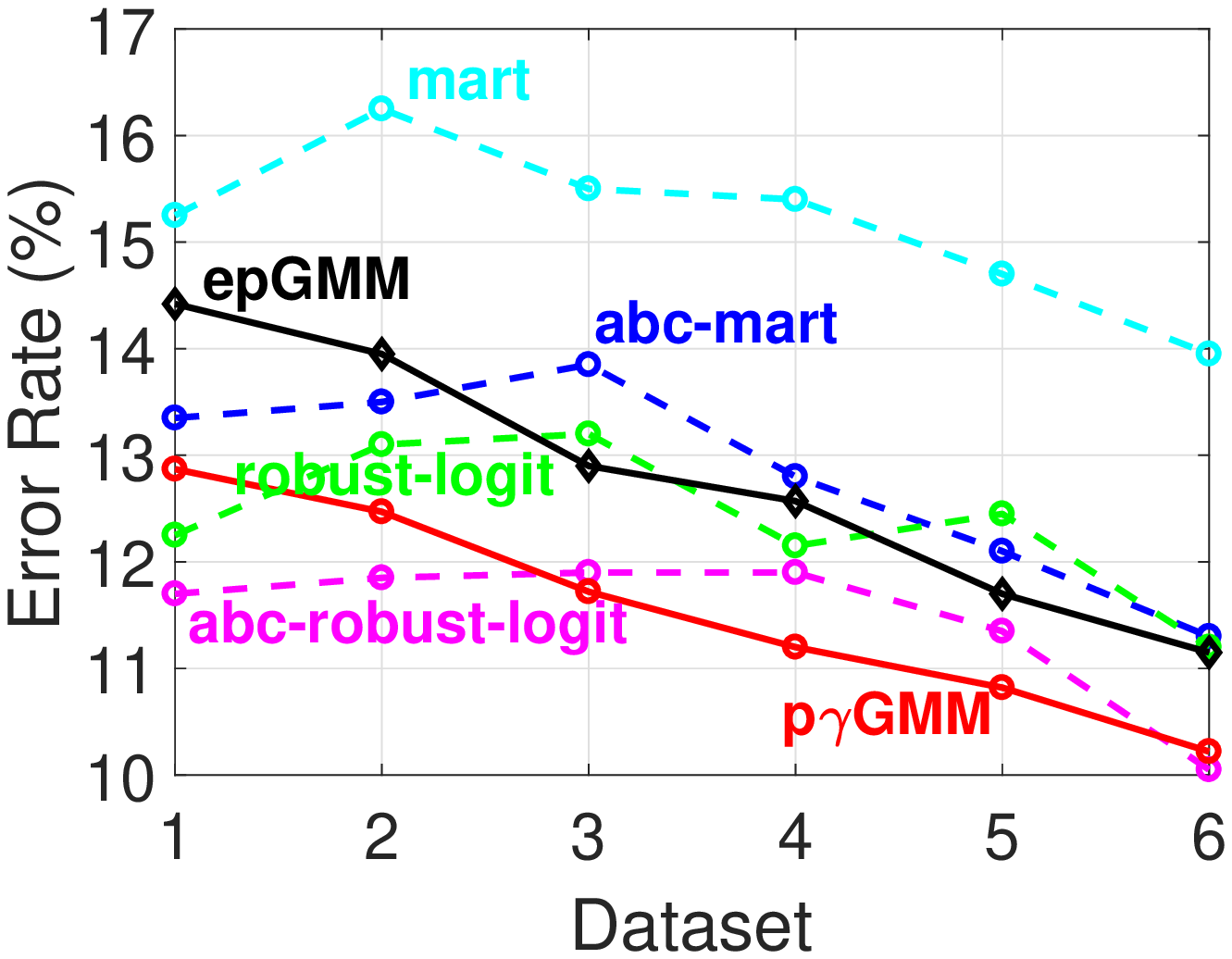}

\end{center}
\vspace{-0.15in}
\caption{Classification test error rates on M-Noise1, M-Noise2, ..., M-Noise6 datasets, for evaluating $ep$GMM and $p\gamma$ GMM kernels.  The upper panel includes the results of two deep learning algorithms (SAA-3 and DBN-3) as reported in~\cite{Proc:Larochelle_ICML07}. The bottom panel compares the $ep$GMM and $p\gamma$GMM kernels with four boosted tree methods as reported in~\cite{Proc:ABC_UAI10}.}\label{fig_Noise6}\vspace{-0.02in}
\end{figure}

Following~\cite{Proc:Larochelle_ICML07,Proc:ABC_UAI10}, we report the results on these 11 datasets in terms of the test error rates instead of accuracies, in Figure~\ref{fig_Noise6} (for M-Noise1, M-Noise2, M-Noise3, M-Noise4, M-Noise5, M-Noise6) and Table~\ref{tab_deep} (for M-Basic, M-Rotate, M-Image, M-Rand, M-RotImg).

The results presented in Figure~\ref{fig_Noise6} and Table~\ref{tab_deep} are  quite exciting, because, at this point, we merely use kernel SVM with single kernels. The performances of  tunable GMM kernels are already  comparable to four boosting \& tree methods:  mart, abc-mart, robust logitboost, and abc-robust-logitboost, whose training procedures are time-consuming with large model sizes (up to 10000 boosting iterations). It is  reasonable to expect that additional improvements  might be achieved in near future.\\

The ``mart'' tree algorithm~\cite{Article:Friedman_AS01} has been  popular in industry practice, especially in search. At each boosting step, it uses the first derivative of the logistic loss function as the residual response to fit regression trees, to achieve excellent robustness and fairly good accuracy. The earlier work on ``logitboost''~\cite{Article:FHT_AS00} were believed to exhibit numerical issues (which in part motivated the development of mart). It turns out that the numerical issue does not actually exist after~\cite{Proc:ABC_UAI10} derived the tree-split formula using both the first and second order derivatives of the logistic loss function. ~\cite{Proc:ABC_UAI10} showed the ``robust logitboost'' in general improves ``mart'', as can be seen from Figure~\ref{fig_Noise6} and Table~\ref{tab_deep}.

\cite{Article:Li_ABC_arXiv08,Proc:ABC_ICML09,Proc:ABC_UAI10} made an interesting  observation that the derivatives (as in text books) of the classical logistic loss function can be written in a different form for the multi-class case, by enforcing the ``sum-to-zero'' constraints.  At each boosting step, they identify a ``base class'' either by the ``worst-class'' criterion~\cite{Article:Li_ABC_arXiv08} or the exhaustive search method as reported in~\cite{Proc:ABC_ICML09,Proc:ABC_UAI10}. This ``adaptive base class (abc)'' strategy can be combined with either mart or robust logitboost; hence the names ``abc-mart'' and ``abc-robust-logitboost''. The improvements due to the use of ``abc'' strategy can also be substantial. In all the tree implementations, they~\cite{Article:Li_ABC_arXiv08,Proc:ABC_ICML09,Proc:ABC_UAI10} always used the adaptive-binning strategy for simplifying the implementation and speeding up training. Also, they followed the ``best-first'' criterion whereas  many tree implementations used balanced trees.

\begin{table*}[t]
\caption{Test error rates of five  datasets reported in~\cite{Proc:Larochelle_ICML07,Proc:ABC_UAI10}. The results in group 1 are from~\cite{Proc:Larochelle_ICML07}, where they compared kernel SVM, neural nets, and deep nets. The results in group 3 are  from~\cite{Proc:ABC_UAI10}, which compared four boosted tree methods.\vspace{-0.1in}}
\begin{center}{
\begin{tabular}{c l c c c c c c}
\hline \hline
Group &Method &M-Basic & M-Rotate &M-Image &M-Rand & M-RotImg\\\hline
&SVM-RBF &{3.05}\% &11.11\% &22.61\% &14.58\% &55.18\%\\
&SVM-POLY &3.69\% &15.42\% &24.01\% &16.62\% &56.41\%\\
1&NNET &4.69\% &18.11\% &27.41\% &20.04\% &62.16\%\\
&DBN-3 &3.11\% &{\bf10.30\%} &16.31\% &{\bf6.73\%} &47.39\%\\
&SAA-3 &3.46\% &{\bf10.30\%} &23.00\% &11.28\% &51.93\%\\
&DBN-1 &3.94\% &14.69\% &16.15\% &9.80\% &52.21\%\\\hline
&Linear &10.02\% &52.01\% &29.29\% &21.10\% &68.56\% \\
&RBF &{\bf2.79}\% &{\bf10.32}\% &22.16\% &14.61\% &54.16\%\\
2&GMM &3.64\% &15.24\% &19.15\% &15.78\% &59.02\%\\
&$e$GMM &3.53\%&13.98\% &18.80\%&15.05\% &57.12\%\\
&$p$GMM &3.60\%&13.44\% &10.47\%&10.91\% &45.42\%\\
&$\gamma$GMM &3.16\%&12.67\% &18.68\%&14.83\% &56.78\%\\
&$ep$GMM &3.29\%&11.64\%&10.04\%&10.57\%&{44.27}\%\\
&$p\gamma$GMM &3.00\%&10.94\%&{9.04}\%&9.53\%&{\bf40.18}\%\\\hline
&{mart} & 4.12\% &15.35\% &11.64\% & 13.15\% &49.82\%\\
3&{abc-mart} &3.69\% &13.27\% &9.45\% & 10.60\% &46.14\%\\
&{robust logitboost} &3.45\% &13.63\% & 9.41\% &10.04\%&45.92\%\\
&{abc-robust-logitboost} &3.20\% &11.92\% & \textbf{8.54\%} &9.45\%&{44.69\%}\\
\hline\hline
\end{tabular}

}
\end{center}
\label{tab_deep}
\end{table*}

\vspace{0.05in}

Table~\ref{tab_deep} reports the test error rates on five  datasets: M-Basic, M-Rotate, M-Image, M-Rand, and M-RotImg. In group 1 (as  reported in~\cite{Proc:Larochelle_ICML07}), the results show that (i) the kernel SVM with RBF kernel outperforms the kernel SVM with polynomial kernel; (ii) deep learning algorithms usually beat kernel SVM and neural nets. Group 2 presents the same results as in Table~\ref{tab_MNIST} (in terms of error rates as opposed to accuracies). In group 3, overall the tree methods especially abc-robust-logitboost  achieve very good accuracies. The results of tunable GMM kernels are largely comparable.

\vspace{0.08in}

The training of boosted trees is typically  slow (especially in high-dimensional data) because a large number of trees are usually needed in order to achieve good accuracies. Consequently, the model sizes of tree methods are usually  large. Therefore, it would be exciting to have methods which are much  simpler than trees and achieve comparable accuracies.


\section{Hashing the $p$GMM Kernel}

It is now well-understood that it is highly beneficial to be able to linearize nonlinear kernels so that learning algorithms can be easily scaled to massive data. Linearization can be done either through hashing~\cite{Report:Manasse_CWS10,Proc:Ioffe_ICDM10,Proc:Li_KDD15} or the Nystrom method~\cite{Article:Nystrom1930}.

It turns out that developing a hashing method for the $p$GMM kernel is quite straightforward, by modifying the prior algorithms. Algorithm~\ref{alg_GCWS} summarizes the modified GCWS (generalized consistent weighted sampling) .

\begin{algorithm}{

\hspace{-1.2in}\textbf{Input:} Data vector $u_i$  ($i=1$ to $D$)

\hspace{-1.1in}Generate vector $\tilde{u}$ in $2D$-dim by (\ref{eqn_transform}).

\vspace{0.08in}

\hspace{-2in}For $i$ from 1 to $2D$

\hspace{-0in}$r_i\sim Gamma(2, 1)$, $c_i\sim Gamma(2, 1)$,  $\beta_i\sim Uniform(0, 1)$

\hspace{-0.4in} $t_i\leftarrow \lfloor p\frac{\log \tilde{u}_i }{r_i}+\beta_i\rfloor$, $a_i\leftarrow \log(c_i)- r_i(t_i+1-\beta_i)$

\hspace{-2.5in} End For

\hspace{-0.7in}\textbf{Output:} $i^* \leftarrow arg\min_i \ a_i$,\hspace{0.3in}  $t^* \leftarrow t_{i^*}$
}\caption{Modified generalized consistent weighted sampling (GCWS) for hashing the $p$GMM kernel with a tuning parameter $p$.}
\label{alg_GCWS}
\end{algorithm}

\noindent With $k$ samples, we  can  estimate $p$GMM$(u,v)$ according to the following collision probability:
\begin{align}
&\mathbf{Pr}\left\{i^*_{\tilde{u},j} = i^*_{\tilde{v},j} \ \text{and} \ t^*_{\tilde{u},j} = t^*_{\tilde{v},j}\right\} = p\text{GMM}(u,v),
\end{align}
or, for implementation  convenience, the approximate collision probability~\cite{Proc:Li_KDD15}:
\begin{align}\label{eqn_GCWS_Prob}
\mathbf{Pr}\left\{i^*_{\tilde{u},j} =  i^*_{\tilde{v},j}\right\} \approx p\text{GMM}({u},{v})
\end{align}

For each  vector $u$, we obtain $k$ random samples $i^*_{\tilde{u},j}$, $j=1$ to $k$. We store only the lowest $b$ bits of $i^*$. We need to view those $k$ integers as locations (of the nonzeros). For example, when $b=2$, we should view $i^*$ as a  binary vector of length $2^b=4$.  We  concatenate all $k$ such vectors into a binary vector of length $2^b\times k$ and then feed the new data vectors to a linear classifier if the task is classification. The storage and computational cost is largely determined by the number of nonzeros in each data vector, i.e., the $k$ in our case. This scheme can  also be used for   other tasks including clustering, regression, and near neighbor search.

\vspace{0.08in}

\begin{figure}[t]
\begin{center}

\mbox{
\includegraphics[width=1.75in]{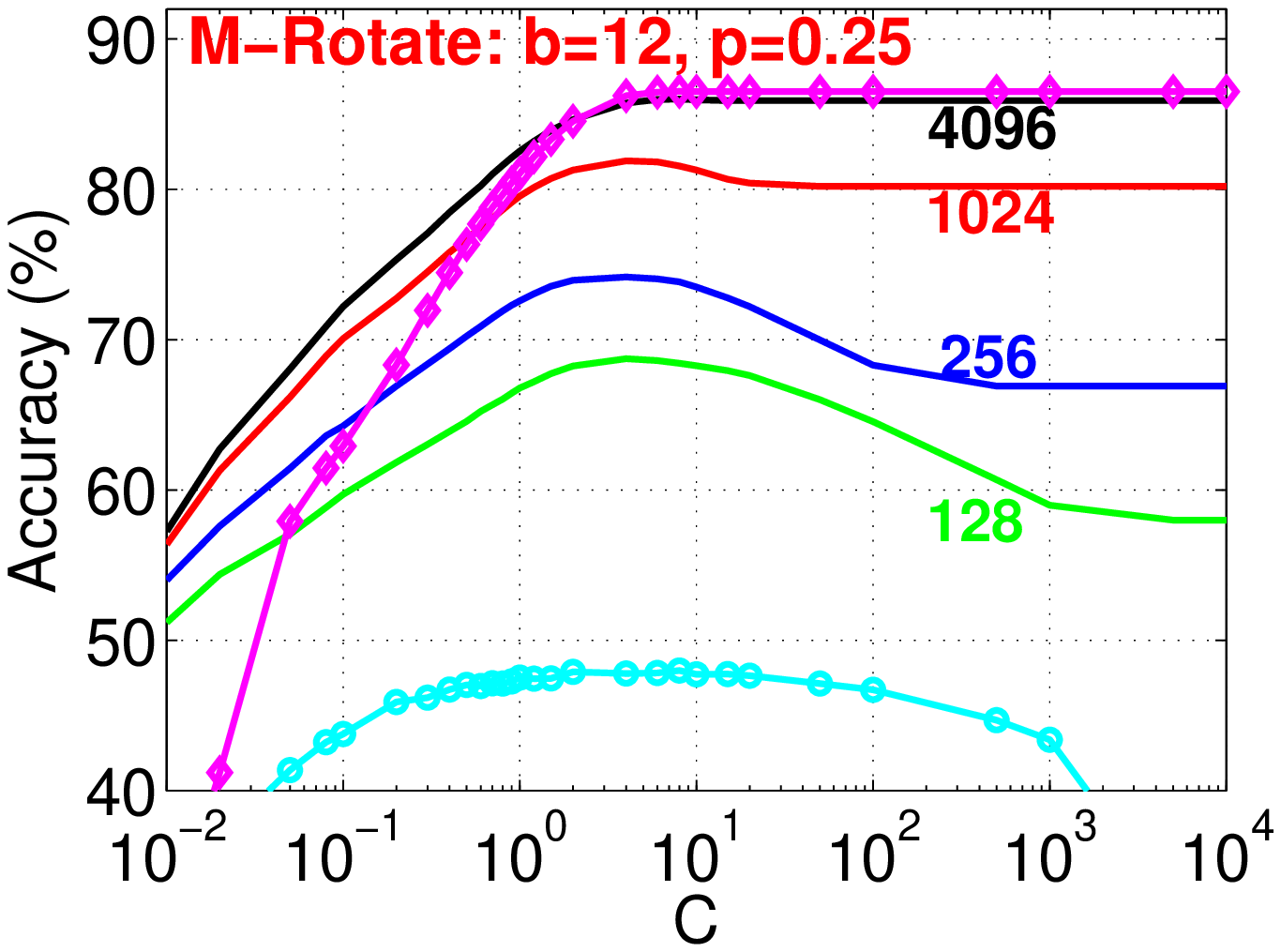}\hspace{-0.12in}
\includegraphics[width=1.75in]{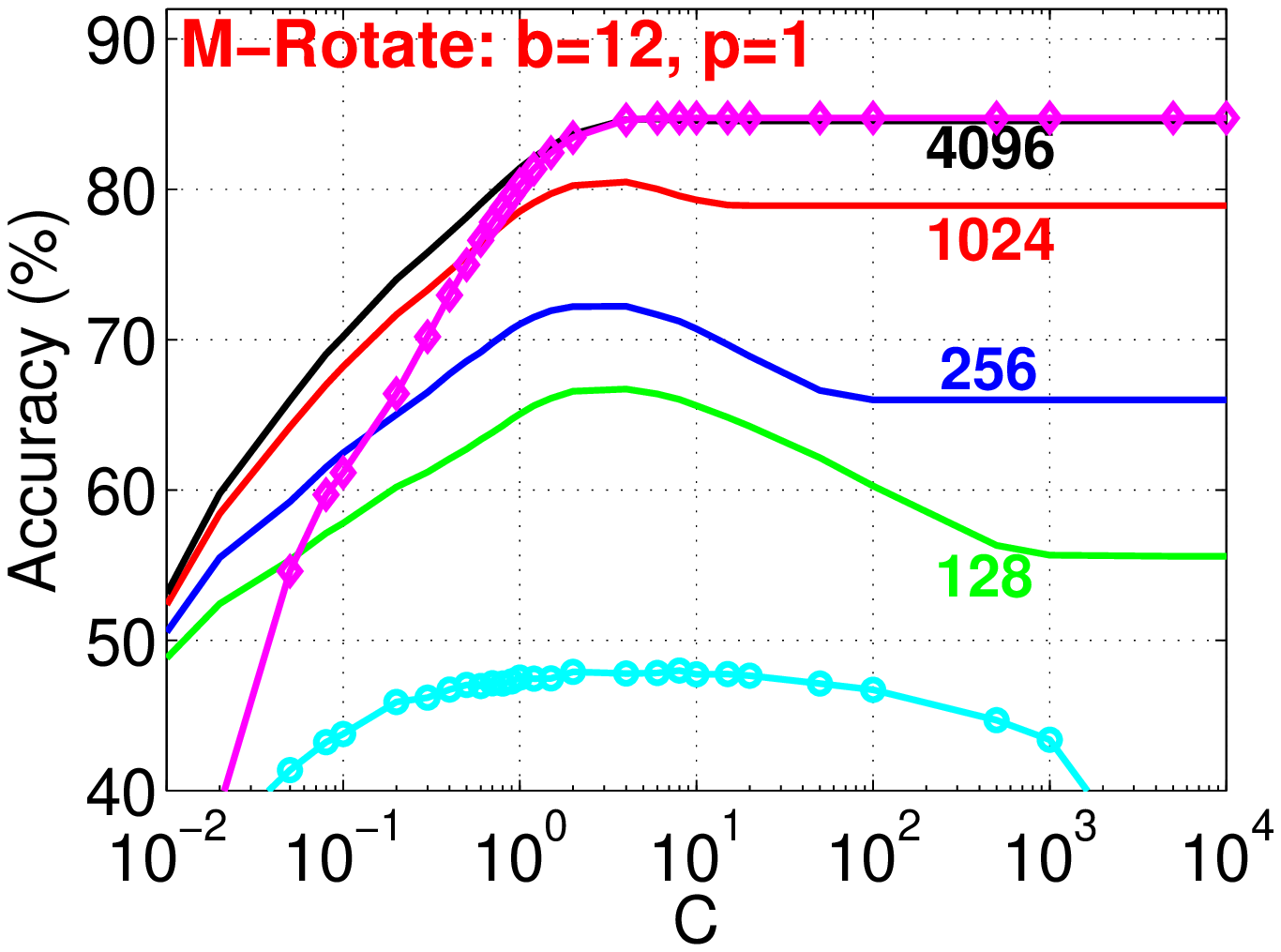}
}

\mbox{
\includegraphics[width=1.75in]{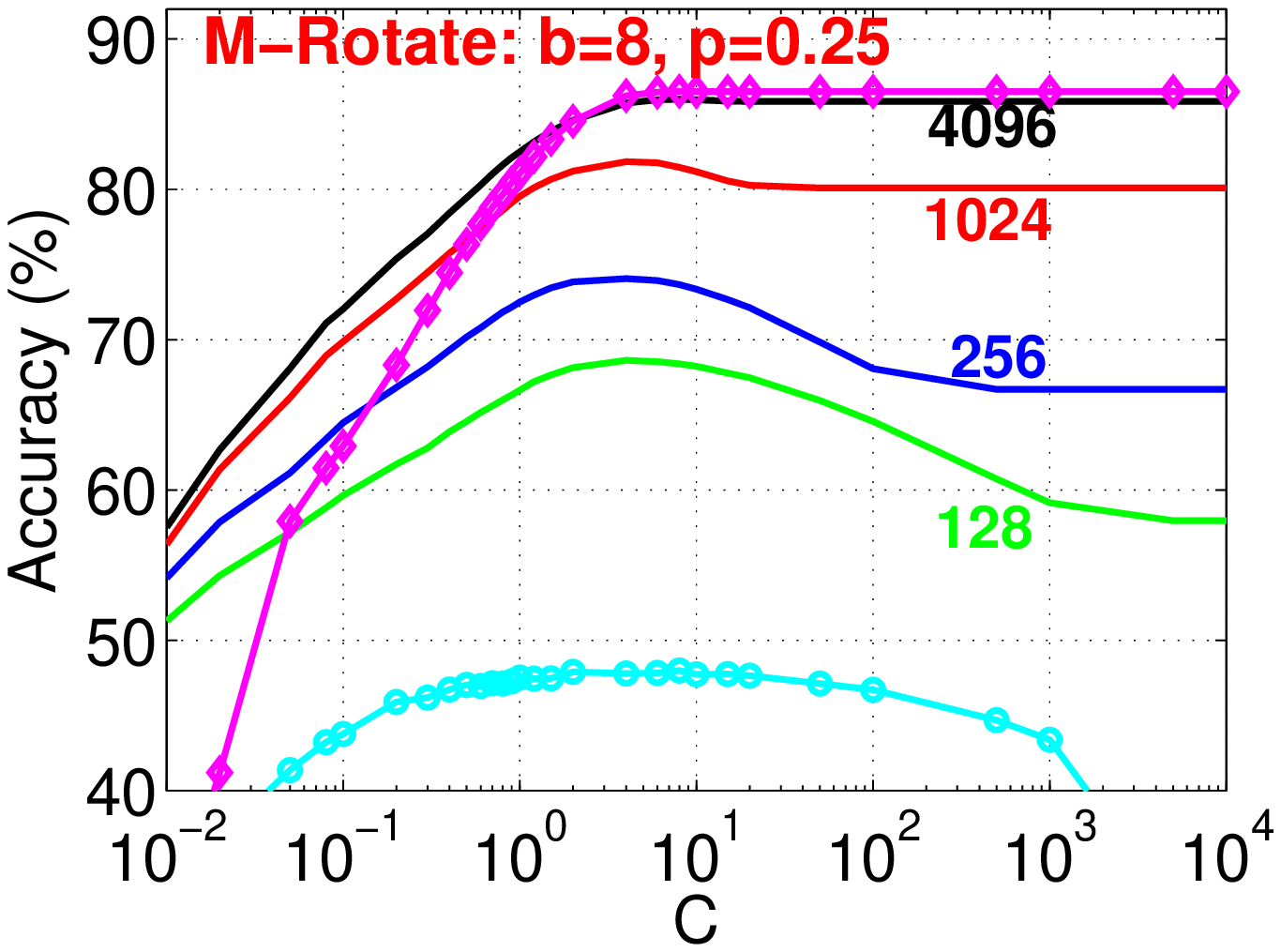}\hspace{-0.12in}
\includegraphics[width=1.75in]{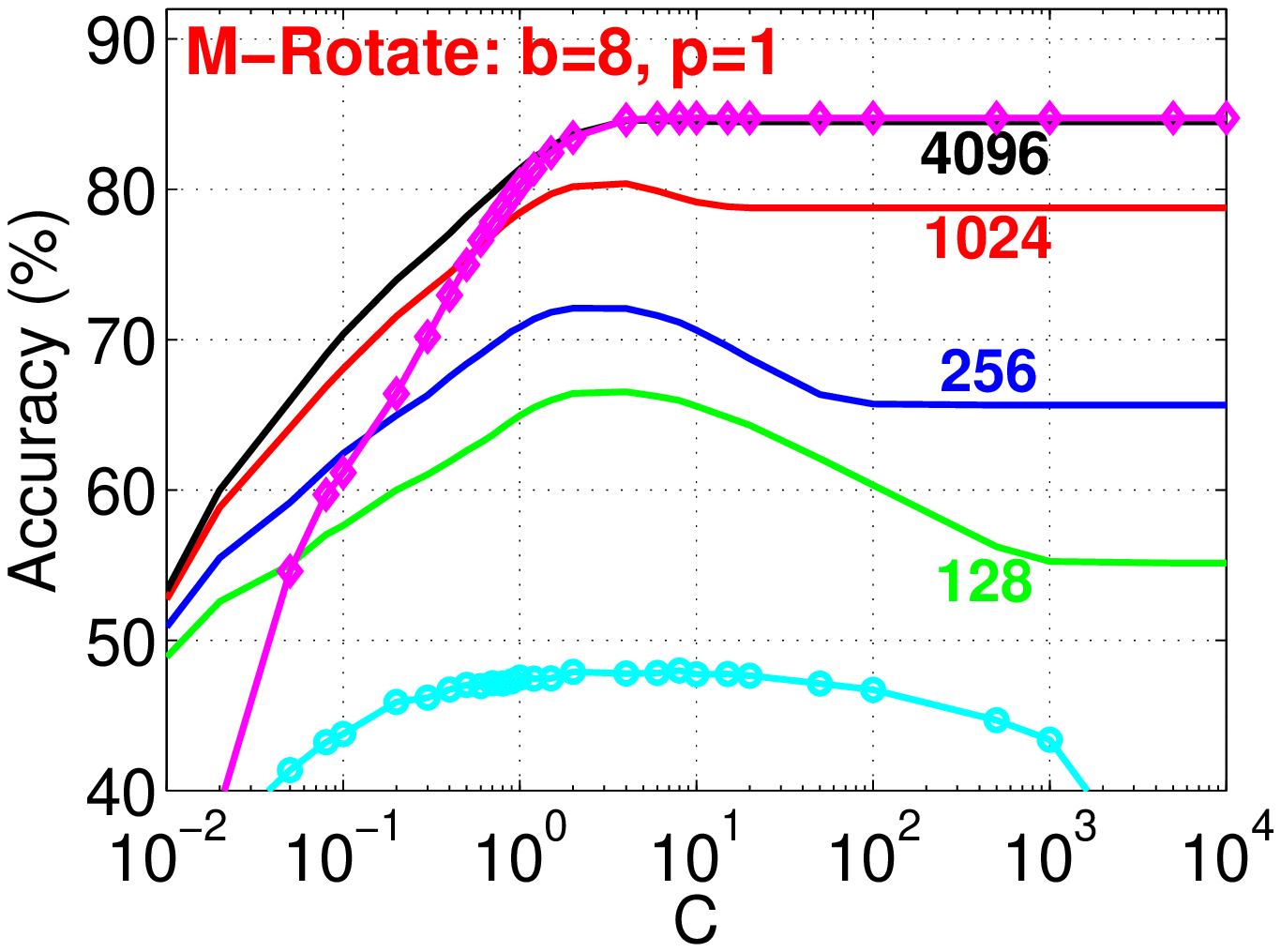}
}

\mbox{
\includegraphics[width=1.75in]{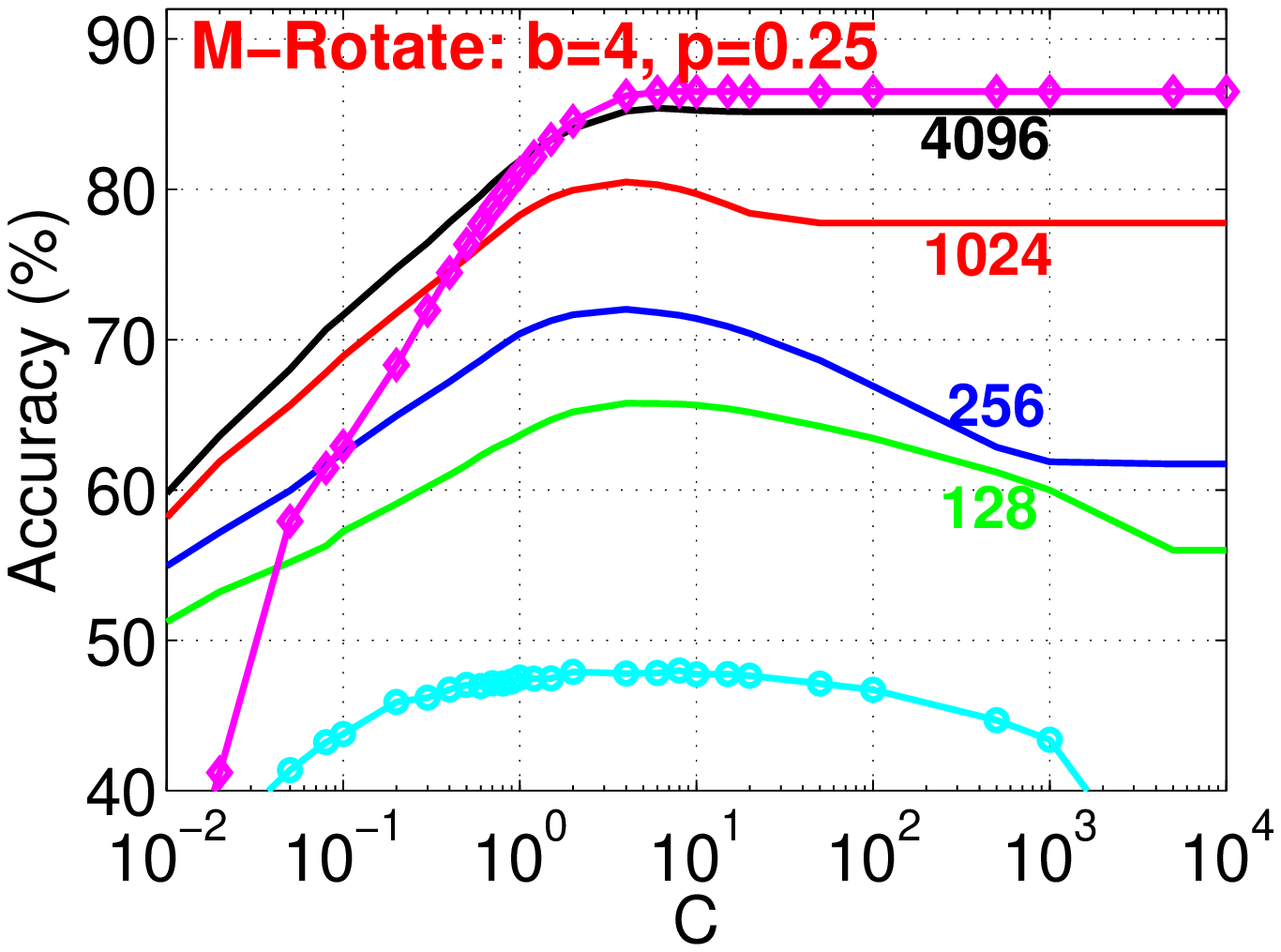}\hspace{-0.12in}
\includegraphics[width=1.75in]{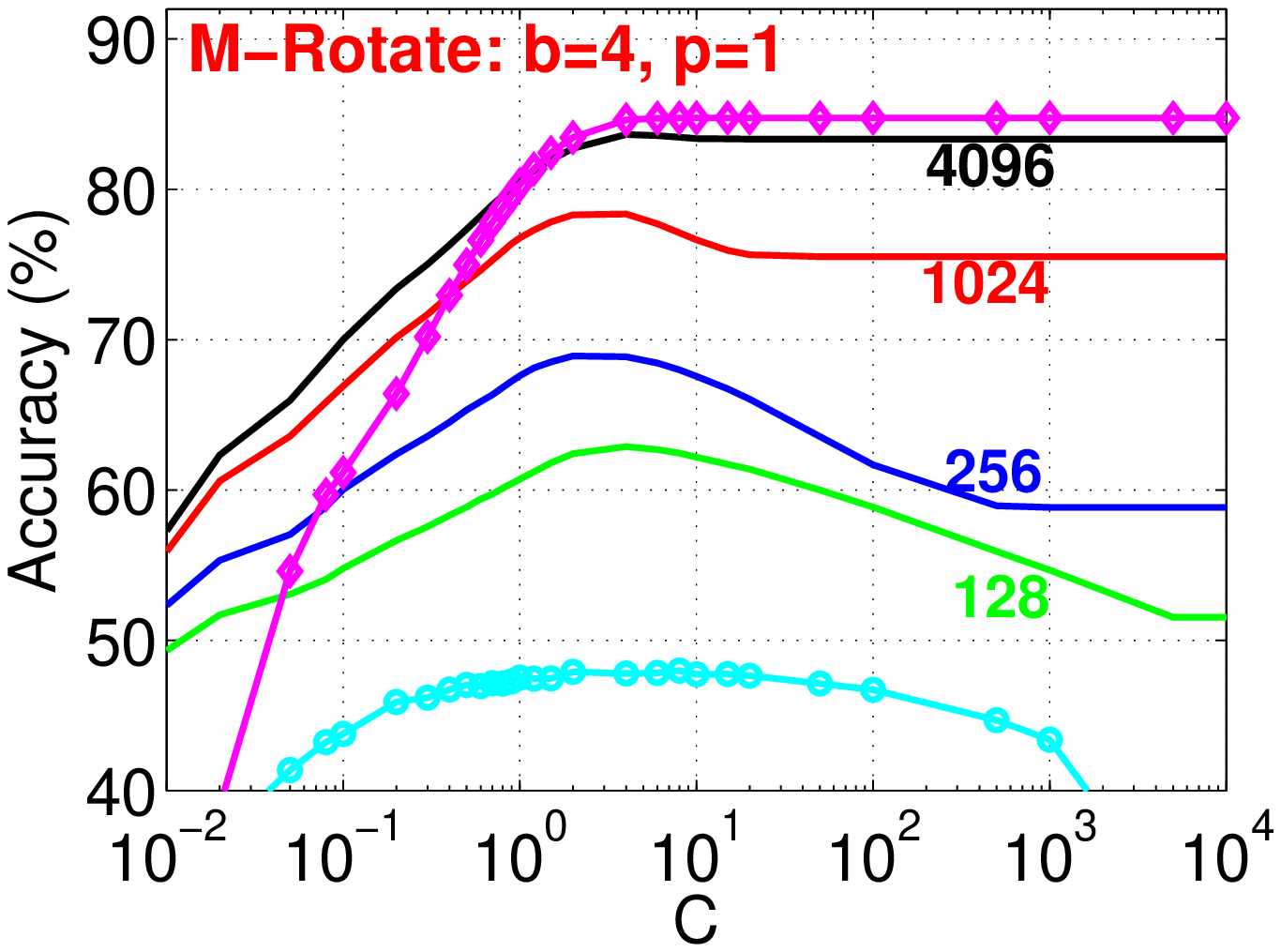}
}

\mbox{
\includegraphics[width=1.75in]{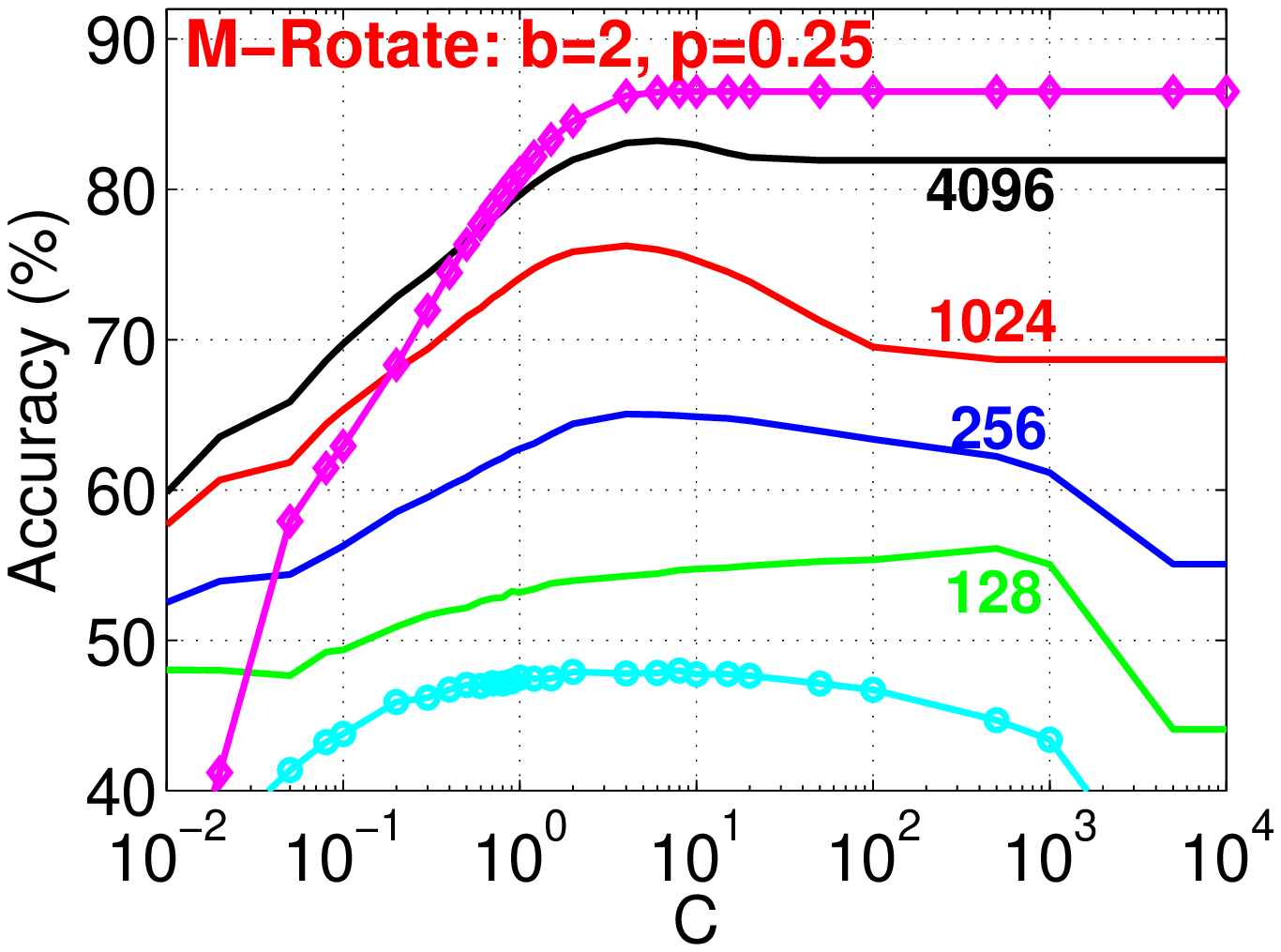}\hspace{-0.12in}
\includegraphics[width=1.75in]{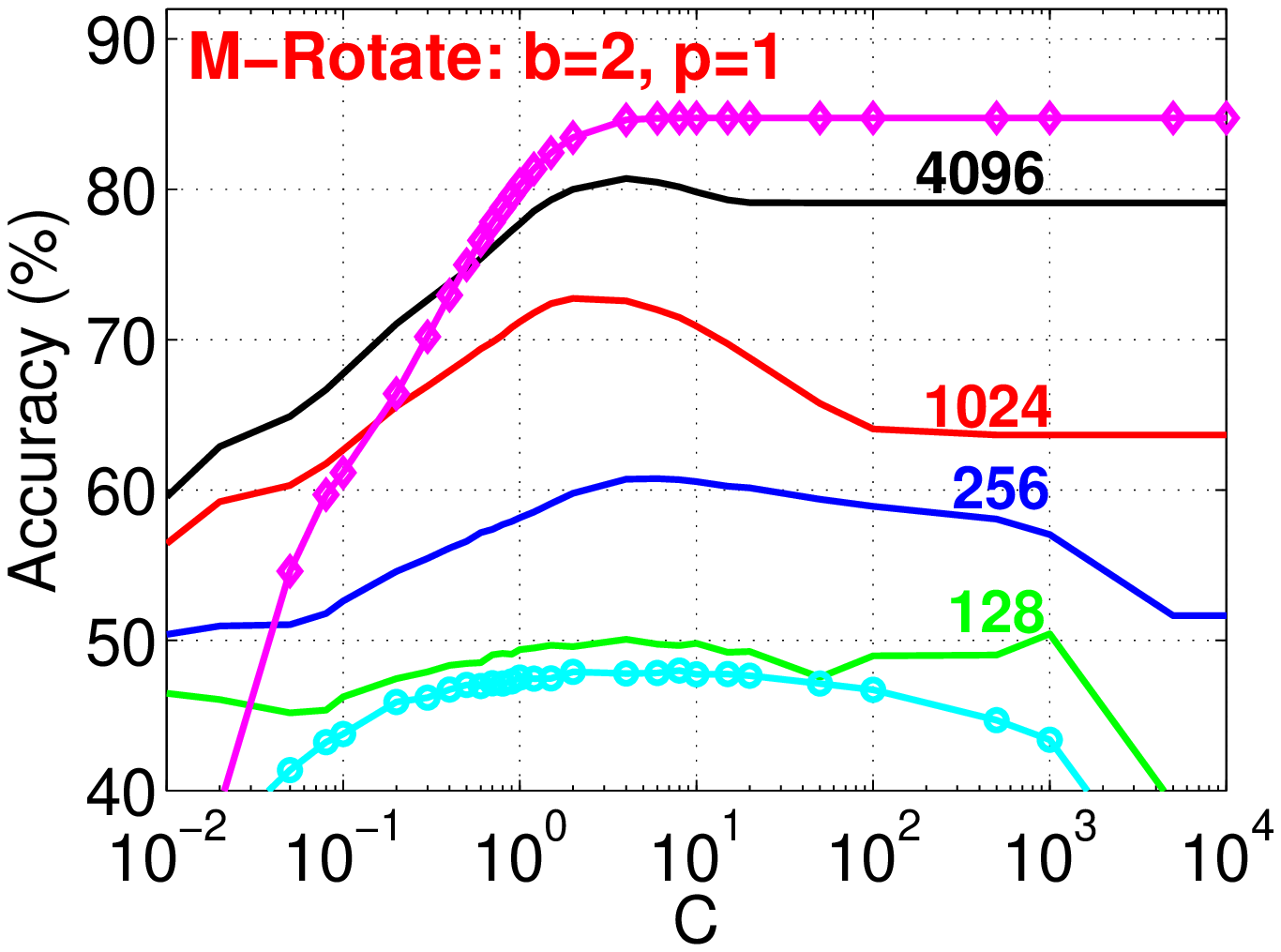}
}

\end{center}
\vspace{-0.15in}
\caption{Test classification accuracies for using linear classifiers combined with hashing in Algorithm~\ref{alg_GCWS} on M-Rotate dataset, for $p = 0.25$ (left panels) and $p = 1$ (right panels), and $b\in\{12, 8, 4, 2\}$. In each panel, the four solid curves correspond to results with $k$ hashes for $k\in\{128, 256,1024, 4096\}$. For comparisons, each panel  also plots the results of linear classifiers on the original data (lower marked curve) and the results of $p$GMM kernel SVMs (higher marked curve). }\label{fig_HashM-Rotate}
\end{figure}

Figure~\ref{fig_HashM-Rotate} presents the experimental results on hashing for  M-Rotate. For this dataset, $p=0.25$ is the best choice (among the range of $p$ values we have searched). Figure~\ref{fig_HashM-Rotate} plots the results for both $p=0.25$ (left panels) and $p=1$ (right panels), for $b\in\{12, 8, 4, 2\}$. Recall here $b$ is the number of bits for representing each hashed value. The results demonstrate that: (i) hashing using $p=0.25$ produces better results than hashing using $p=1$; (ii) It is preferable  to use a fairly large $b$ value, for example, $b\geq 4$ or 8. Using smaller $b$ values (e.g., $b=2$) hurts the accuracy; (iii) With merely a small number of hashes (e.g., $k=128$), the linearized $p$GMM kernel can significantly outperform the original linear kernel. Note that the original dimensionality is 784. This example illustrates the significant advantage of nonlinear kernel and hashing.

\begin{figure*}[h!]
\begin{center}
\mbox{
\includegraphics[width=1.75in]{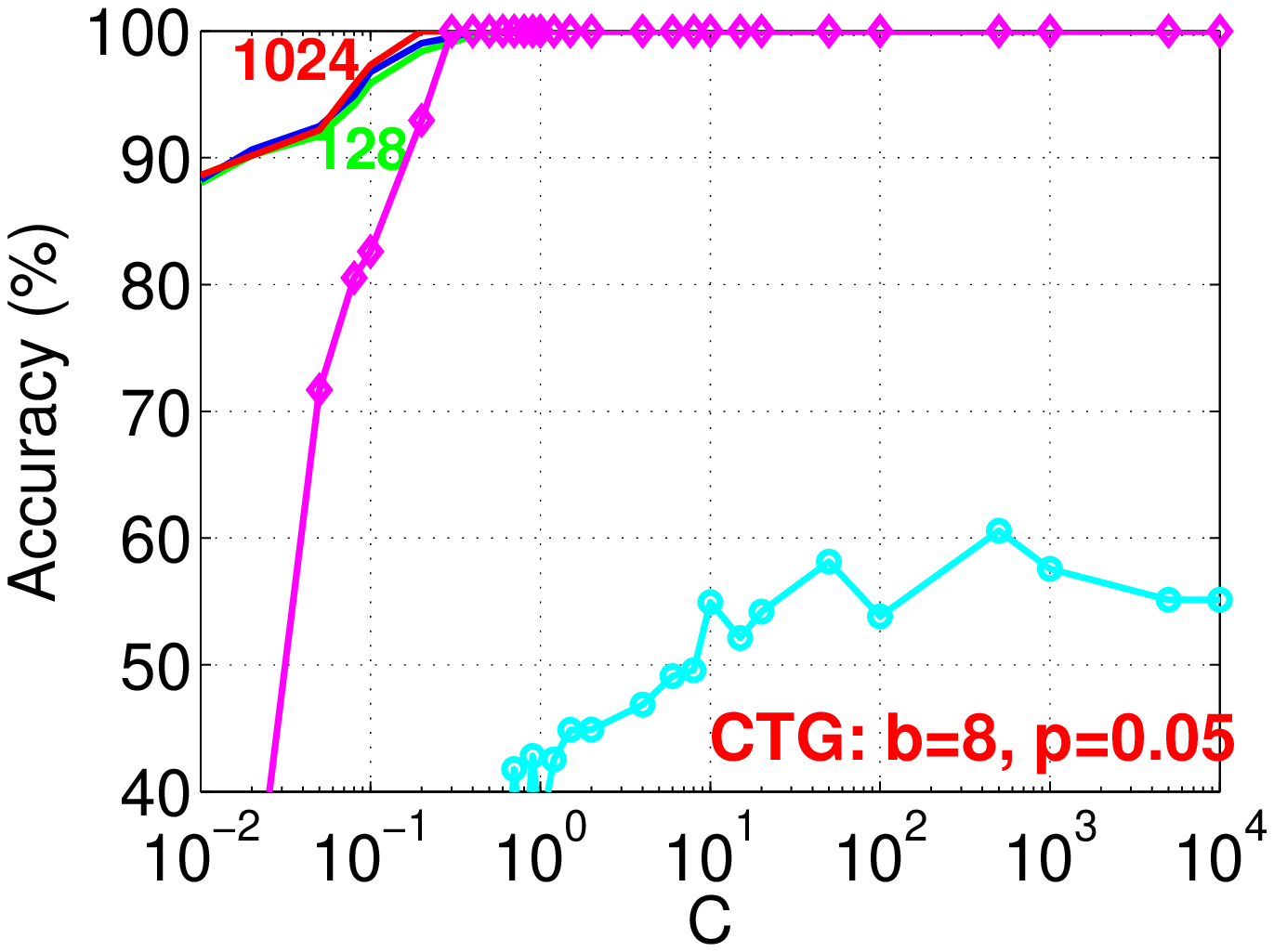}\hspace{-0.14in}
\includegraphics[width=1.75in]{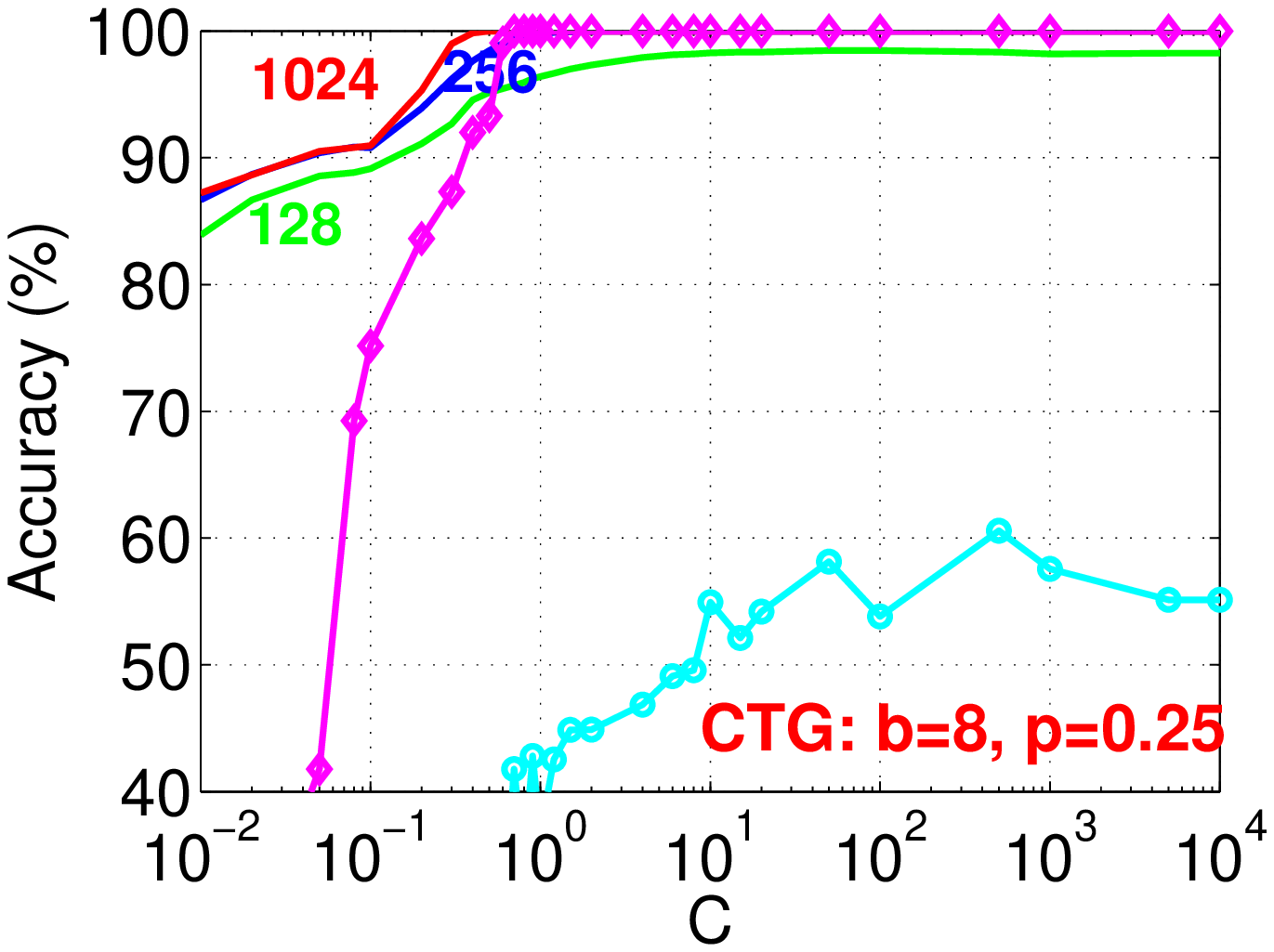}\hspace{-0.14in}
\includegraphics[width=1.75in]{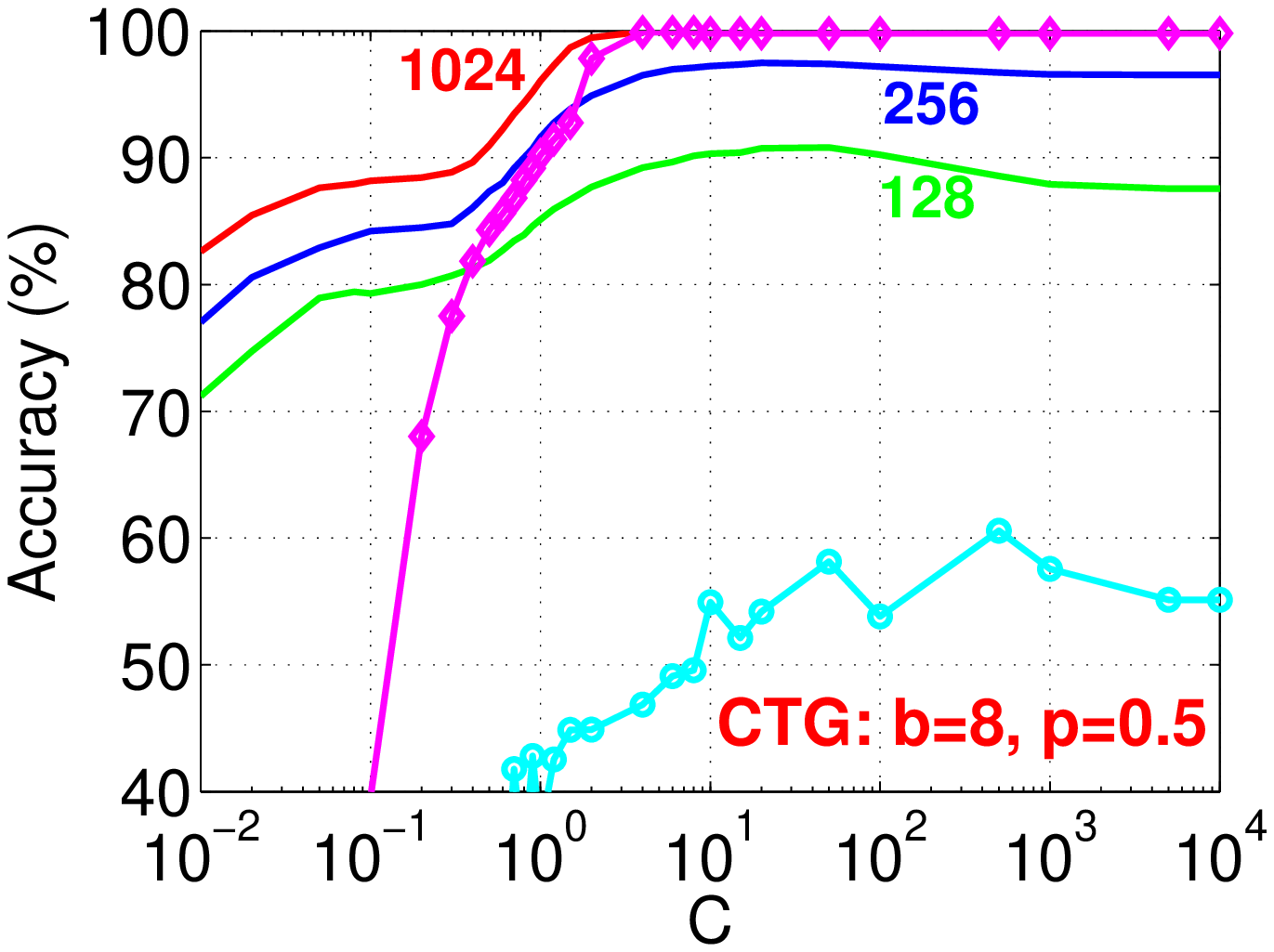}\hspace{-0.14in}
\includegraphics[width=1.75in]{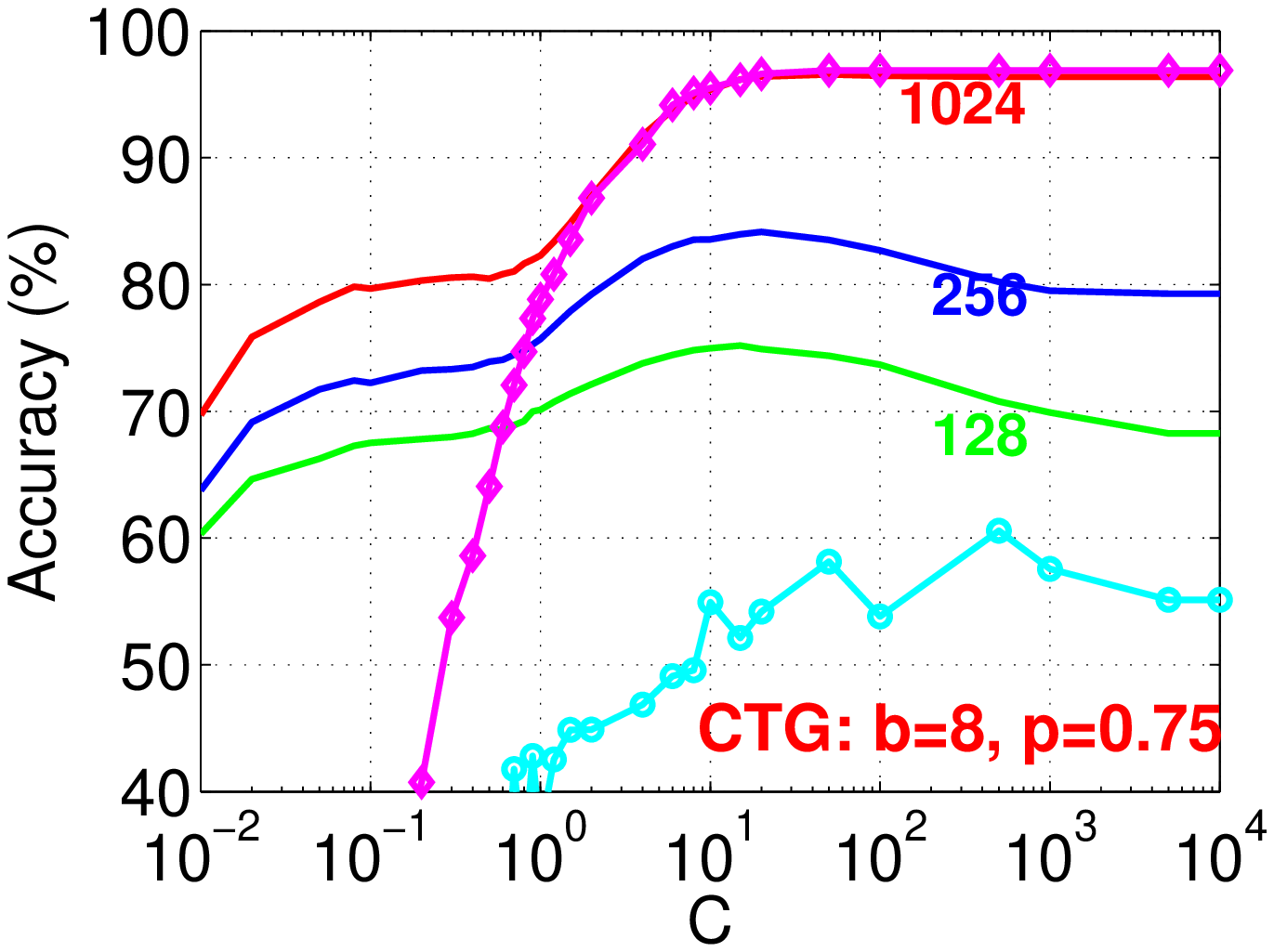}
}

\mbox{
\includegraphics[width=1.75in]{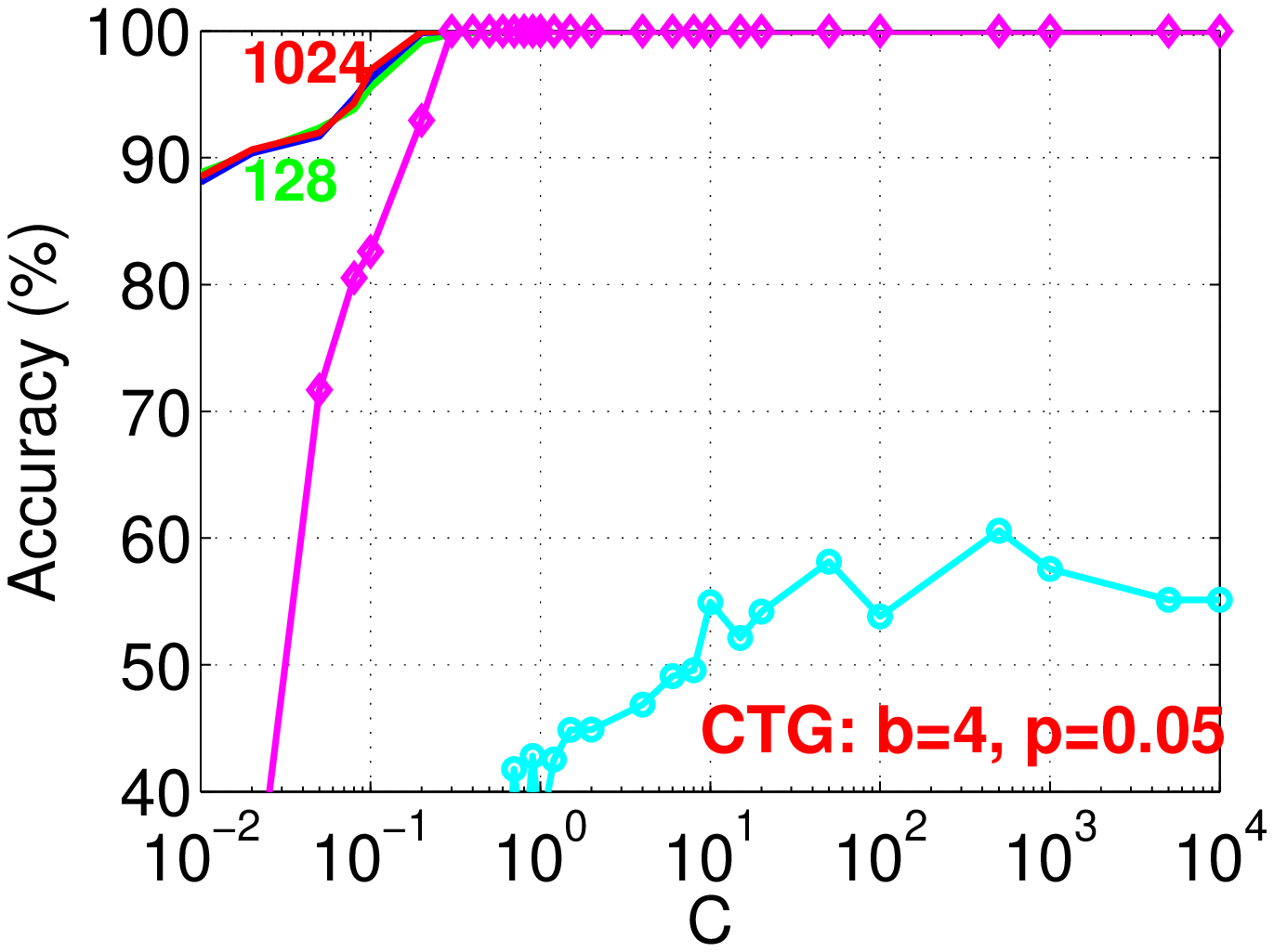}\hspace{-0.14in}
\includegraphics[width=1.75in]{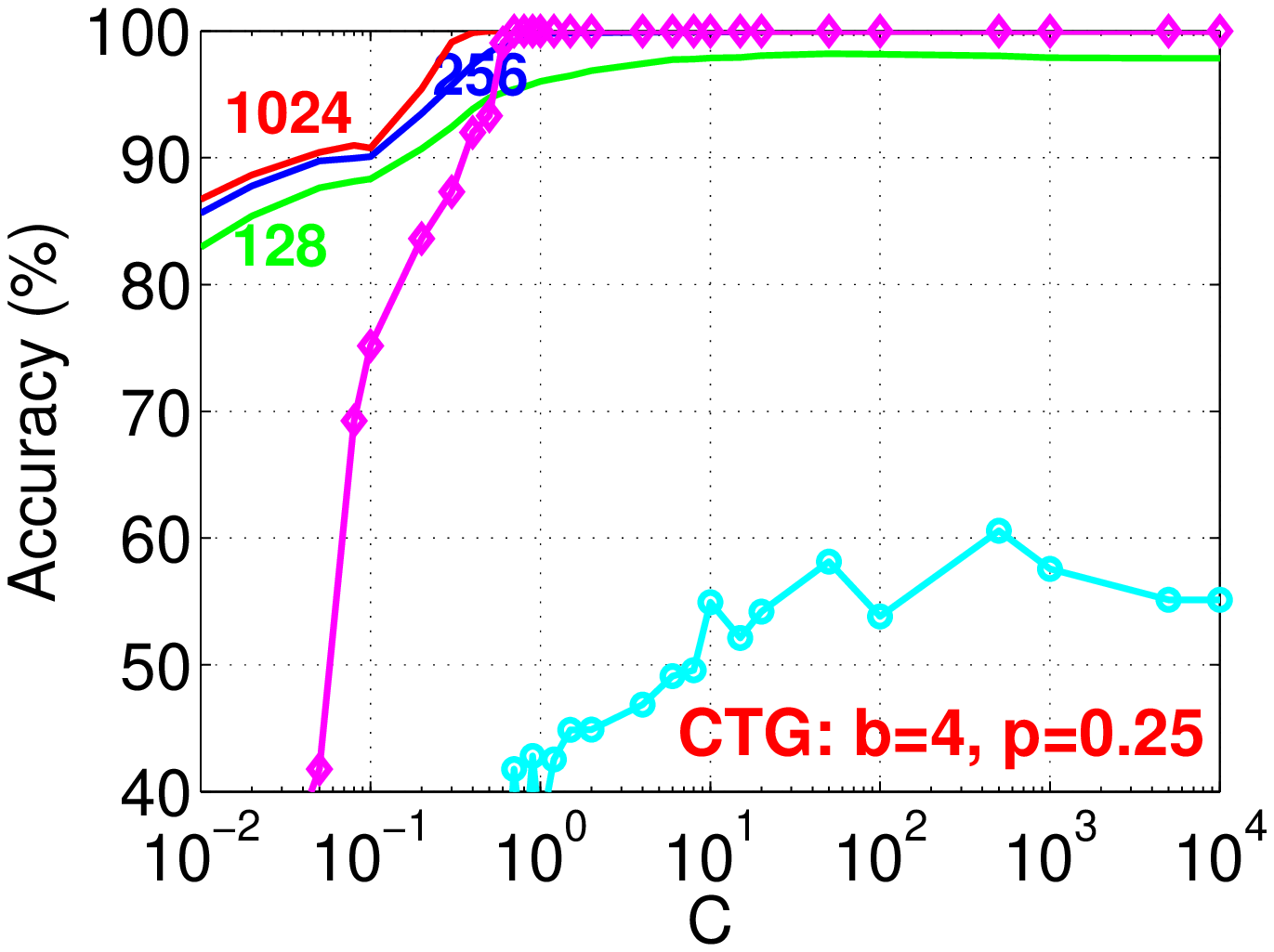}\hspace{-0.14in}
\includegraphics[width=1.75in]{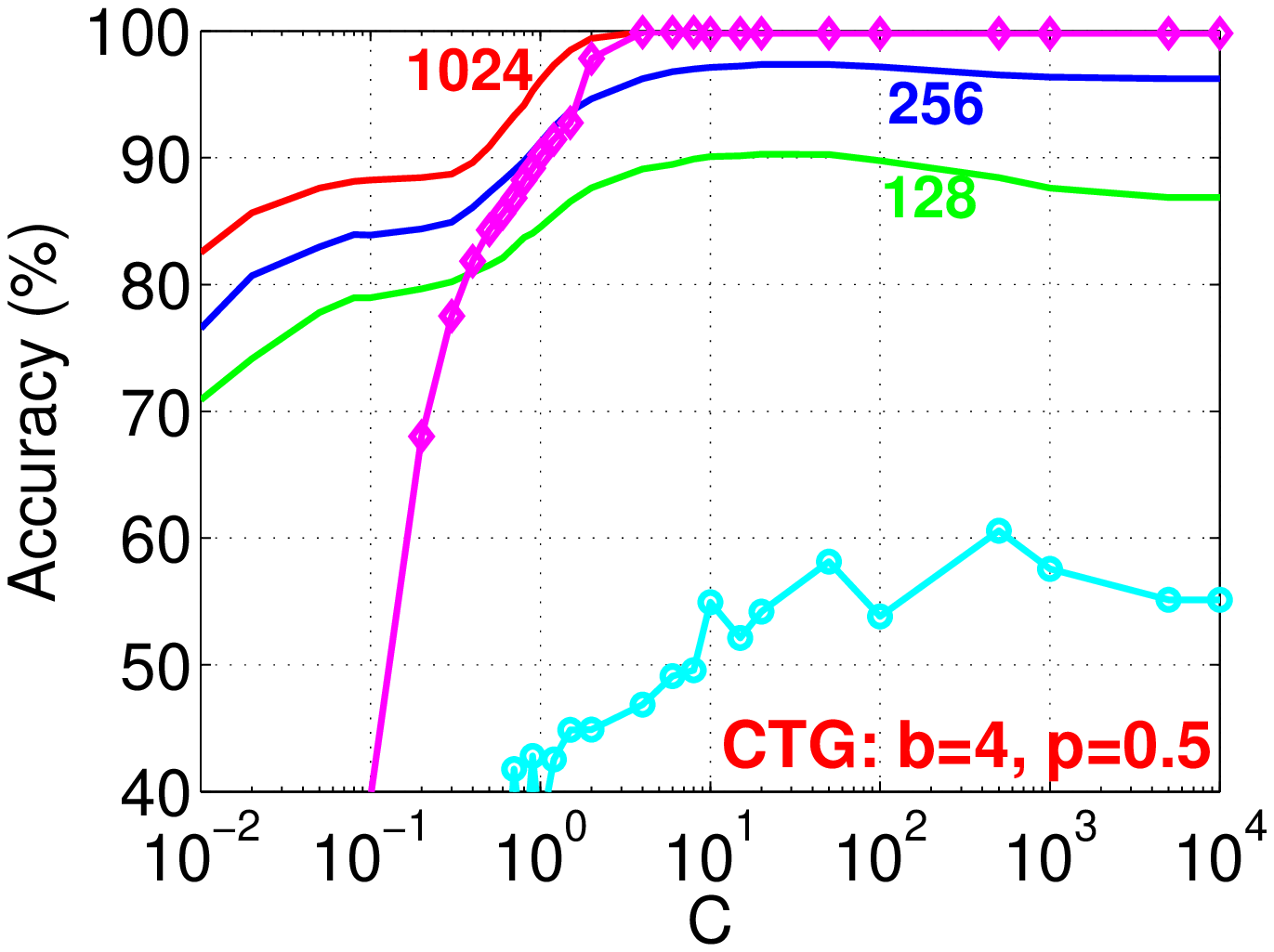}\hspace{-0.14in}
\includegraphics[width=1.75in]{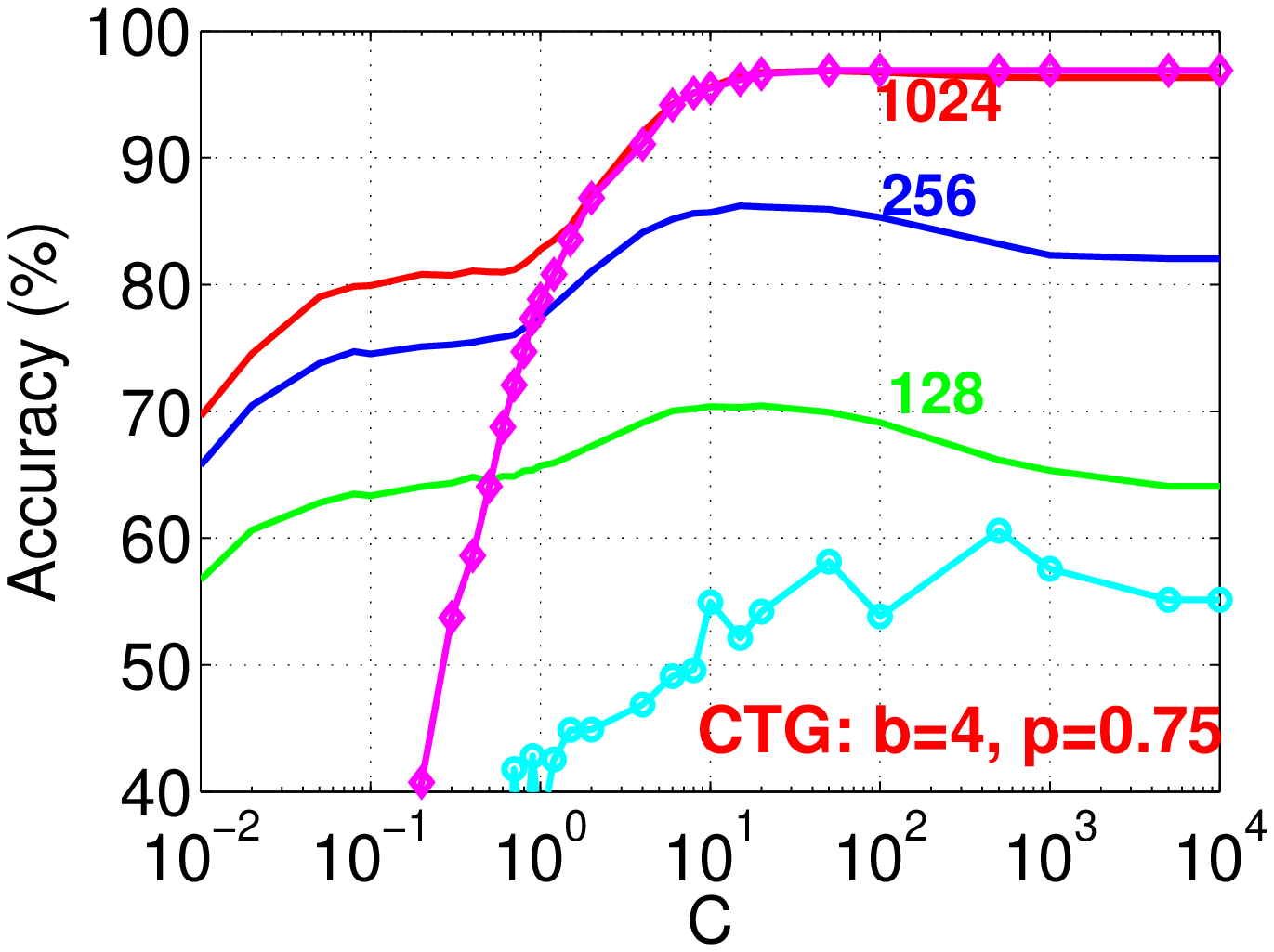}
}

\mbox{
\includegraphics[width=1.75in]{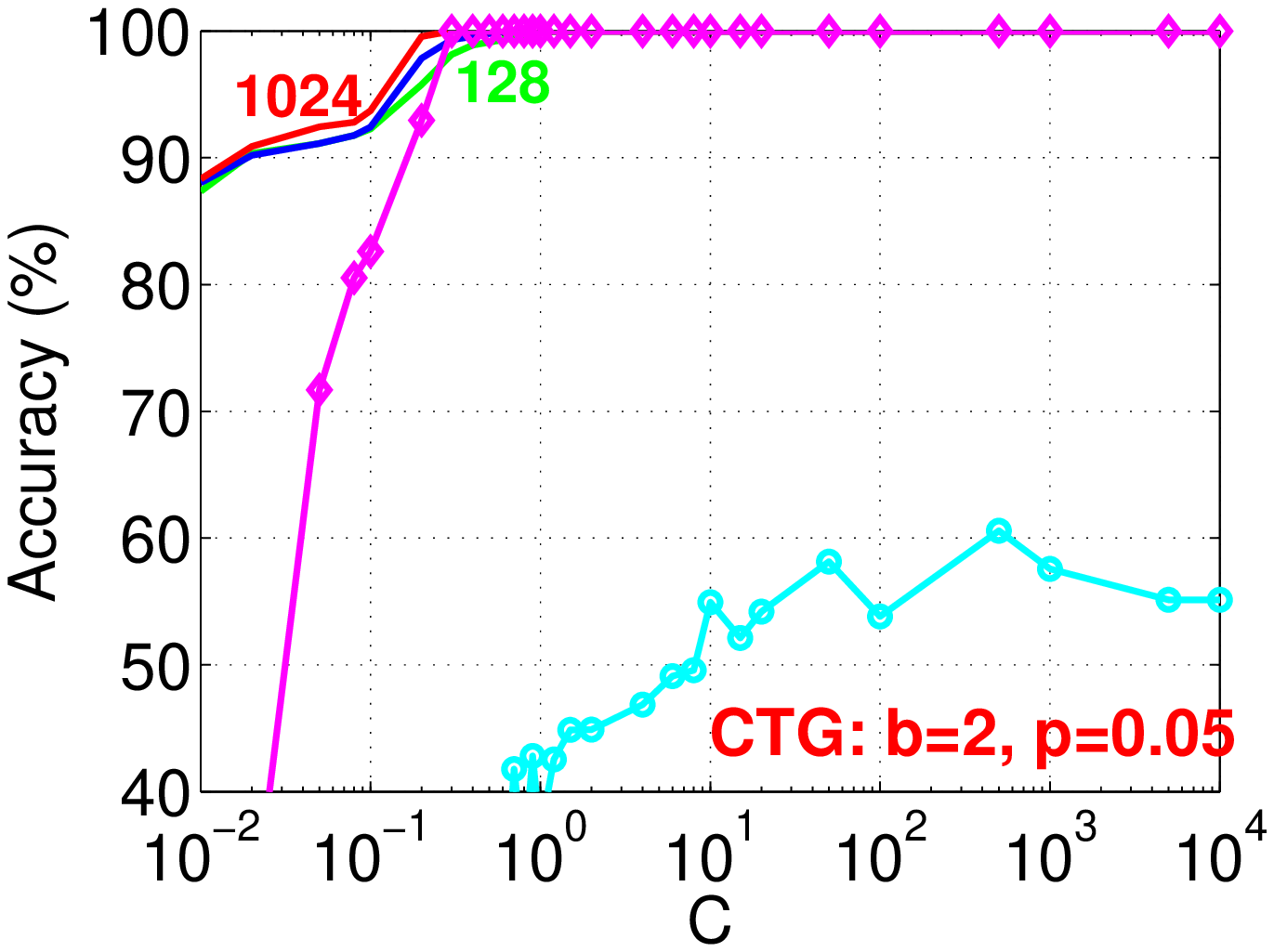}\hspace{-0.14in}
\includegraphics[width=1.75in]{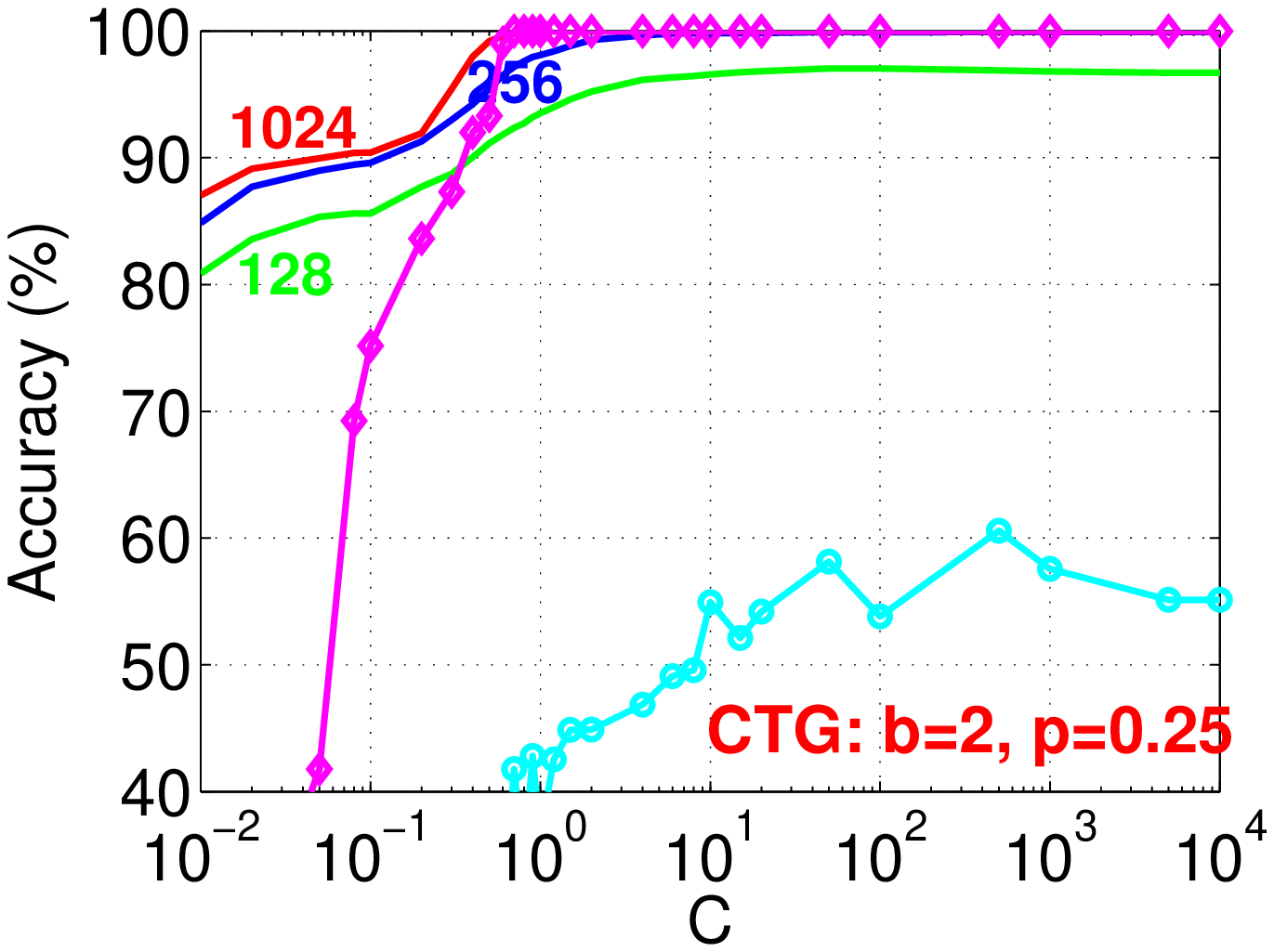}\hspace{-0.14in}
\includegraphics[width=1.75in]{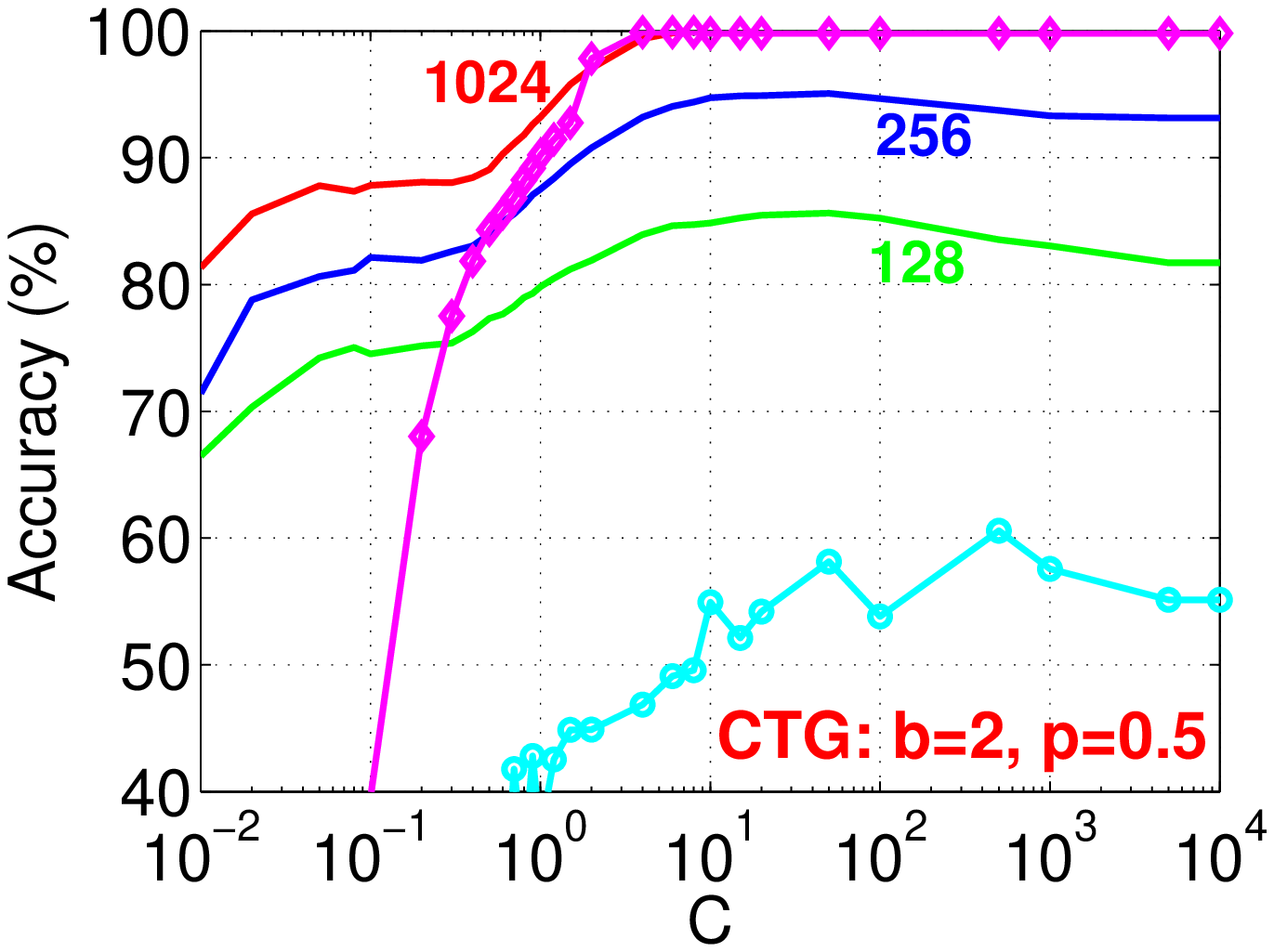}\hspace{-0.14in}
\includegraphics[width=1.75in]{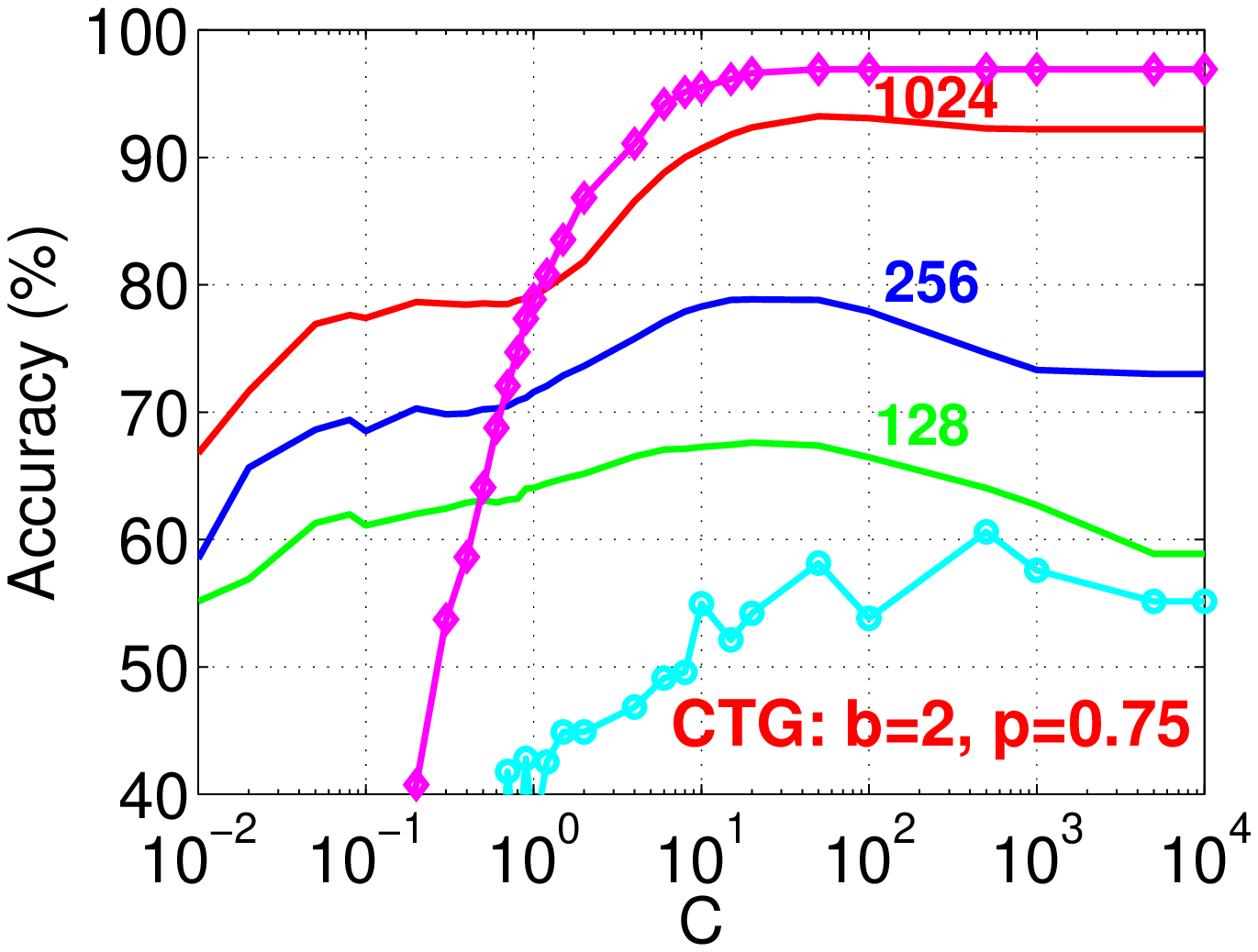}
}

\end{center}
\vspace{-0.15in}
\caption{Test classification accuracies for using linear classifiers combined with hashing in Algorithm~\ref{alg_GCWS} on CTG dataset, for $p\in\{0.05, 0.25, 0.5, 0.75\}$ to visualize the trend that, for this dataset, the accuracy decreases with increasing $p$.  }\label{fig_HashCTG10C}\vspace{-0.1in}
\end{figure*}

\begin{figure*}[h!]
\begin{center}
\mbox{
\includegraphics[width=1.75in]{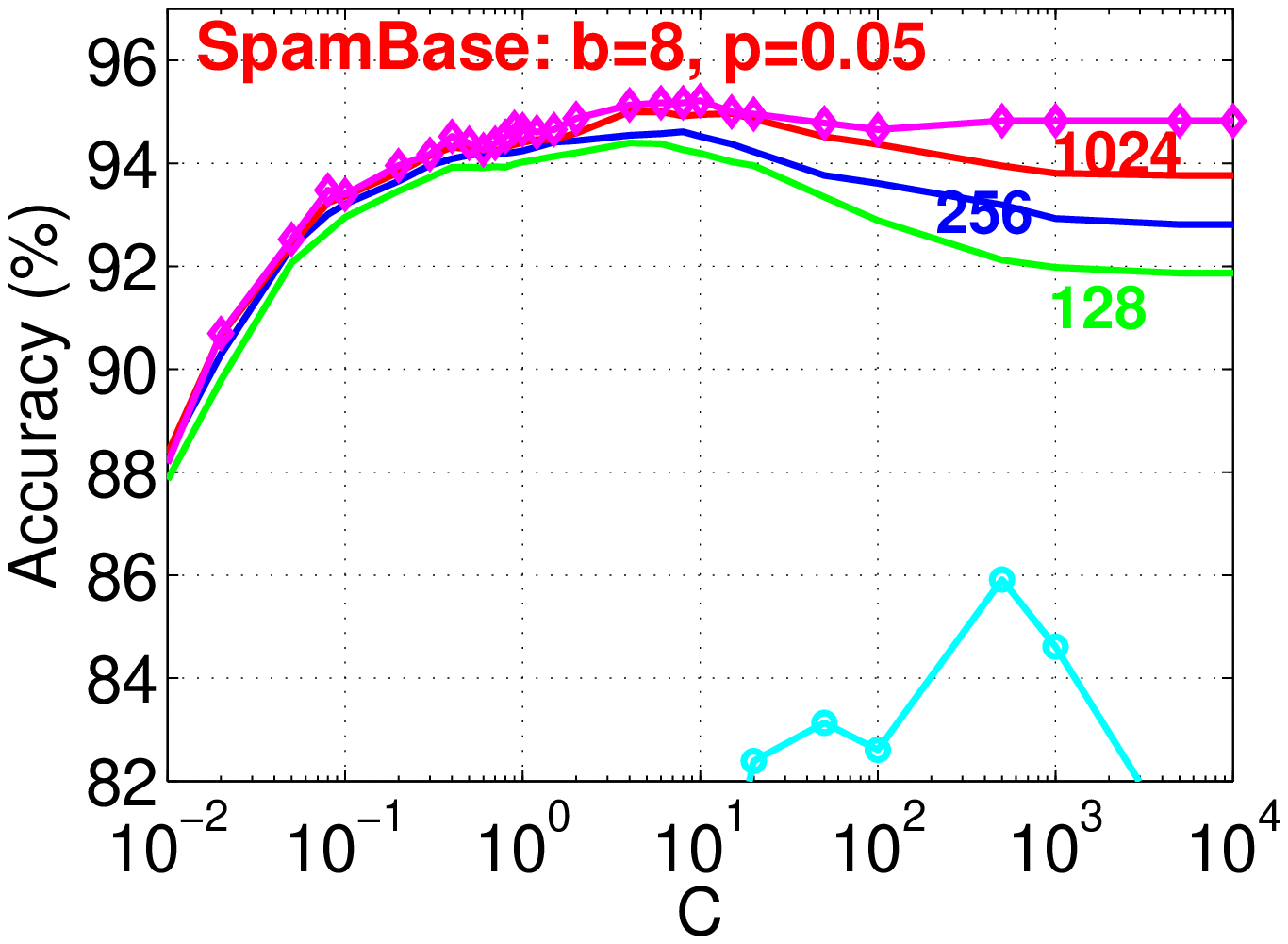}\hspace{-0.14in}
\includegraphics[width=1.75in]{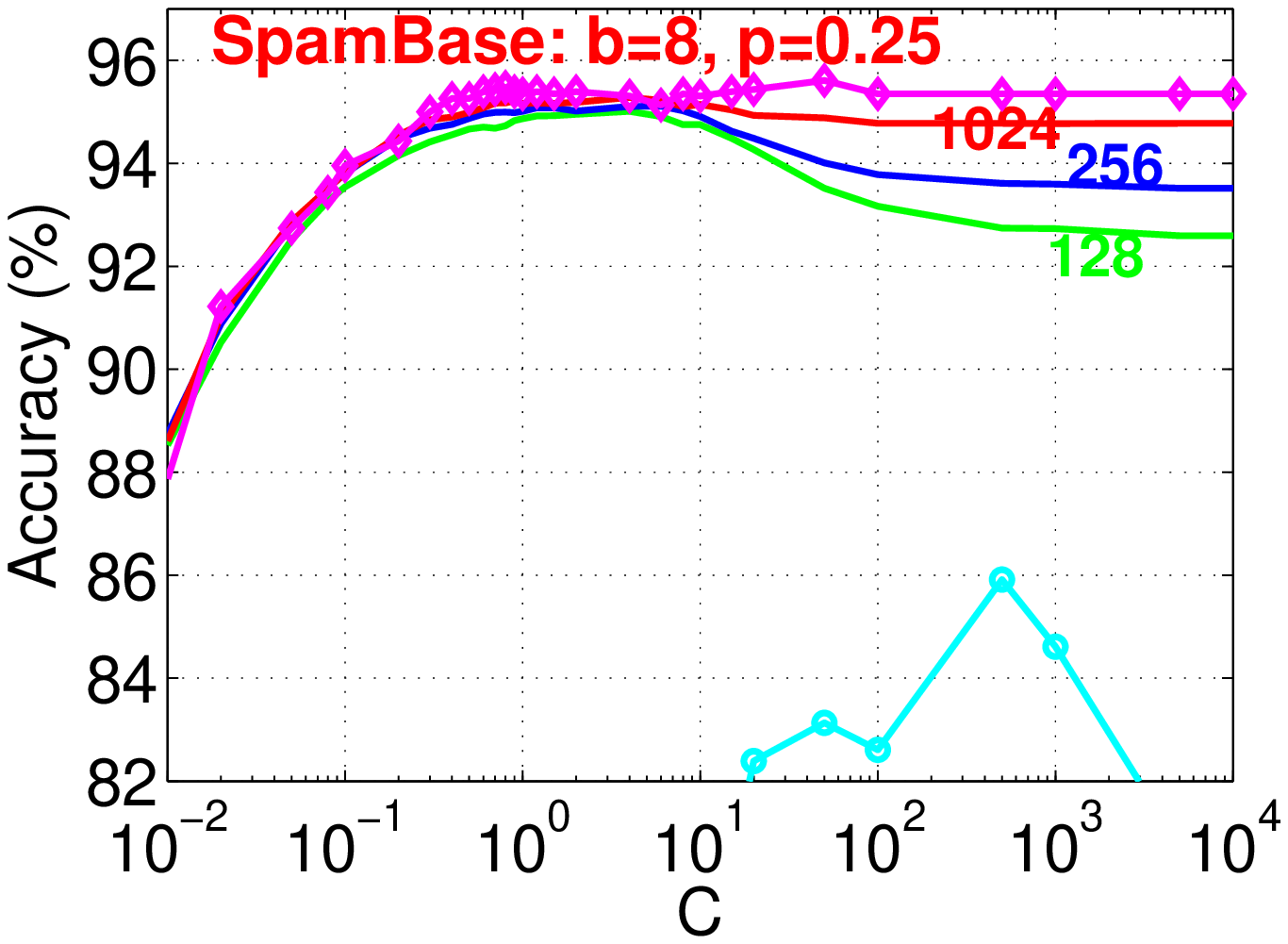}\hspace{-0.14in}
\includegraphics[width=1.75in]{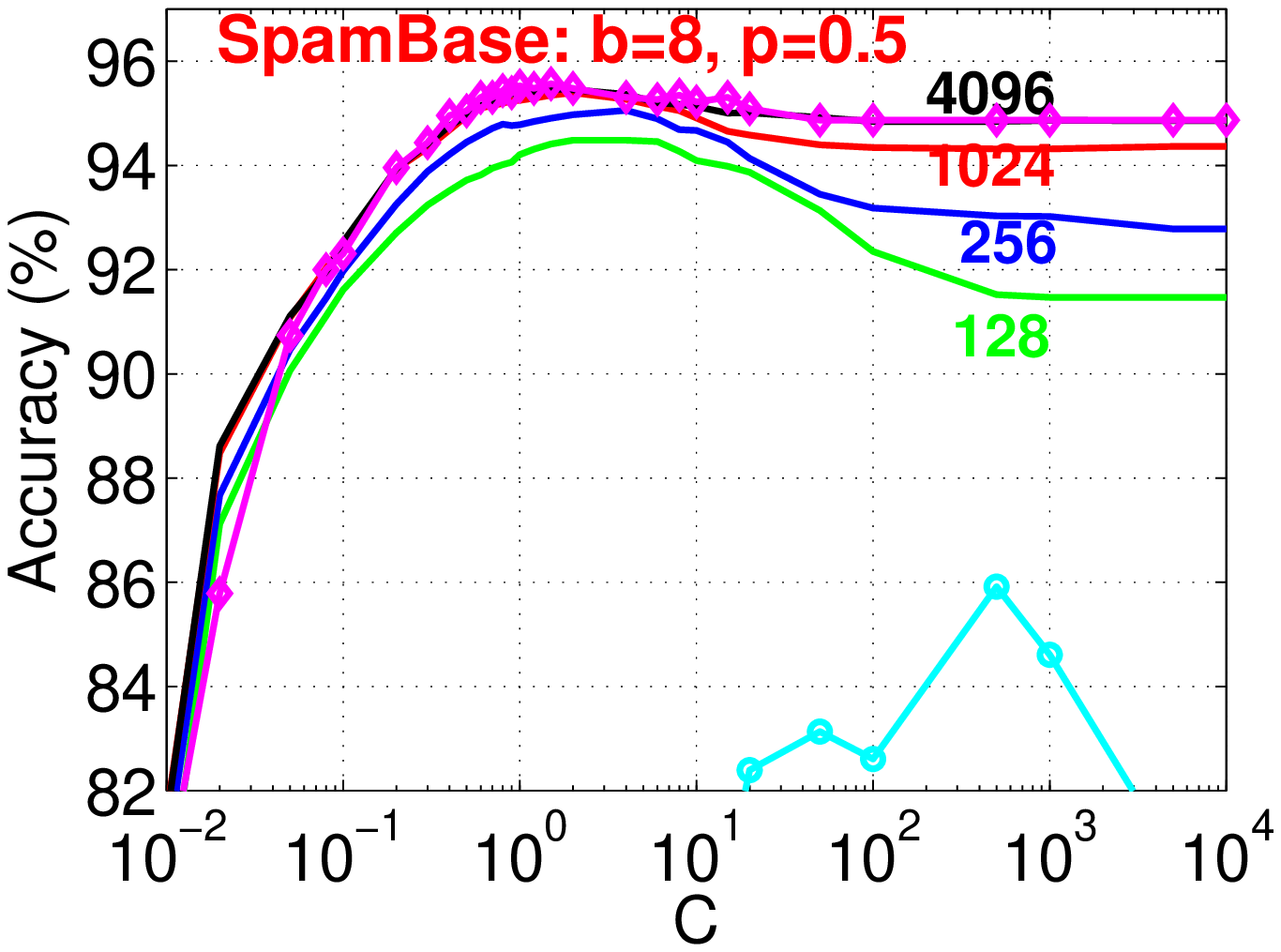}\hspace{-0.14in}
\includegraphics[width=1.75in]{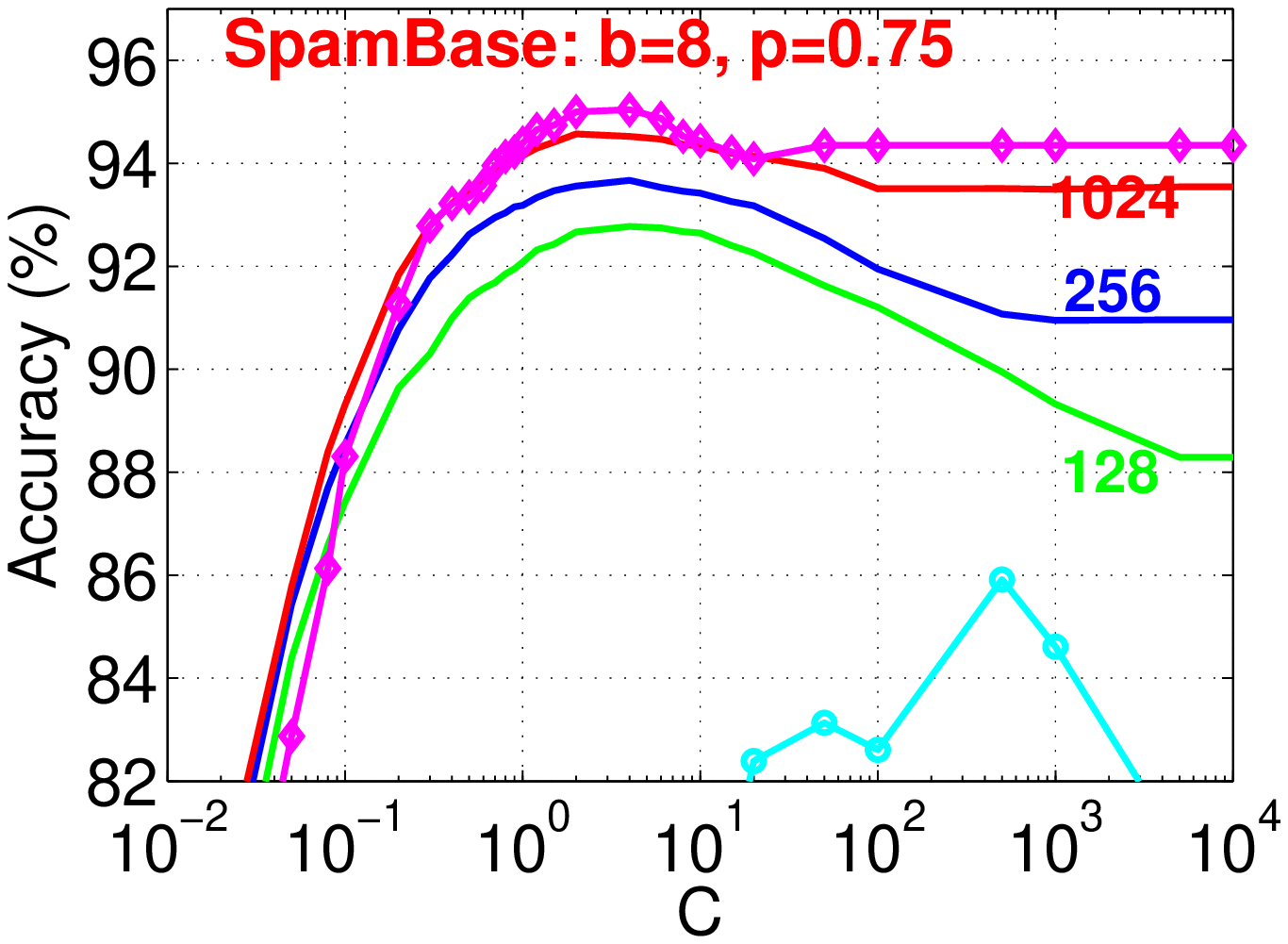}
}

\mbox{
\includegraphics[width=1.75in]{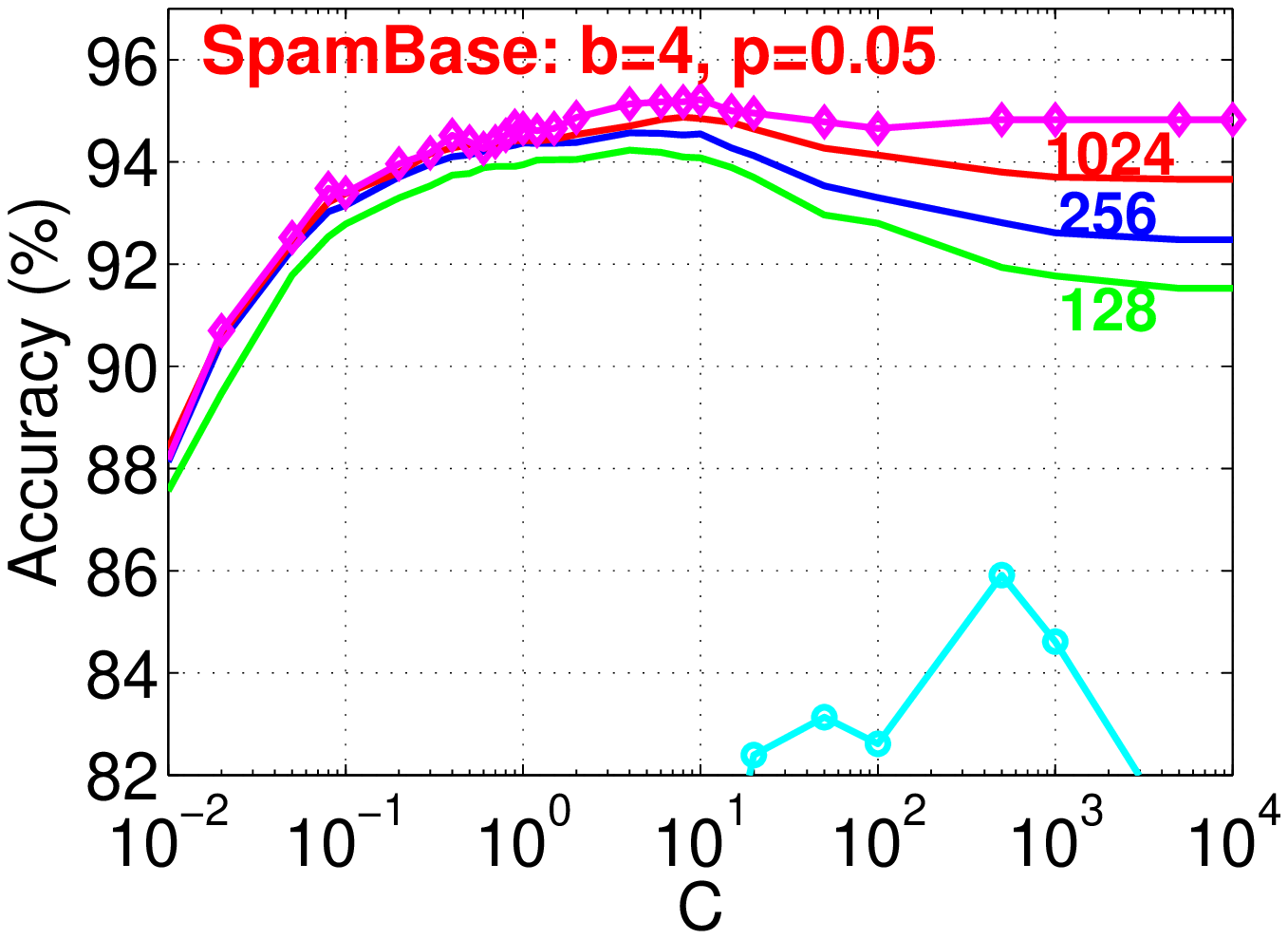}\hspace{-0.14in}
\includegraphics[width=1.75in]{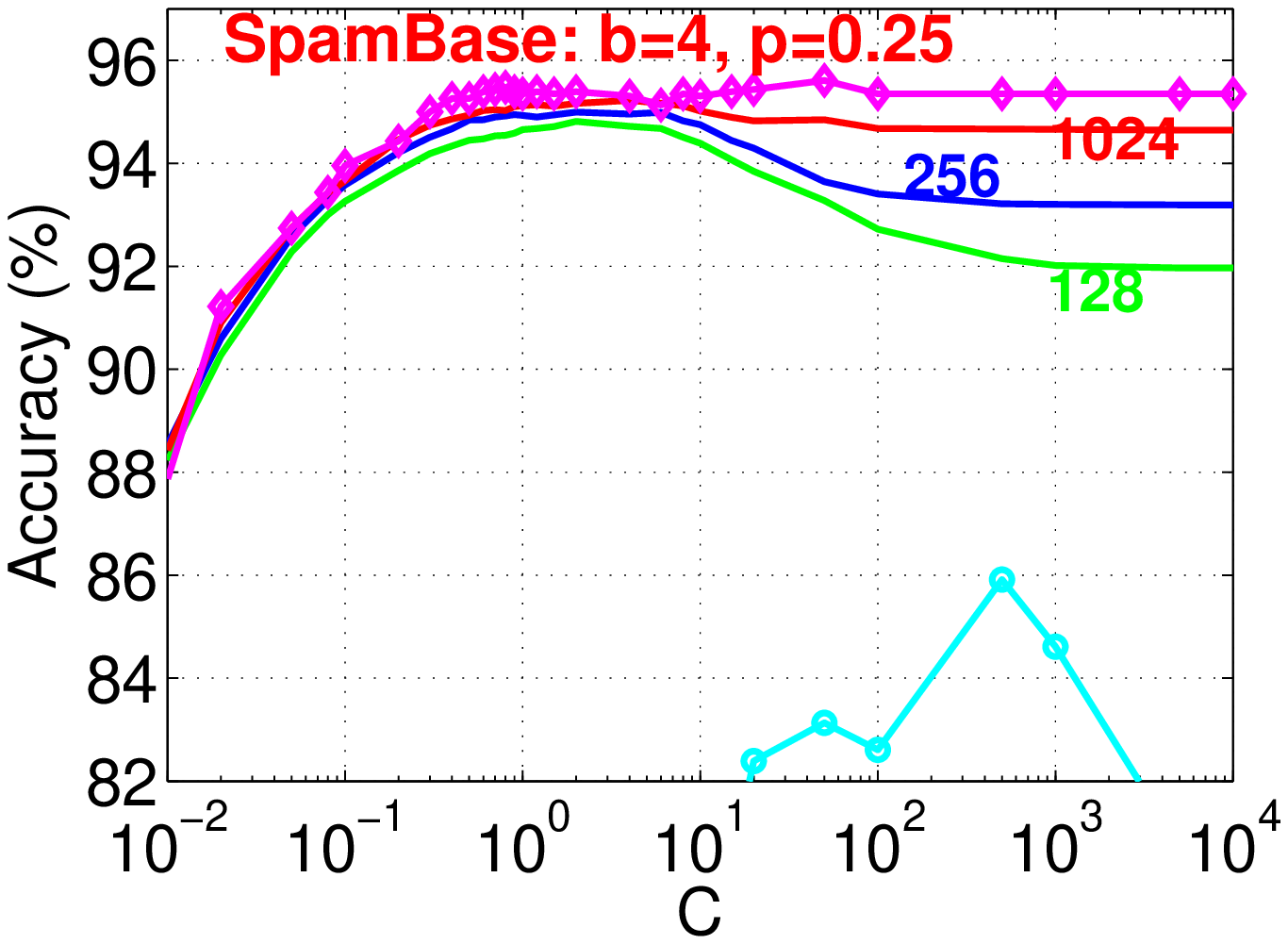}\hspace{-0.14in}
\includegraphics[width=1.75in]{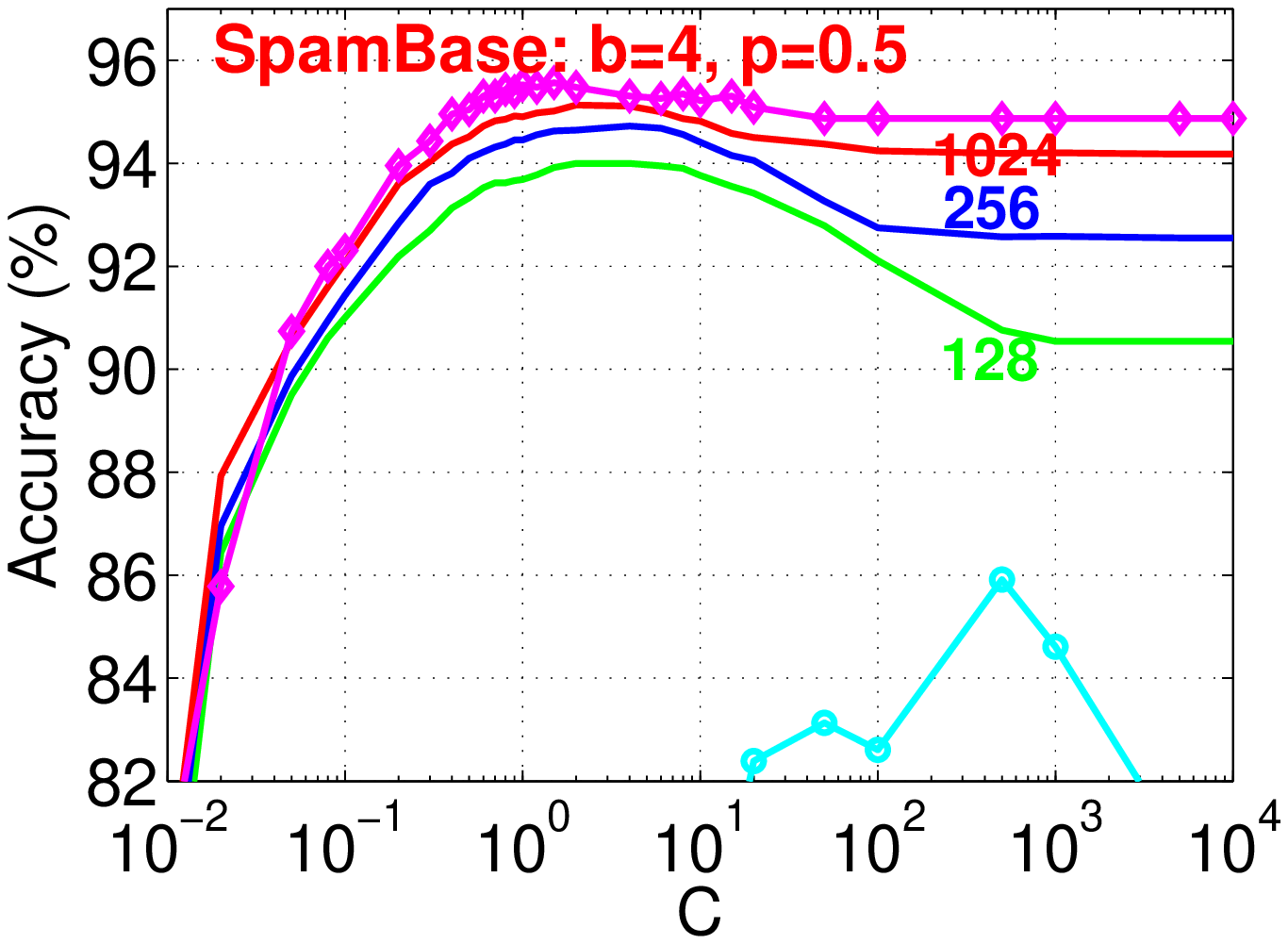}\hspace{-0.14in}
\includegraphics[width=1.75in]{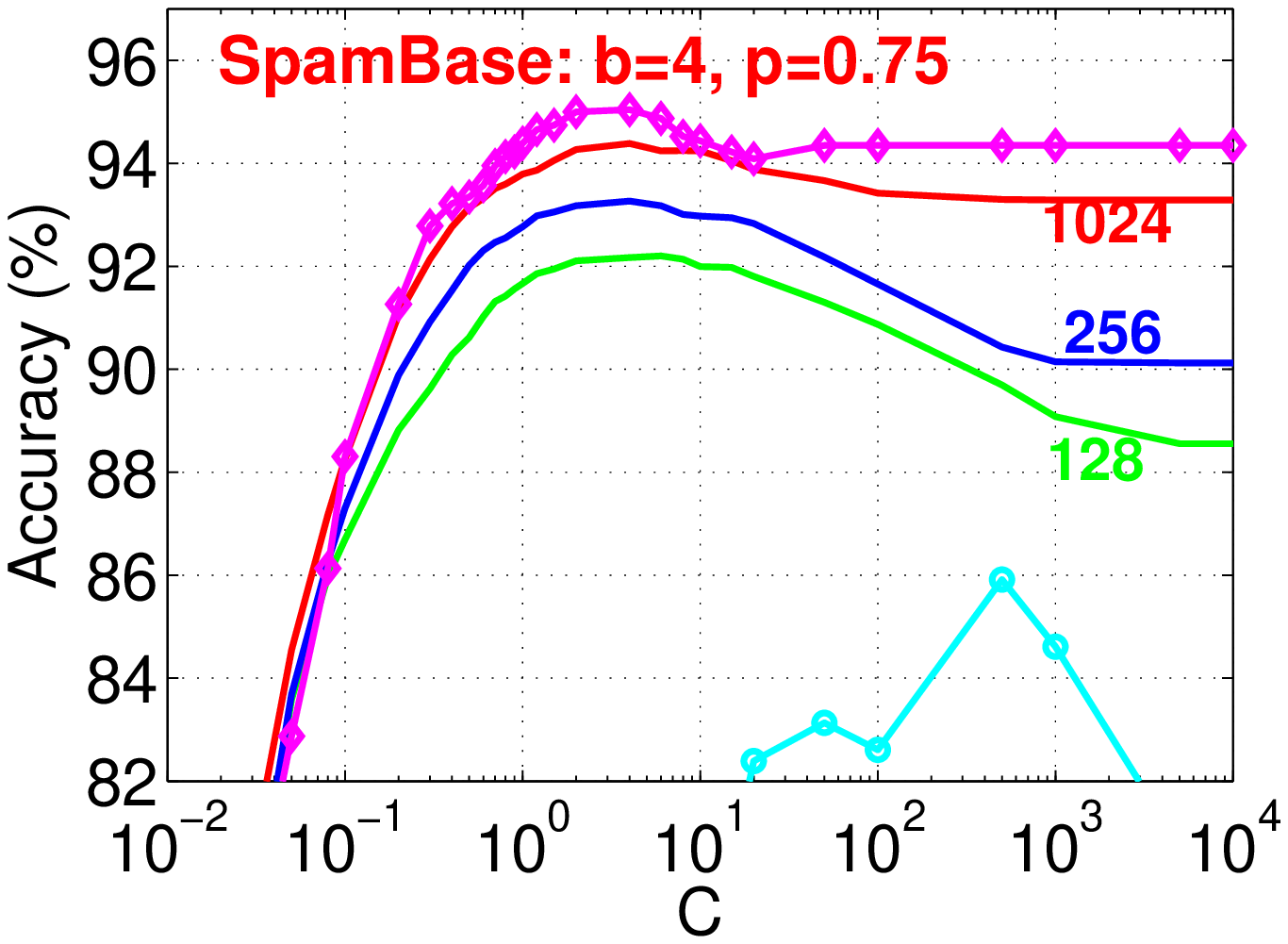}
}

\mbox{
\includegraphics[width=1.75in]{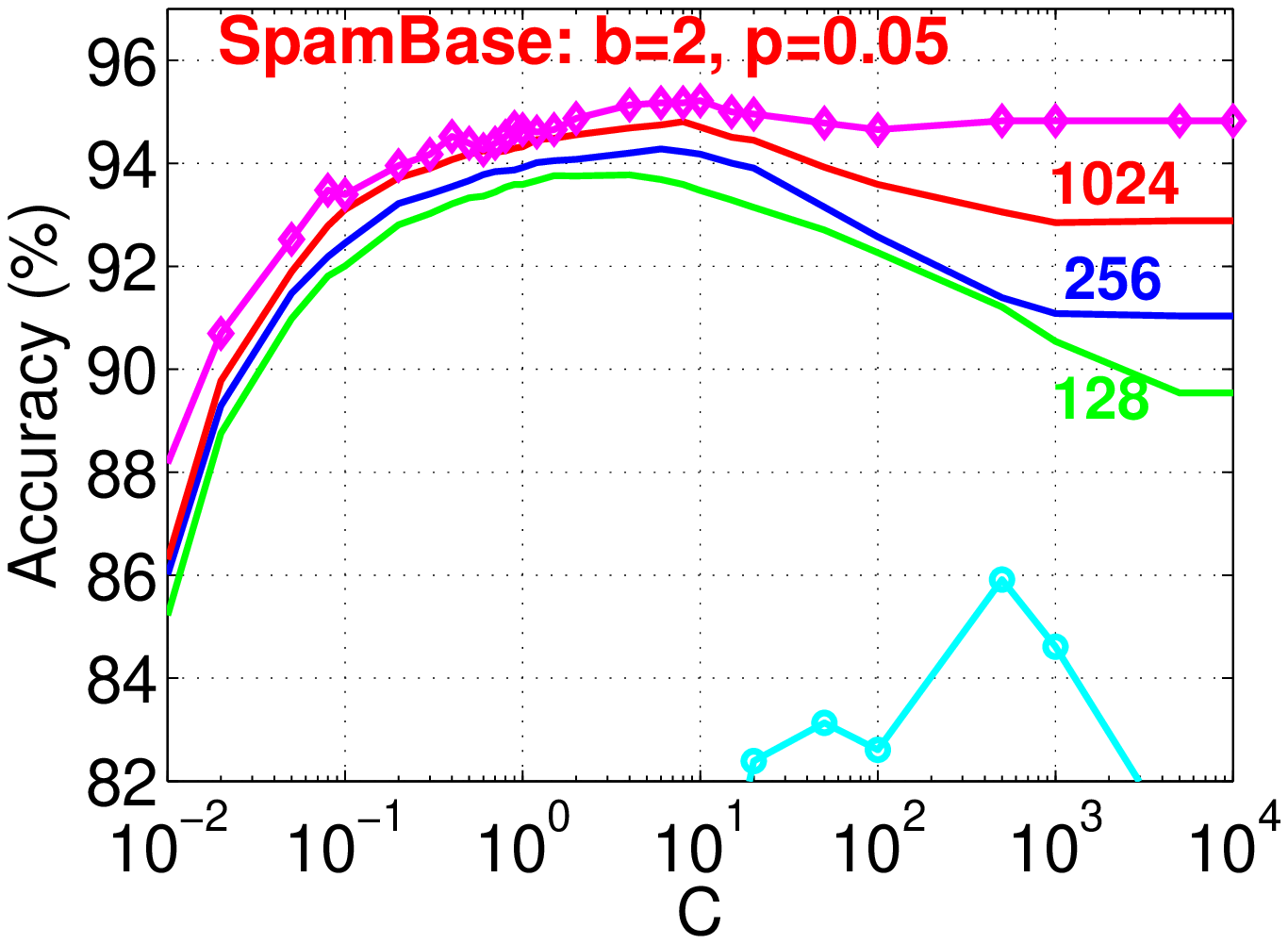}\hspace{-0.14in}
\includegraphics[width=1.75in]{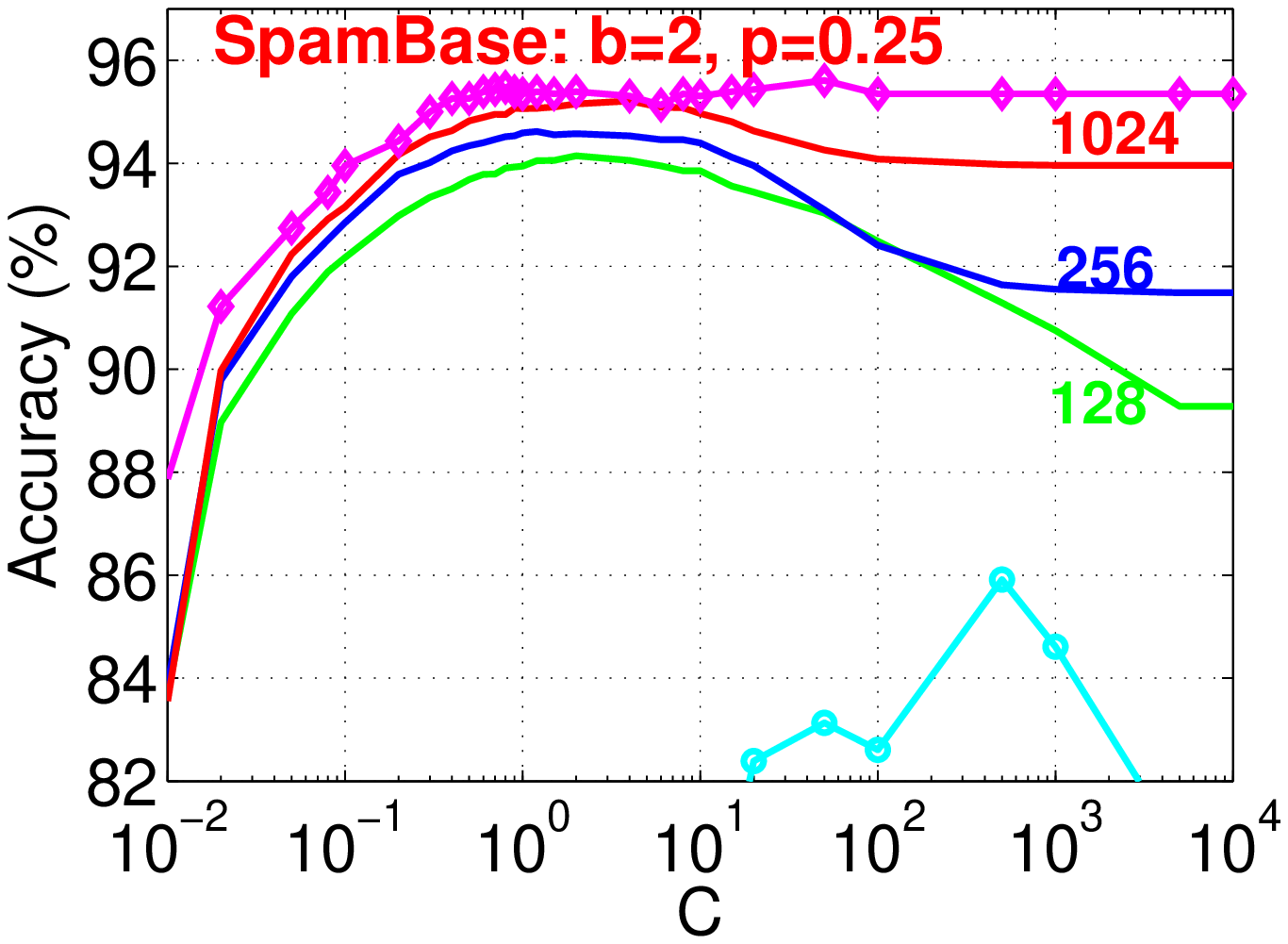}\hspace{-0.14in}
\includegraphics[width=1.75in]{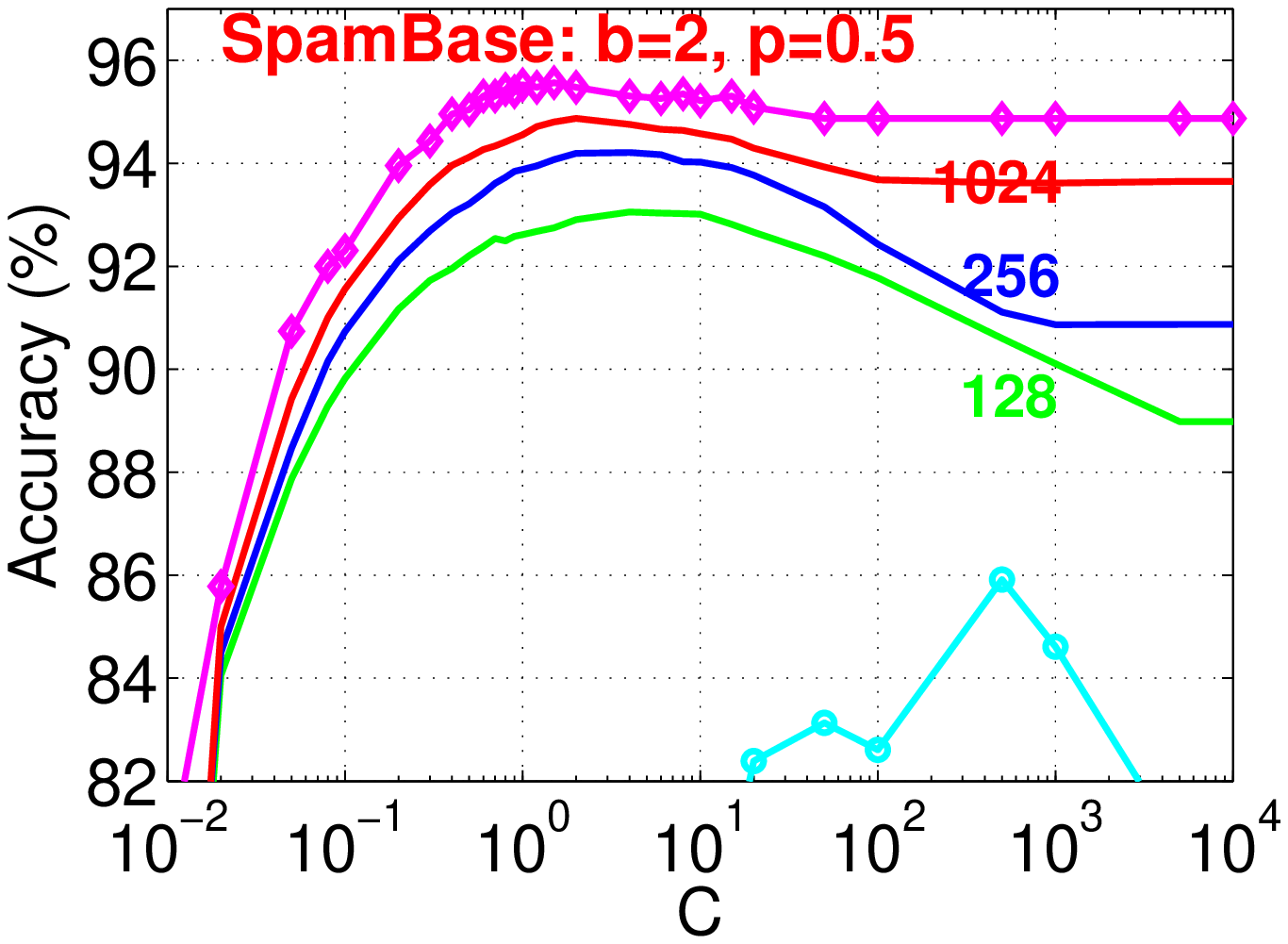}\hspace{-0.14in}
\includegraphics[width=1.75in]{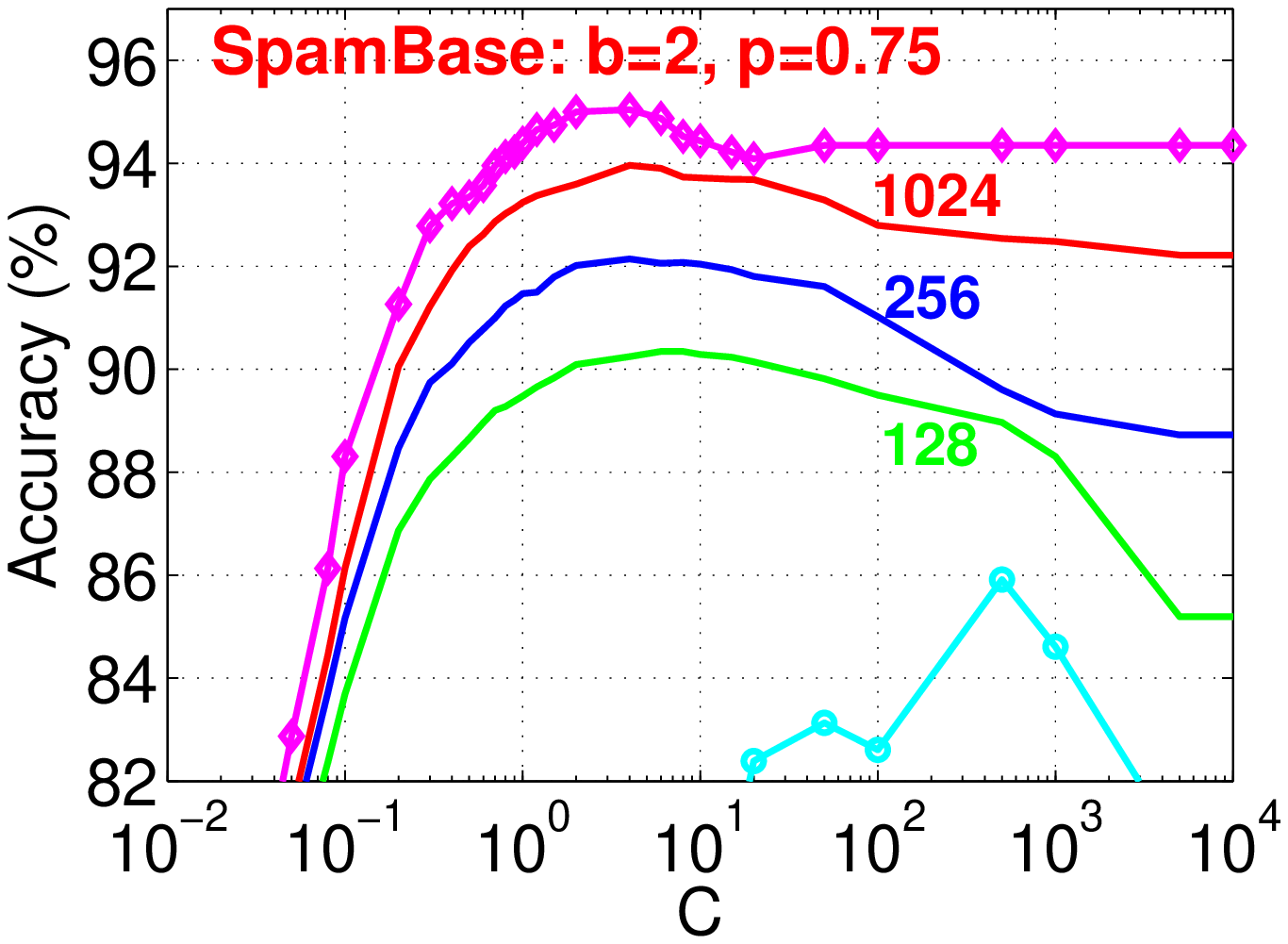}
}

\end{center}
\vspace{-0.15in}
\caption{Test classification accuracies for using linear classifiers combined with hashing in Algorithm~\ref{alg_GCWS} on SpamBase dataset, for $p\in\{0.05, 0.25, 0.5, 0.75\}$ to visualize the trend that, for this dataset, the accuracy decreases with increasing $p$.   }\label{fig_HashSpamBase}\vspace{-0.1in}
\end{figure*}

Figure~\ref{fig_HashCTG10C} (for CTG dataset) and Figure~\ref{fig_HashSpamBase} (for SpamBase dataset) are somewhat different from the previous figures. For both datasets, using $\gamma=0.05$ achieves the best accuracy. We plot the results for $\gamma=0.05, 0.25, 0.5, 0.75$, and $b=8, 4, 2$, to visualize the trend.

\vspace{0.08in}

We have conducted significantly more experiments than we have presented here, but we hope they are convincing enough.

\vspace{0.08in}

\section{Discussion: Hashing $\gamma$GMM and $e$GMM}\hspace{0.1in} At least for $\gamma$ being integers, it is possible to modify the Algorithm~\ref{alg_GCWS} to develop a hashing method for the $\gamma$GMM kernel. Basically, for each hash value, we just need to generate $\gamma$ independent samples. For the original data vectors $u$ and $v$, we require all $\gamma$ samples of $u$ to match all $\gamma$ samples of $v$. The collision probability will be exactly equal to the $\gamma$GMM kernel.

Developing a hashing method for the $e$GMM is more challenging. A more straightforward approach is a two-stage hashing scheme. We first generate hashed values using Algorithm~\ref{alg_GCWS} (with $p=1$), then we apply random Fourier features (RFF)~\cite{Proc:Rahimi_NIPS07} on the hashed values. Based on the analysis in~\cite{Proc:Li_KDD17}, the RFF method needs a very large number of samples in order to reach a satisfactory accuracy. Therefore, we do not expect this two-stage scheme would be practical for hashing the   $e$GMM kernel.

\vspace{-0.1in}
\section{Conclusion}

It is  commonly believed that deep learning algorithms and tree methods can produce the state-of-the-art results in many statistical machine learning tasks. In 2010, ~\cite{Proc:ABC_UAI10} reported a set of surprising experiments on the datasets used by the deep learning community~\cite{Proc:Larochelle_ICML07}, to show that tree methods can outperform deep nets on a majority (but not all) of those datasets and the improvements can be substantial on a good portion of datasets. \cite{Proc:ABC_UAI10} introduced several ideas including the second-order tree-split formula and the new derivatives for multi-class logistic loss function. Nevertheless, tree methods are slow and their model sizes are typically large.

\vspace{0.06in}

\noindent In machine learning practice with massive data, it is desirable to develop algorithms which run almost as efficient as linear methods (such as linear logistic regression or linear SVM) and achieve similar accuracies as nonlinear methods. In this study, the tunable linearized GMM kernels are promising tools for achieving those goals. Our extensive experiments on the same datasets used for testing tree methods and deep nets demonstrate that tunable GMM kernels and their linearized versions through hashing can achieve comparable accuracies as trees. On some datasets, ``abc-robust-logitboost''  achieves better accuracies than the proposed tunable GMM kernels. Also, on some datasets, deep learning methods or RBF kernel SVM outperform tunable GMM kernels. Therefore, there is still  room for future improvements.

\vspace{0.06in}

\noindent In this study, we focus on testing tunable GMM kernels and their linearized versions using classification tasks. It is clear that  these techniques basically generate new data representations and hence can be applied to a wide variety of statistical learning tasks including clustering and regression. Due to the discrete name of the hashed values, the techniques naturally can also be used for building hash tables for fast near neighbor search.

\bibliography{../bib/mybibfile}


\begin{thebibliography}{00}


\ifx \showCODEN    \undefined \def \showCODEN     #1{\unskip}     \fi
\ifx \showDOI      \undefined \def \showDOI       #1{{\tt DOI:}\penalty0{#1}\ }
  \fi
\ifx \showISBNx    \undefined \def \showISBNx     #1{\unskip}     \fi
\ifx \showISBNxiii \undefined \def \showISBNxiii  #1{\unskip}     \fi
\ifx \showISSN     \undefined \def \showISSN      #1{\unskip}     \fi
\ifx \showLCCN     \undefined \def \showLCCN      #1{\unskip}     \fi
\ifx \shownote     \undefined \def \shownote      #1{#1}          \fi
\ifx \showarticletitle \undefined \def \showarticletitle #1{#1}   \fi
\ifx \showURL      \undefined \def \showURL       #1{#1}          \fi
\providecommand\bibfield[2]{#2}
\providecommand\bibinfo[2]{#2}
\providecommand\natexlab[1]{#1}
\providecommand\showeprint[2][]{arXiv:#2}

\bibitem[\protect\citeauthoryear{Bendersky and Croft}{Bendersky and
  Croft}{2009}]%
        {Proc:Bendersky_WSDM09}
\bibfield{author}{\bibinfo{person}{Michael Bendersky} {and}
  \bibinfo{person}{W.~Bruce Croft}.} \bibinfo{year}{2009}\natexlab{}.
\newblock \showarticletitle{Finding text reuse on the web}. In
  \bibinfo{booktitle}{{\em WSDM}}. \bibinfo{address}{Barcelona, Spain},
  \bibinfo{pages}{262--271}.
\newblock
\showISBNx{978-1-60558-390-7}


\bibitem[\protect\citeauthoryear{Bottou}{Bottou}{}]%
        {URL:Bottou_SGD}
\bibfield{author}{\bibinfo{person}{Leon Bottou}.}
\newblock \bibinfo{howpublished}{http://leon.bottou.org/projects/sgd}.
  (\bibinfo{year}{????}).
\newblock


\bibitem[\protect\citeauthoryear{Bottou, Chapelle, DeCoste, and Weston}{Bottou
  et~al\mbox{.}}{2007}]%
        {Book:Bottou_07}
\bibfield{editor}{\bibinfo{person}{L\'eon Bottou}, \bibinfo{person}{Olivier
  Chapelle}, \bibinfo{person}{Dennis DeCoste}, {and} \bibinfo{person}{Jason
  Weston}} (Eds.). \bibinfo{year}{2007}\natexlab{}.
\newblock \bibinfo{booktitle}{{\em Large-Scale Kernel Machines}}.
\newblock \bibinfo{publisher}{The MIT Press}, \bibinfo{address}{Cambridge, MA}.
\newblock


\bibitem[\protect\citeauthoryear{Broder, Glassman, Manasse, and Zweig}{Broder
  et~al\mbox{.}}{1997}]%
        {Proc:Broder_WWW97}
\bibfield{author}{\bibinfo{person}{Andrei~Z. Broder},
  \bibinfo{person}{Steven~C. Glassman}, \bibinfo{person}{Mark~S. Manasse},
  {and} \bibinfo{person}{Geoffrey Zweig}.} \bibinfo{year}{1997}\natexlab{}.
\newblock \showarticletitle{Syntactic clustering of the Web}. In
  \bibinfo{booktitle}{{\em WWW}}. \bibinfo{address}{Santa Clara, CA},
  \bibinfo{pages}{1157 -- 1166}.
\newblock


\bibitem[\protect\citeauthoryear{Charikar}{Charikar}{2002}]%
        {Proc:Charikar}
\bibfield{author}{\bibinfo{person}{Moses~S. Charikar}.}
  \bibinfo{year}{2002}\natexlab{}.
\newblock \showarticletitle{Similarity estimation techniques from rounding
  algorithms}. In \bibinfo{booktitle}{{\em STOC}}. \bibinfo{address}{Montreal,
  Canada}, \bibinfo{pages}{380--388}.
\newblock


\bibitem[\protect\citeauthoryear{Cherkasova, Eshghi, III, Tucek, and
  Veitch}{Cherkasova et~al\mbox{.}}{2009}]%
        {Proc:Cherkasova_KDD09}
\bibfield{author}{\bibinfo{person}{Ludmila Cherkasova}, \bibinfo{person}{Kave
  Eshghi}, \bibinfo{person}{Charles B.~Morrey III}, \bibinfo{person}{Joseph
  Tucek}, {and} \bibinfo{person}{Alistair~C. Veitch}.}
  \bibinfo{year}{2009}\natexlab{}.
\newblock \showarticletitle{Applying syntactic similarity algorithms for
  enterprise information management}. In \bibinfo{booktitle}{{\em KDD}}.
  \bibinfo{address}{Paris, France}, \bibinfo{pages}{1087--1096}.
\newblock


\bibitem[\protect\citeauthoryear{Chierichetti, Kumar, Lattanzi, Mitzenmacher,
  Panconesi, and Raghavan}{Chierichetti et~al\mbox{.}}{2009}]%
        {Proc:Chierichetti_KDD09}
\bibfield{author}{\bibinfo{person}{Flavio Chierichetti}, \bibinfo{person}{Ravi
  Kumar}, \bibinfo{person}{Silvio Lattanzi}, \bibinfo{person}{Michael
  Mitzenmacher}, \bibinfo{person}{Alessandro Panconesi}, {and}
  \bibinfo{person}{Prabhakar Raghavan}.} \bibinfo{year}{2009}\natexlab{}.
\newblock \showarticletitle{On compressing social networks}. In
  \bibinfo{booktitle}{{\em KDD}}. \bibinfo{address}{Paris, France},
  \bibinfo{pages}{219--228}.
\newblock


\bibitem[\protect\citeauthoryear{Dourisboure, Geraci, and
  Pellegrini}{Dourisboure et~al\mbox{.}}{2009}]%
        {Article:Dourisboure09}
\bibfield{author}{\bibinfo{person}{Yon Dourisboure}, \bibinfo{person}{Filippo
  Geraci}, {and} \bibinfo{person}{Marco Pellegrini}.}
  \bibinfo{year}{2009}\natexlab{}.
\newblock \showarticletitle{Extraction and classification of dense implicit
  communities in the Web graph}.
\newblock \bibinfo{journal}{{\em ACM Trans. Web\/}} \bibinfo{volume}{3},
  \bibinfo{number}{2} (\bibinfo{year}{2009}), \bibinfo{pages}{1--36}.
\newblock


\bibitem[\protect\citeauthoryear{Fan, Chang, Hsieh, Wang, and Lin}{Fan
  et~al\mbox{.}}{2008}]%
        {Article:Fan_JMLR08}
\bibfield{author}{\bibinfo{person}{Rong-En Fan}, \bibinfo{person}{Kai-Wei
  Chang}, \bibinfo{person}{Cho-Jui Hsieh}, \bibinfo{person}{Xiang-Rui Wang},
  {and} \bibinfo{person}{Chih-Jen Lin}.} \bibinfo{year}{2008}\natexlab{}.
\newblock \showarticletitle{LIBLINEAR: A Library for Large Linear
  Classification}.
\newblock \bibinfo{journal}{{\em Journal of Machine Learning Research\/}}
  \bibinfo{volume}{9} (\bibinfo{year}{2008}), \bibinfo{pages}{1871--1874}.
\newblock


\bibitem[\protect\citeauthoryear{Fetterly, Manasse, Najork, and
  Wiener}{Fetterly et~al\mbox{.}}{2003}]%
        {Proc:Fetterly_WWW03}
\bibfield{author}{\bibinfo{person}{Dennis Fetterly}, \bibinfo{person}{Mark
  Manasse}, \bibinfo{person}{Marc Najork}, {and} \bibinfo{person}{Janet~L.
  Wiener}.} \bibinfo{year}{2003}\natexlab{}.
\newblock \showarticletitle{A large-scale study of the evolution of web pages}.
  In \bibinfo{booktitle}{{\em WWW}}. \bibinfo{address}{Budapest, Hungary},
  \bibinfo{pages}{669--678}.
\newblock


\bibitem[\protect\citeauthoryear{Forman, Eshghi, and Suermondt}{Forman
  et~al\mbox{.}}{2009}]%
        {Article:Forman09}
\bibfield{author}{\bibinfo{person}{George Forman}, \bibinfo{person}{Kave
  Eshghi}, {and} \bibinfo{person}{Jaap Suermondt}.}
  \bibinfo{year}{2009}\natexlab{}.
\newblock \showarticletitle{Efficient detection of large-scale redundancy in
  enterprise file systems}.
\newblock \bibinfo{journal}{{\em SIGOPS Oper. Syst. Rev.\/}}
  \bibinfo{volume}{43}, \bibinfo{number}{1} (\bibinfo{year}{2009}),
  \bibinfo{pages}{84--91}.
\newblock
\showISSN{0163-5980}


\bibitem[\protect\citeauthoryear{Friedman}{Friedman}{2001}]%
        {Article:Friedman_AS01}
\bibfield{author}{\bibinfo{person}{Jerome~H. Friedman}.}
  \bibinfo{year}{2001}\natexlab{}.
\newblock \showarticletitle{Greedy function approximation: A gradient boosting
  machine}.
\newblock \bibinfo{journal}{{\em The Annals of Statistics\/}}
  \bibinfo{volume}{29}, \bibinfo{number}{5} (\bibinfo{year}{2001}),
  \bibinfo{pages}{1189--1232}.
\newblock


\bibitem[\protect\citeauthoryear{Friedman, Hastie, and Tibshirani}{Friedman
  et~al\mbox{.}}{2000}]%
        {Article:FHT_AS00}
\bibfield{author}{\bibinfo{person}{Jerome~H. Friedman},
  \bibinfo{person}{Trevor~J. Hastie}, {and} \bibinfo{person}{Robert
  Tibshirani}.} \bibinfo{year}{2000}\natexlab{}.
\newblock \showarticletitle{Additive logistic regression: a statistical view of
  boosting}.
\newblock \bibinfo{journal}{{\em The Annals of Statistics\/}}
  \bibinfo{volume}{28}, \bibinfo{number}{2} (\bibinfo{year}{2000}),
  \bibinfo{pages}{337--407}.
\newblock


\bibitem[\protect\citeauthoryear{Gollapudi and Sharma}{Gollapudi and
  Sharma}{2009}]%
        {Proc:Gollapudi_WWW09}
\bibfield{author}{\bibinfo{person}{Sreenivas Gollapudi} {and}
  \bibinfo{person}{Aneesh Sharma}.} \bibinfo{year}{2009}\natexlab{}.
\newblock \showarticletitle{An axiomatic approach for result diversification}.
  In \bibinfo{booktitle}{{\em WWW}}. \bibinfo{address}{Madrid, Spain},
  \bibinfo{pages}{381--390}.
\newblock


\bibitem[\protect\citeauthoryear{Ioffe}{Ioffe}{2010}]%
        {Proc:Ioffe_ICDM10}
\bibfield{author}{\bibinfo{person}{Sergey Ioffe}.}
  \bibinfo{year}{2010}\natexlab{}.
\newblock \showarticletitle{Improved Consistent Sampling, Weighted Minhash and
  \text{L1} Sketching}. In \bibinfo{booktitle}{{\em ICDM}}.
  \bibinfo{address}{Sydney, AU}, \bibinfo{pages}{246--255}.
\newblock


\bibitem[\protect\citeauthoryear{Kleinberg and Tardos}{Kleinberg and
  Tardos}{1999}]%
        {Proc:Kleinberg_FOCS99}
\bibfield{author}{\bibinfo{person}{Jon Kleinberg} {and} \bibinfo{person}{Eva
  Tardos}.} \bibinfo{year}{1999}\natexlab{}.
\newblock \showarticletitle{Approximation Algorithms for Classification
  Problems with Pairwise Relationships: Metric Labeling and \text{Markov}
  Random Fields}. In \bibinfo{booktitle}{{\em FOCS}}. \bibinfo{address}{New
  York}, \bibinfo{pages}{14--23}.
\newblock


\bibitem[\protect\citeauthoryear{Larochelle, Erhan, Courville, Bergstra, and
  Bengio}{Larochelle et~al\mbox{.}}{2007}]%
        {Proc:Larochelle_ICML07}
\bibfield{author}{\bibinfo{person}{Hugo Larochelle}, \bibinfo{person}{Dumitru
  Erhan}, \bibinfo{person}{Aaron~C. Courville}, \bibinfo{person}{James
  Bergstra}, {and} \bibinfo{person}{Yoshua Bengio}.}
  \bibinfo{year}{2007}\natexlab{}.
\newblock \showarticletitle{An empirical evaluation of deep architectures on
  problems with many factors of variation}. In \bibinfo{booktitle}{{\em ICML}}.
  \bibinfo{address}{Corvalis, Oregon}, \bibinfo{pages}{473--480}.
\newblock


\bibitem[\protect\citeauthoryear{Li}{Li}{2008}]%
        {Article:Li_ABC_arXiv08}
\bibfield{author}{\bibinfo{person}{Ping Li}.} \bibinfo{year}{2008}\natexlab{}.
\newblock \showarticletitle{Adaptive Base Class Boost for Multi-class
  Classification}.
\newblock \bibinfo{journal}{{\em CoRR\/}}  \bibinfo{volume}{abs/0811.1250}
  (\bibinfo{year}{2008}).
\newblock


\bibitem[\protect\citeauthoryear{Li}{Li}{2009}]%
        {Proc:ABC_ICML09}
\bibfield{author}{\bibinfo{person}{Ping Li}.} \bibinfo{year}{2009}\natexlab{}.
\newblock \showarticletitle{ABC-Boost: Adaptive Base Class Boost for
  Multi-Class Classification}. In \bibinfo{booktitle}{{\em ICML}}.
  \bibinfo{address}{Montreal, Canada}, \bibinfo{pages}{625--632}.
\newblock


\bibitem[\protect\citeauthoryear{Li}{Li}{2010}]%
        {Proc:ABC_UAI10}
\bibfield{author}{\bibinfo{person}{Ping Li}.} \bibinfo{year}{2010}\natexlab{}.
\newblock \showarticletitle{Robust LogitBoost and Adaptive Base Class (ABC)
  LogitBoost}. In \bibinfo{booktitle}{{\em UAI}}. \bibinfo{address}{Catalina
  Island, CA}.
\newblock


\bibitem[\protect\citeauthoryear{Li}{Li}{2015}]%
        {Proc:Li_KDD15}
\bibfield{author}{\bibinfo{person}{Ping Li}.} \bibinfo{year}{2015}\natexlab{}.
\newblock \showarticletitle{0-Bit Consistent Weighted Sampling}. In
  \bibinfo{booktitle}{{\em KDD}}. \bibinfo{address}{Sydney, Australia},
  \bibinfo{pages}{665--674}.
\newblock


\bibitem[\protect\citeauthoryear{Li}{Li}{2016}]%
        {Report:Li_GMM16}
\bibfield{author}{\bibinfo{person}{Ping Li}.} \bibinfo{year}{2016}\natexlab{}.
\newblock \bibinfo{booktitle}{{\em Linearized \text{GMM} Kernels and Normalized
  Random Fourier Features}}.
\newblock \bibinfo{type}{{T}echnical {R}eport}.
  \bibinfo{institution}{arXiv:1605.05721}.
\newblock


\bibitem[\protect\citeauthoryear{Li}{Li}{2017}]%
        {Proc:Li_KDD17}
\bibfield{author}{\bibinfo{person}{Ping Li}.} \bibinfo{year}{2017}\natexlab{}.
\newblock \showarticletitle{Linearized GMM Kernels and Normalized Random
  Fourier Features}. In \bibinfo{booktitle}{{\em KDD}}.
  \bibinfo{pages}{315--324}.
\newblock


\bibitem[\protect\citeauthoryear{Li and {K\"{o}nig}}{Li and
  {K\"{o}nig}}{2010}]%
        {Proc:Li_Konig_WWW10}
\bibfield{author}{\bibinfo{person}{Ping Li} {and}
  \bibinfo{person}{Arnd~Christian {K\"{o}nig}}.}
  \bibinfo{year}{2010}\natexlab{}.
\newblock \showarticletitle{b-Bit Minwise Hashing}. In \bibinfo{booktitle}{{\em
  WWW}}. \bibinfo{address}{Raleigh, NC}, \bibinfo{pages}{671--680}.
\newblock


\bibitem[\protect\citeauthoryear{Li, Shrivastava, Moore, and K\"onig}{Li
  et~al\mbox{.}}{2011}]%
        {Proc:HashLearning_NIPS11}
\bibfield{author}{\bibinfo{person}{Ping Li}, \bibinfo{person}{Anshumali
  Shrivastava}, \bibinfo{person}{Joshua Moore}, {and}
  \bibinfo{person}{Arnd~Christian K\"onig}.} \bibinfo{year}{2011}\natexlab{}.
\newblock \showarticletitle{Hashing Algorithms for Large-Scale Learning}. In
  \bibinfo{booktitle}{{\em NIPS}}. \bibinfo{address}{Granada, Spain},
  \bibinfo{pages}{2672--2680}.
\newblock


\bibitem[\protect\citeauthoryear{Manasse, McSherry, and Talwar}{Manasse
  et~al\mbox{.}}{2010}]%
        {Report:Manasse_CWS10}
\bibfield{author}{\bibinfo{person}{Mark Manasse}, \bibinfo{person}{Frank
  McSherry}, {and} \bibinfo{person}{Kunal Talwar}.}
  \bibinfo{year}{2010}\natexlab{}.
\newblock \bibinfo{booktitle}{{\em Consistent Weighted Sampling}}.
\newblock \bibinfo{type}{{T}echnical {R}eport} MSR-TR-2010-73.
  \bibinfo{institution}{Microsoft Research}.
\newblock


\bibitem[\protect\citeauthoryear{Najork, Gollapudi, and Panigrahy}{Najork
  et~al\mbox{.}}{2009}]%
        {Proc:Najork_WSDM09}
\bibfield{author}{\bibinfo{person}{Marc Najork}, \bibinfo{person}{Sreenivas
  Gollapudi}, {and} \bibinfo{person}{Rina Panigrahy}.}
  \bibinfo{year}{2009}\natexlab{}.
\newblock \showarticletitle{Less is more: sampling the neighborhood graph makes
  SALSA better and faster}. In \bibinfo{booktitle}{{\em WSDM}}.
  \bibinfo{address}{Barcelona, Spain}, \bibinfo{pages}{242--251}.
\newblock


\bibitem[\protect\citeauthoryear{Nystr{\"o}m}{Nystr{\"o}m}{1930}]%
        {Article:Nystrom1930}
\bibfield{author}{\bibinfo{person}{E.~J. Nystr{\"o}m}.}
  \bibinfo{year}{1930}\natexlab{}.
\newblock \showarticletitle{{\"U}ber Die Praktische Aufl{\"o}sung von
  Integralgleichungen mit Anwendungen auf Randwertaufgaben}.
\newblock \bibinfo{journal}{{\em Acta Mathematica\/}} \bibinfo{volume}{54},
  \bibinfo{number}{1} (\bibinfo{year}{1930}), \bibinfo{pages}{185--204}.
\newblock


\bibitem[\protect\citeauthoryear{Rahimi and Recht}{Rahimi and Recht}{2007}]%
        {Proc:Rahimi_NIPS07}
\bibfield{author}{\bibinfo{person}{A. Rahimi} {and} \bibinfo{person}{B.
  Recht}.} \bibinfo{year}{2007}\natexlab{}.
\newblock \showarticletitle{Random features for large-scale kernel machines}.
  In \bibinfo{booktitle}{{\em NIPS}}. \bibinfo{address}{Vancouver, Canada},
  \bibinfo{pages}{1177--1184}.
\newblock


\bibitem[\protect\citeauthoryear{Sch\"olkopf and Smola}{Sch\"olkopf and
  Smola}{2002}]%
        {Book:Scholkopf_02}
\bibfield{author}{\bibinfo{person}{Bernhard Sch\"olkopf} {and}
  \bibinfo{person}{Alexander~J. Smola}.} \bibinfo{year}{2002}\natexlab{}.
\newblock \bibinfo{booktitle}{{\em Learning with Kernels}}.
\newblock \bibinfo{publisher}{The MIT Press}, \bibinfo{address}{Cambridge, MA}.
\newblock


\end{thebibliography}

\bibliographystyle{ACM-Reference-Format}

\end{document}